\documentclass[twoside,11pt]{article}

\usepackage{jmlr2e}

\usepackage[utf8]{inputenc}
\usepackage[T1]{fontenc}
\usepackage{lmodern}
\usepackage{url}
\usepackage{booktabs}
\usepackage{amsfonts}
\usepackage{nicefrac}
\usepackage{microtype}
\usepackage{enumitem}
\usepackage{xfrac}
\usepackage{amsmath}
\usepackage{changepage}
\usepackage{xcolor}
\usepackage{rotating}
\usepackage{amssymb}

\definecolor{darkblue}{rgb}{0, 0.2, 0.7}
\hypersetup{
    colorlinks = true,
    linkcolor = darkblue,
    anchorcolor = darkblue,
    citecolor = darkblue,
    filecolor = darkblue,
    urlcolor = darkblue
}

\newlength{\offsetpage}
\newenvironment{widepage}{\begin{adjustwidth}{-\offsetpage}{-\offsetpage}%
    \addtolength{\textwidth}{2\offsetpage}}%
{\end{adjustwidth}}

\newcommand{\xa}{\makebox[0pt][l]{$^a$}}
\newcommand{\xb}{\makebox[0pt][l]{$^b$}}
\newcommand{\xc}{\makebox[0pt][l]{$^c$}}
\newcommand{\xd}{\makebox[0pt][l]{$^d$}}
\newcommand{\xe}{\makebox[0pt][l]{$^e$}}
\newcommand{\xf}{\makebox[0pt][l]{$^f$}}
\newcommand{\xg}{\makebox[0pt][l]{$^g$}}

\newcommand{\bsl}{\makebox[0pt][r]{\raisebox{0.05em}{$\bigstar\,$}}}

\def\*#1{\mathbf{#1}}

\makeatletter
\newcommand\footnoteref[1]{\protected@xdef\@thefnmark{\ref{#1}}\@footnotemark}
\makeatother

\usepackage[noabbrev,capitalize]{cleveref}

\usepackage{lastpage}
\jmlrheading{21}{2020}{1-\pageref{LastPage}}{1/20; Revised 6/20}{6/20}{20-074}{Colin Raffel, Noam Shazeer, Adam Roberts, Katherine Lee, Sharan Narang, Michael Matena, Yanqi Zhou, Wei Li, and Peter J. Liu}

\ShortHeadings{Exploring the Limits of Transfer Learning}{Raffel, Shazeer, Roberts, Lee, Narang, Matena, Zhou, Li and Liu}
\firstpageno{1}

\begin{document}

\title{Exploring the Limits of Transfer Learning with a Unified Text-to-Text Transformer}

\author{\name Colin Raffel\thanks{Equal contribution. A description of each author's contribution is available in \cref{sec:contributions}. Correspondence to \texttt{craffel@gmail.com}.} \email craffel@gmail.com
  \AND
  \name Noam Shazeer$^*$ \email noam@google.com
  \AND
  \name Adam Roberts$^*$ \email adarob@google.com
  \AND
  \name Katherine Lee$^*$ \email katherinelee@google.com
  \AND
  \name Sharan Narang \email sharannarang@google.com
  \AND
  \name Michael Matena \email mmatena@google.com
  \AND
  \name Yanqi Zhou \email yanqiz@google.com
  \AND
  Wei Li \email mweili@google.com
  \AND
  \name Peter J.\ Liu \email peterjliu@google.com
  \AND \addr Google, Mountain View, CA 94043, USA
}

\editor{Ivan Titov}

\maketitle

\begin{abstract}%
Transfer learning, where a model is first pre-trained on a data-rich task before being fine-tuned on a downstream task, has emerged as a powerful technique in natural language processing (NLP).
The effectiveness of transfer learning has given rise to a diversity of approaches, methodology, and practice.
In this paper, we explore the landscape of transfer learning techniques for NLP by introducing a unified framework that converts all text-based language problems into a text-to-text format.
Our systematic study compares pre-training objectives, architectures, unlabeled data sets, transfer approaches, and other factors on dozens of language understanding tasks.
By combining the insights from our exploration with scale and our new ``Colossal Clean Crawled Corpus'', we achieve state-of-the-art results on many benchmarks covering summarization, question answering, text classification, and more.
To facilitate future work on transfer learning for NLP, we release our data set, pre-trained models, and code.\footnote{\label{fn:oss}\url{https://github.com/google-research/text-to-text-transfer-transformer}}
\end{abstract}

\begin{keywords}
  transfer learning, natural language processing, multi-task learning, attention-based models, deep learning
\end{keywords}

\section{Introduction}

Training a machine learning model to perform natural language processing (NLP) tasks often requires that the model can process text in a way that is amenable to downstream learning.
This can be loosely viewed as developing general-purpose knowledge that allows the model to ``understand'' text.
This knowledge can range from low-level (e.g.\ the spelling or meaning of words) to high-level (e.g.\ that a tuba is too large to fit in most backpacks).
In modern machine learning practice, providing this knowledge is rarely done explicitly; instead, it is often learned as part of an auxiliary task.
For example, a historically common approach is to use word vectors \citep{mikolov2013distributed,mikolov2013efficient,pennington2014glove} to map word identities to a continuous representation where, ideally, similar words map to similar vectors.
These vectors are often learned through an objective that, for example, encourages co-occurring words to be positioned nearby in the continuous space \citep{mikolov2013distributed}.

Recently, it has become increasingly common to pre-train the entire model on a data-rich task.
Ideally, this pre-training causes the model to develop general-purpose abilities and knowledge that can then be transferred to downstream tasks.
In applications of transfer learning to computer vision \citep{oquab2014learning,jia2014caffe,huh2016makes,yosinski2014transferable}, pre-training is typically done via supervised learning on a large labeled data set like ImageNet \citep{russakovsky2015imagenet,deng2009imagenet}.
In contrast, modern techniques for transfer learning in NLP often pre-train using unsupervised learning on unlabeled data.
This approach has recently been used to obtain state-of-the-art results in many of the most common NLP benchmarks \citep{devlin2018bert,yang2019xlnet,dong2019unified,liu2019roberta,lan2019albert}.
Beyond its empirical strength, unsupervised pre-training for NLP is particularly attractive because unlabeled text data is available en masse thanks to the Internet---for example, the Common Crawl project\footnote{\url{http://commoncrawl.org}} produces about 20TB of text data extracted from web pages each month.
This is a natural fit for neural networks, which have been shown to exhibit remarkable scalability, i.e.\ it is often possible to achieve better performance simply by training a larger model on a larger data set \citep{hestness2017deep,shazeer2017outrageously,jozefowicz2016exploring,mahajan2018exploring,radford2019language,shazeer2018mesh,huang2018gpipe,keskar2019ctrl}.

This synergy has resulted in a great deal of recent work developing transfer learning methodology for NLP, which has produced a wide landscape of pre-training objectives \citep{howard2018universal,devlin2018bert,yang2019xlnet,dong2019unified}, unlabeled data sets \citep{yang2019xlnet,liu2019roberta,zellers2019defending}, benchmarks \citep{wang2019superglue,wang2018glue,conneau2018senteval}, fine-tuning methods \citep{howard2018universal,houlsby2019parameter,peters2019tune}, and more.
The rapid rate of progress and diversity of techniques in this burgeoning field can make it difficult to compare different algorithms, tease apart the effects of new contributions, and understand the space of existing methods for transfer learning.
Motivated by a need for more rigorous understanding, we leverage a unified approach to transfer learning that allows us to systematically study different approaches and push the current limits of the field.

\begin{figure}[t]
    \centering
    \includegraphics[width=\textwidth]{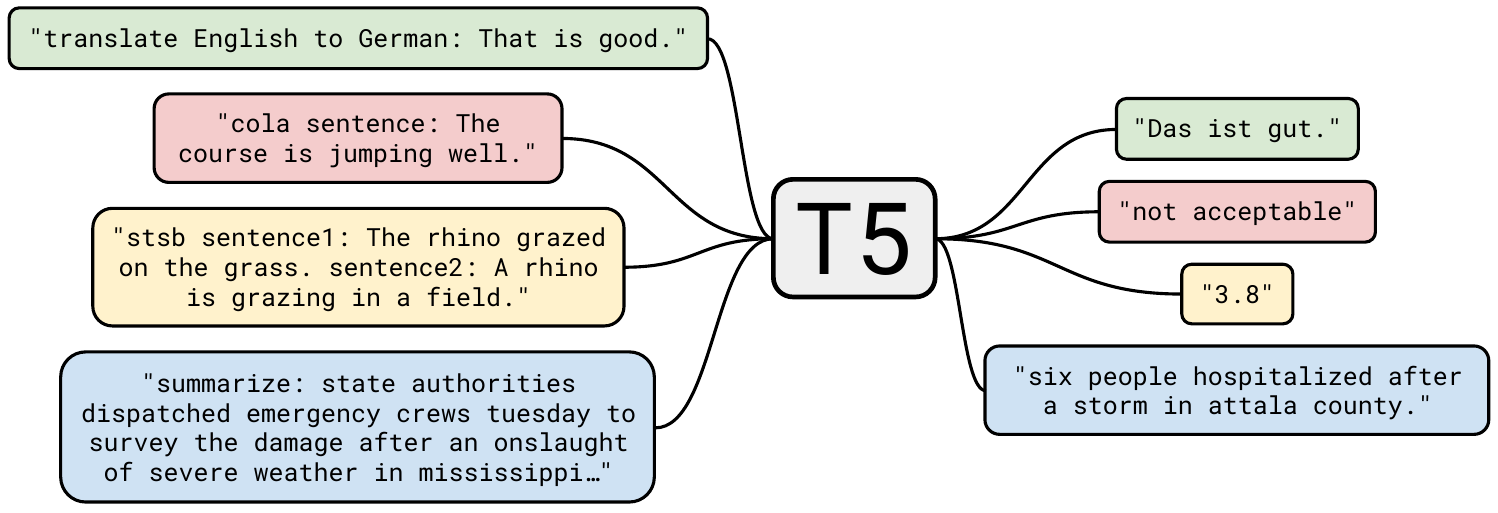}
    \caption{
    A diagram of our text-to-text framework.
    Every task we consider---including translation, question answering, and classification---is cast as feeding our model text as input and training it to generate some target text.
    This allows us to use the same model, loss function, hyperparameters, etc.\ across our diverse set of tasks.
    It also provides a standard testbed for the methods included in our empirical survey.
    ``T5'' refers to our model, which we dub the ``\textbf{T}ext-\textbf{t}o-\textbf{T}ext \textbf{T}ransfer \textbf{T}ransformer''.
    }
    \label{fig:text_to_text}
\end{figure}

The basic idea underlying our work is to treat every text processing problem as a ``text-to-text'' problem, i.e.\ taking text as input and producing new text as output.
This approach is inspired by previous unifying frameworks for NLP tasks, including casting all text problems as question answering  \citep{mccann2018natural}, language modeling \citep{radford2019language}, or span extraction \cite{keskar2019unifying} tasks.
Crucially, the text-to-text framework allows us to directly apply the same model, objective, training procedure, and decoding process to every task we consider.
We leverage this flexibility by evaluating performance on a wide variety of English-based NLP problems, including question answering, document summarization, and sentiment classification, to name a few.
With this unified approach, we can compare the effectiveness of different transfer learning objectives, unlabeled data sets, and other factors, while exploring the limits of transfer learning for NLP by scaling up models and data sets beyond what has previously been considered.

We emphasize that our goal is not to propose new methods but instead to provide a comprehensive perspective on where the field stands.
As such, our work primarily comprises a survey, exploration, and empirical comparison of existing techniques.
% Our approach of casting every problem as a text-to-text task constitutes an additional major contribution.
% This unifying framework differs from current practice and boasts both simplicity and strong performance.
We also explore the limits of current approaches by scaling up the insights from our systematic study (training models up to $11$ billion parameters) to obtain state-of-the-art results in many of the tasks we consider.
In order to perform experiments at this scale, we introduce the ``Colossal Clean Crawled Corpus'' (C4), a data set consisting of hundreds of gigabytes of clean English text scraped from the web.
Recognizing that the main utility of transfer learning is the possibility of leveraging pre-trained models in data-scarce settings, we release our code, data sets, and pre-trained models.\footnoteref{fn:oss}

The remainder of the paper is structured as follows:
In the following section, we discuss our base model and its implementation, our procedure for formulating every text processing problem as a text-to-text task, and the suite of tasks we consider.
In \cref{sec:experiments}, we present a large set of experiments that explore the field of transfer learning for NLP.
At the end of the section (\cref{sec:together}), we combine insights from our systematic study to obtain state-of-the-art results on a wide variety of benchmarks.
Finally, we provide a summary of our results and wrap up with a look towards the future in \cref{sec:conclusion}.

\section{Setup}
\label{sec:setup}

Before presenting the results from our large-scale empirical study, we review the necessary background topics required to understand our results, including the Transformer model architecture and the downstream tasks we evaluate on.
We also introduce our approach for treating every problem as a text-to-text task and describe our ``Colossal Clean Crawled Corpus'' (C4), the Common Crawl-based data set we created as a source of unlabeled text data.
We refer to our model and framework as the ``\textbf{T}ext-\textbf{t}o-\textbf{T}ext \textbf{T}ransfer \textbf{T}ransformer'' (T5).

\subsection{Model}
\label{sec:model}

Early results on transfer learning for NLP leveraged recurrent neural networks \citep{peters2018deep,howard2018universal}, but it has recently become more common to use models based on the ``Transformer'' architecture \citep{vaswani2017attention}.
The Transformer was initially shown to be effective for machine translation, but it has subsequently been used in a wide variety of NLP settings \citep{radford2018improving,devlin2018bert,mccann2018natural,yu2018qanet}.
Due to its increasing ubiquity, all of the models we study are based on the Transformer architecture.
Apart from the details mentioned below and the variants we explore in \cref{sec:architectures}, we do not deviate significantly from this architecture as originally proposed.
Instead of providing a comprehensive definition of this model, we refer the interested reader to the original paper \citep{vaswani2017attention} or follow-up tutorials\footnote{\url{http://nlp.seas.harvard.edu/2018/04/03/attention.html}}\textsuperscript{,}\footnote{\url{http://jalammar.github.io/illustrated-transformer/}} for a more detailed introduction.

The primary building block of the Transformer is self-attention \citep{cheng2016long}.
Self-attention is a variant of attention \citep{graves2013generating,bahdanau2014neural} that processes a sequence by replacing each element by a weighted average of the rest of the sequence.
The original Transformer consisted of an encoder-decoder architecture and was intended for sequence-to-sequence \citep{sutskever2014sequence,kalchbrenner2014convolutional} tasks.
It has recently also become common to use models consisting of a single Transformer layer stack, with varying forms of self-attention used to produce architectures appropriate for language modeling \citep{radford2018improving,al2019character} or classification and span prediction tasks \citep{devlin2018bert,yang2019xlnet}.
We empirically explore these architectural variants in \cref{sec:architectures}.

Overall, our encoder-decoder Transformer implementation closely follows its originally-proposed form \citep{vaswani2017attention}.
First, an input sequence of tokens is mapped to a sequence of embeddings, which is then passed into the encoder.
The encoder consists of a stack of ``blocks'', each of which comprises two subcomponents: a self-attention layer followed by a small feed-forward network.
Layer normalization \citep{ba2016layer} is applied to the input of each subcomponent.
We use a simplified version of layer normalization where the activations are only rescaled and no additive bias is applied.
After layer normalization, a residual skip connection \citep{he2016deep} adds each subcomponent's input to its output.
Dropout \citep{srivastava2014dropout} is applied within the feed-forward network, on the skip connection, on the attention weights, and at the input and output of the entire stack.
The decoder is similar in structure to the encoder except that it includes a standard attention mechanism after each self-attention layer that attends to the output of the encoder.
The self-attention mechanism in the decoder also uses a form of autoregressive or causal self-attention, which only allows the model to attend to past outputs.
The output of the final decoder block is fed into a dense layer with a softmax output, whose weights are shared with the input embedding matrix.
All attention mechanisms in the Transformer are split up into independent ``heads'' whose outputs are concatenated before being further processed.

Since self-attention is order-independent (i.e.\ it is an operation on sets), it is common to provide an explicit position signal to the Transformer.
While the original Transformer used a sinusoidal position signal or learned position embeddings, it has recently become more common to use relative position embeddings \citep{shaw2018self,huang2018music}.
Instead of using a fixed embedding for each position, relative position embeddings produce a different learned embedding according to the offset between the ``key'' and ``query'' being compared in the self-attention mechanism.
We use a simplified form of position embeddings where each ``embedding'' is simply a scalar that is added to the corresponding logit used for computing the attention weights.
For efficiency, we also share the position embedding parameters across all layers in our model, though within a given layer each attention head uses a different learned position embedding.
Typically, a fixed number of embeddings are learned, each corresponding to a range of possible key-query offsets.
In this work, we use $32$ embeddings for all of our models with ranges that increase in size logarithmically up to an offset of $128$ beyond which we assign all relative positions to the same embedding.
Note that a given layer is insensitive to relative position beyond $128$ tokens, but subsequent layers can build a sensitivity to larger offsets by combining local information from previous layers.
To summarize, our model is roughly equivalent to the original Transformer proposed by \citet{vaswani2017attention} with the exception of removing the Layer Norm bias, placing the layer normalization outside the residual path, and using a different position embedding scheme.
Since these architectural changes are orthogonal to the experimental factors we consider in our empirical survey of transfer learning, we leave the ablation of their impact for future work.

As part of our study, we experiment with the scalability of these models, i.e.\ how their performance changes as they are made to have more parameters or layers.
Training large models can be non-trivial since they might not fit on a single machine and require a great deal of computation.
As a result, we use a combination of model and data parallelism and train models on ``slices'' of Cloud TPU Pods.\footnote{\url{https://cloud.google.com/tpu/}}
TPU pods are are multi-rack ML supercomputers that contain $1{,}024$ TPU v3 chips connected via a high-speed 2D mesh interconnect with supporting CPU host machines.
We leverage the Mesh TensorFlow library \citep{shazeer2018mesh} for ease of implementation of both model parallelism and data parallelism \citep{krizhevsky2014one}.

\subsection{The Colossal Clean Crawled Corpus}
\label{sec:dataset}

Much of the previous work on transfer learning for NLP makes use of large unlabeled data sets for unsupervised learning.
In this paper, we are interested in measuring the effect of the quality, characteristics, and size of this unlabeled data.
To generate data sets that satisfy our needs, we leverage Common Crawl as a source of text scraped from the web.
Common Crawl has previously been used as a source of text data for NLP, for example to train an n-gram language model \citep{buck2014n}, as training data for commonsense reasoning \citep{trinh2018simple}, for mining parallel texts for machine translation \citep{smith2013dirt}, as a pre-training data set \citep{grave2018learning,zellers2019defending,liu2019roberta}, and even simply as a giant text corpus for testing optimizers \citep{anil2019memory}.

Common Crawl is a publicly-available web archive that provides ``web extracted text'' by removing markup and other non-text content from the scraped HTML files.
This process produces around 20TB of scraped text data each month.
Unfortunately, the majority of the resulting text is not natural language.
Instead, it largely comprises gibberish or boiler-plate text like menus, error messages, or duplicate text.
Furthermore, a good deal of the scraped text contains content that is unlikely to be helpful for any of the tasks we consider (offensive language, placeholder text, source code, etc.).
To address these issues, we used the following heuristics for cleaning up Common Crawl's web extracted text:
\begin{itemize}
  \item We only retained lines that ended in a terminal punctuation mark (i.e.\ a period, exclamation mark, question mark, or end quotation mark).
  \item We discarded any page with fewer than 3 sentences and only retained lines that contained at least 5 words.
  \item We removed any page that contained any word on the ``List of Dirty, Naughty, Obscene or Otherwise Bad Words''.\footnote{\url{https://github.com/LDNOOBW/List-of-Dirty-Naughty-Obscene-and-Otherwise-Bad-Words}}
  \item Many of the scraped pages contained warnings stating that Javascript should be enabled so we removed any line with the word Javascript.
  \item Some pages had placeholder ``lorem ipsum'' text; we removed any page where the phrase ``lorem ipsum'' appeared.
  \item Some pages inadvertently contained code. Since the curly bracket ``\{'' appears in many programming languages (such as Javascript, widely used on the web) but not in natural text, we removed any pages that contained a curly bracket.
  \item Since some of the scraped pages were sourced from Wikipedia and had citation markers (e.g.\ [1], [citation needed], etc.), we removed any such markers.
  \item Many pages had boilerplate policy notices, so we removed any lines containing the strings ``terms of use'', ``privacy policy'', ``cookie policy'', ``uses cookies'', ``use of cookies'', or ``use cookies''.
  \item To deduplicate the data set, we discarded all but one of any three-sentence span occurring more than once in the data set.
\end{itemize}
Additionally, since most of our downstream tasks are focused on English-language text, we used \texttt{langdetect}\footnote{\url{https://pypi.org/project/langdetect/}} to filter out any pages that were not classified as English with a probability of at least 0.99.
Our heuristics are inspired by past work on using Common Crawl as a source of data for NLP:
For example, \citet{grave2018learning} also filter text using an automatic language detector and discard short lines and \citet{smith2013dirt,grave2018learning} both perform line-level deduplication.
However, we opted to create a new data set because prior data sets use a more limited set of filtering heuristics, are not publicly available, and/or are different in scope (e.g.\ are limited to News data \citep{zellers2019defending,liu2019roberta}, comprise only Creative Commons content \citep{habernal2016c4corpus}, or are focused on parallel training data for machine translation \citep{smith2013dirt}).

To assemble our base data set, we downloaded the web extracted text from April 2019 and applied the aforementioned filtering.
This produces a collection of text that is not only orders of magnitude larger than most data sets used for pre-training (about 750 GB) but also comprises reasonably clean and natural English text.
We dub this data set the ``\textbf{C}olossal \textbf{C}lean \textbf{C}rawled \textbf{C}orpus'' (or C4 for short) and release it as part of TensorFlow Datasets.\footnote{\url{https://www.tensorflow.org/datasets/catalog/c4}}
We consider the impact of using various alternative versions of this data set in \cref{sec:datasets}.

\subsection{Downstream Tasks}
\label{sec:tasks}

Our goal in this paper is to measure general language learning abilities.
As such, we study downstream performance on a diverse set of benchmarks, including machine translation, question answering, abstractive summarization, and text classification.
Specifically, we measure performance on the GLUE and SuperGLUE text classification meta-benchmarks; CNN/Daily Mail abstractive summarization; SQuAD question answering; and WMT English to German, French, and Romanian translation.
All data was sourced from TensorFlow Datasets.\footnote{\url{https://www.tensorflow.org/datasets}}

GLUE \citep{wang2018glue} and SuperGLUE \citep{wang2019superglue} each comprise a collection of text classification tasks meant to test general language understanding abilities:
\begin{itemize}
  \item Sentence acceptability judgment (CoLA \citep{warstadt2018neural})
  \item Sentiment analysis (SST-2 \citep{socher2013recursive})
  \item Paraphrasing/sentence similarity (MRPC \citep{dolan2005automatically}, STS-B \citep{cer2017semeval}, QQP \citep{shankar2017first})
  \item Natural language inference (MNLI \citep{williams2017broad}, QNLI \citep{rajpurkar2016squad}, RTE \citep{dagan2005pascal}, CB \citep{de2019commitmentbank})
  \item Coreference resolution (WNLI and WSC \citep{levesque2012winograd})
  \item Sentence completion (COPA \citep{roemmele2011choice})
  \item Word sense disambiguation (WIC \citep{pilehvar2018wic})
  \item Question answering (MultiRC \citep{khashabi2018looking}, ReCoRD \citep{zhang2018record}, BoolQ \citep{clark2019boolq})
\end{itemize}
We use the data sets as distributed by the GLUE and SuperGLUE benchmarks.
For simplicity, when fine-tuning we treat all of the tasks in the GLUE benchmark (and similarly for SuperGLUE) as a single task by concatenating all of the constituent data sets.
As suggested by \cite{kocijan2019surprisingly} we also include the Definite Pronoun Resolution (DPR) data set \citep{rahman2012resolving} in the combined SuperGLUE task.

The CNN/Daily Mail \citep{hermann2015teaching} data set was introduced as a question-answering task but was adapted for text summarization by  \citet{nallapati2016abstractive}; we use the non-anonymized version from \citet{see2017get} as an abstractive summarization task.
SQuAD \citep{rajpurkar2016squad} is a common question-answering benchmark.
In our experiments, the model is fed the question and its context and asked to generate the answer token-by-token.
For WMT English to German, we use the same training data as \citep{vaswani2017attention} (i.e.\ News Commentary v13, Common Crawl, Europarl v7) and \texttt{newstest2013} as a validation set \citep{bojar2014findings}.
For English to French, we use the standard training data from 2015 and \texttt{newstest2014} as a validation set \citep{bojar2015findings}.
For English to Romanian, which is a standard lower-resource machine translation benchmark, we use the train and validation sets from WMT 2016 \citep{bojar2016findings}.
Note that we only pre-train on English data, so in order to learn to translate a given model will need to learn to generate text in a new language.

\subsection{Input and Output Format}
\label{sec:format}

In order to train a single model on the diverse set of tasks described above, we cast all of the tasks we consider into a ``text-to-text'' format---that is, a task where the model is fed some text for context or conditioning and is then asked to produce some output text.
This framework provides a consistent training objective both for pre-training and fine-tuning.
Specifically, the model is trained with a maximum likelihood objective (using ``teacher forcing'' \citep{williams1989learning}) regardless of the task.
To specify which task the model should perform, we add a task-specific (text) prefix to the original input sequence before feeding it to the model.

As an example, to ask the model to translate the sentence ``That is good.''\ from English to German, the model would be fed the sequence ``translate English to German: That is good.''\ and would be trained to output ``Das ist gut.''
For text classification tasks, the model simply predicts a single word corresponding to the target label.
For example, on the MNLI benchmark \citep{williams2017broad} the goal is to predict whether a premise implies (``entailment''), contradicts (``contradiction''), or neither (``neutral'') a hypothesis.
With our preprocessing, the input sequence becomes ``mnli premise: I hate pigeons. hypothesis: My feelings towards pigeons are filled with animosity.'' with the corresponding target word ``entailment''.
Note that an issue arises if our model outputs text on a text classification task that does not correspond to any of the possible labels (for example if the model outputs ``hamburger'' when the only possible labels for a task were ``entailment'', ``neutral'', or ``contradiction'').
In this case, we always count the model's output as wrong, though we never observed this behavior in any of our trained models.
Note that the choice of text prefix used for a given task is essentially a hyperparameter; we found that changing the exact wording of the prefix had limited impact and so did not perform extensive experiments into different prefix choices.
A diagram of our text-to-text framework with a few input/output examples is shown in \cref{fig:text_to_text}.
We provide full examples of preprocessed inputs for every task we studied in \cref{sec:preprocessing}.

Our text-to-text framework follows previous work that casts multiple NLP tasks into a common format:
\citet{mccann2018natural} propose the ``Natural Language Decathlon'', a benchmark that uses a consistent question-answering format for a suite of ten NLP tasks.
The Natural Language Decathlon also stipulates that all models must be multi-task, i.e.\ are able to simultaneously tackle all of the tasks at once.
We instead allow for separately fine-tuning the model on each individual task and use short task prefixes instead of an explicit question-answer format.
\citet{radford2019language} evaluate the zero-shot learning capabilities of language models by feeding some input to the model as a prefix and then autoregressively sampling an output.
For example, automatic summarization is done by feeding in a document followed by the text ``TL;DR:'' (short for ``too long, didn't read'', a common abbreviation) and then the summary is predicted via autoregressive decoding.
We mainly consider models that explicitly process an input with an encoder before generating an output with a separate decoder and we focus on transfer learning rather than zero-shot learning.
Finally, \citet{keskar2019unifying} unify many NLP tasks as ``span extraction'', where text corresponding to possible output choices are appended to the input and the model is trained to extract the input span corresponding to the correct choice.
In contrast, our framework also allows for generative tasks like machine translation and abstractive summarization where it is not possible to enumerate all possible output choices.

We were able to straightforwardly cast all of the tasks we considered into a text-to-text format with the exception of STS-B, which is a regression task where the goal is to predict a similarity score between $1$ and $5$.
We found that \text{most} of these scores were annotated in increments of $0.2$, so we simply rounded any score to the nearest increment of $0.2$ and converted the result to a literal string representation of the number (e.g.\ the floating-point value $2.57$ would be mapped to the string ``2.6'').
At test time, if the model outputs a string corresponding to a number between $1$ and $5$, we convert it to a floating-point value; otherwise, we treat the model's prediction as incorrect.
This effectively recasts the STS-B regression problem as a 21-class classification problem.

Separately, we also convert the Winograd tasks (WNLI from GLUE, WSC from SuperGLUE, and the DPR data set we add to SuperGLUE) into a simpler format that is more amenable to the text-to-text framework.
Examples from the Winograd tasks consist of a text passage containing an ambiguous pronoun that could refer to more than one of the noun phrases in the passage.
For example, the passage might be ``The city councilmen refused the demonstrators a permit because they feared violence.'', which contains the ambiguous pronoun ``they'' that could refer to ``city councilmen'' or ``demonstrators''.
We cast the WNLI, WSC, and DPR tasks as text-to-text problems by highlighting the ambiguous pronoun in the text passage and asking the model to predict the noun that it refers to.
The example mentioned above would be transformed to the input ``The city councilmen refused the demonstrators a permit because *they* feared violence.'' and the model would be trained to predict the target text ``The city councilmen''.

For WSC, examples contain the passage, the ambiguous pronoun, a candidate noun, and a True/False label reflecting whether the candidate matches the pronoun (ignoring any articles).
We only train on examples with a ``True'' label since we do not know the correct noun targets for examples with a ``False'' label.
For evaluation, we assign a ``True'' label if the words in the model's output are a subset of the words in the candidate noun phrase (or vice versa) and assign a ``False'' label otherwise.
This removes roughly half of the WSC training set, but the DPR data set adds about $1{,}000$ pronoun resolution examples.
Examples from DPR are annotated with the correct referent noun, making it easy to use this data set in the format listed above.

The WNLI training and validation sets have a significant overlap with the WSC training set.
To avoid leaking validation examples into our training data (a particular issue in the multi-task experiments of \cref{sec:multitask}), we therefore never train on WNLI and never report results on the WNLI validation set.
Omitting results on the WNLI validation set is standard practice \citep{devlin2018bert} due to the fact that it is ``adversarial'' with respect to the training set, i.e.\ validation examples are all slightly-perturbed versions of training examples with the opposite label.
As such, we do not include WNLI in the average GLUE score whenever we report on the validation set (all sections except \cref{sec:together} where results are presented on the test sets).
Converting examples from WNLI to the ``referent noun prediction'' variant described above is a little more involved; we describe this process in \cref{sec:wnli_preprocessing}.

\section{Experiments}
\label{sec:experiments}

Recent advances in transfer learning for NLP have come from a wide variety of developments, such as new pre-training objectives, model architectures, unlabeled data sets, and more.
In this section, we carry out an empirical survey of these techniques in hopes of teasing apart their contribution and significance.
We then combine the insights gained to attain state-of-the-art in many of the tasks we consider.
Since transfer learning for NLP is a rapidly growing area of research, it is not feasible for us to cover every possible technique or idea in our empirical study.
For a broader literature review, we recommend a recent survey by \cite{ruder2019transfer}.

We systematically study these contributions by taking a reasonable baseline (described in \cref{sec:baseline}) and altering one aspect of the setup at a time.
For example, in \cref{sec:objectives} we measure the performance of different unsupervised objectives while keeping the rest of our experimental pipeline fixed.
This ``coordinate ascent'' approach might miss second-order effects (for example, some particular unsupervised objective may work best on a model larger than our baseline setting), but performing a combinatorial exploration of all of the factors in our study would be prohibitively expensive.
In future work, we expect it could be fruitful to more thoroughly consider combinations of the approaches we study.

Our goal is to compare a variety of different approaches on a diverse set of tasks while keeping as many factors fixed as possible.
In order to satisfy this aim, in some cases we do not exactly replicate existing approaches.
For example, ``encoder-only'' models like BERT \citep{devlin2018bert} are designed to produce a single prediction per input token or a single prediction for an entire input sequence.
This makes them applicable for classification or span prediction tasks but not for generative tasks like translation or abstractive summarization.
As such, none of the model architectures we consider are identical to BERT or consist of an encoder-only structure.
Instead, we test approaches that are similar in spirit---for example, we consider an analogous objective to BERT's ``masked language modeling'' objective in \cref{sec:objectives} and we consider a model architecture that behaves similarly to BERT on text classification tasks in \cref{sec:architectures}.

After outlining our baseline experimental setup in the following subsection, we undertake an empirical comparison of model architectures (\cref{sec:architectures}), unsupervised objectives (\cref{sec:objectives}), pre-training data sets (\cref{sec:datasets}), transfer approaches (\cref{sec:transfer}), and scaling (\cref{sec:scaling}).
At the culmination of this section, we combine insights from our study with scale to obtain state-of-the-art results in many tasks we consider (\cref{sec:together}).

\subsection{Baseline}
\label{sec:baseline}

Our goal for our baseline is to reflect typical, modern practice.
We pre-train a standard Transformer (described in \cref{sec:model}) using a simple denoising objective and then separately fine-tune on each of our downstream tasks.
We describe the details of this experimental setup in the following subsections.

\subsubsection{Model}
\label{sec:model_hparams}

For our model, we use a standard encoder-decoder Transformer as proposed by \cite{vaswani2017attention}.
While many modern approaches to transfer learning for NLP use a Transformer architecture consisting of only a single ``stack'' (e.g.\ for language modeling \citep{radford2018improving,dong2019unified} or classification and span prediction \citep{devlin2018bert,yang2019xlnet}), we found that using a standard encoder-decoder structure achieved good results on both generative and classification tasks.
We explore the performance of different model architectures in \cref{sec:architectures}.

Our baseline model is designed so that the encoder and decoder are each similar in size and configuration to a ``BERT\textsubscript{BASE}'' \citep{devlin2018bert} stack.
Specifically, both the encoder and decoder consist of $12$ blocks (each block comprising self-attention, optional encoder-decoder attention, and a feed-forward network).
The feed-forward networks in each block consist of a dense layer with an output dimensionality of $d_{\mathrm{ff}} = 3072$ followed by a ReLU nonlinearity and another dense layer.
The ``key'' and ``value'' matrices of all attention mechanisms have an inner dimensionality of $d_{\mathrm{kv}} = 64$ and all attention mechanisms have $12$ heads.
All other sub-layers and embeddings have a dimensionality of $d_{\mathrm{model}} = 768$.
In total, this results in a model with about $220$ million parameters.
This is roughly twice the number of parameters of BERT\textsubscript{BASE} since our baseline model contains two layer stacks instead of one.
For regularization, we use a dropout probability of $0.1$ everywhere dropout is applied in the model.

\subsubsection{Training}
\label{sec:training}

As described in \cref{sec:format}, all tasks are formulated as text-to-text tasks.
This allows us to always train using standard maximum likelihood, i.e.\ using teacher forcing \citep{williams1989learning} and a cross-entropy loss.
For optimization, we use AdaFactor \citep{shazeer2018adafactor}.
At test time, we use greedy decoding (i.e.\ choosing the highest-probability logit at every timestep).

We pre-train each model for $2^{19} = 524{,}288$ steps on C4 before fine-tuning.
We use a maximum sequence length of $512$ and a batch size of $128$ sequences.
Whenever possible, we ``pack'' multiple sequences into each entry of the batch\footnote{\url{https://www.pydoc.io/pypi/tensor2tensor-1.5.7/autoapi/data_generators/generator_utils/index.html\#data_generators.generator_utils.pack_examples}} so that our batches contain roughly $2^{16} = 65{,}536$ tokens.
In total, this batch size and number of steps corresponds to pre-training on $2^{35} \approx 34\mathrm{B}$ tokens.
This is considerably less than BERT \citep{devlin2018bert}, which used roughly $137\mathrm{B}$ tokens, or RoBERTa \citep{liu2019roberta}, which used roughly $2.2\mathrm{T}$ tokens.
Using only $2^{35}$ tokens results in a reasonable computational budget while still providing a sufficient amount of pre-training for acceptable performance.
We consider the effect of pre-training for more steps in \cref{sec:scaling,sec:together}.
Note that $2^{35}$ tokens only covers a fraction of the entire C4 data set, so we never repeat any data during pre-training.

During pre-training, we use an ``inverse square root'' learning rate schedule: $1 \big/ \sqrt{\max(n, k)}$ where $n$ is the current training iteration and $k$ is the number of warm-up steps (set to $10^{4}$ in all of our experiments).
This sets a constant learning rate of $0.01$ for the first $10^{4}$ steps, then exponentially decays the learning rate until pre-training is over.
We also experimented with using a triangular learning rate \citep{howard2018universal}, which produced slightly better results but requires knowing the total number of training steps ahead of time.
Since we will be varying the number of training steps in some of our experiments, we opt for the more generic inverse square root schedule.

Our models are fine-tuned for $2^{18} = 262{,}144$ steps on all tasks.
This value was chosen as a trade-off between the high-resource tasks (i.e.\ those with large data sets), which benefit from additional fine-tuning, and low-resource tasks (smaller data sets), which overfit quickly.
During fine-tuning, we continue using batches with $128$ length-$512$ sequences (i.e.\ $2^{16}$ tokens per batch).
We use a constant learning rate of $0.001$ when fine-tuning.
We save a checkpoint every $5{,}000$ steps and report results on the model checkpoint corresponding to the highest validation performance.
For models fine-tuned on multiple tasks, we choose the best checkpoint for each task independently.
For all of the experiments except those in \cref{sec:together}, we report results in the validation set to avoid performing model selection on the test set.

\subsubsection{Vocabulary}
We use SentencePiece \citep{kudo2018sentencepiece} to encode text as WordPiece tokens \citep{sennrich2015neural,kudo2018subword}.
For all experiments, we use a vocabulary of $32{,}000$ wordpieces.
Since we ultimately fine-tune our model on English to German, French, and Romanian translation, we also require that our vocabulary covers these non-English languages.
To address this, we classified pages from the Common Crawl scrape used in C4 as German, French, and Romanian.
Then, we trained our SentencePiece model on a mixture of $10$ parts of English C4 data with $1$ part each of data classified as German, French or Romanian.
This vocabulary was shared across both the input and output of our model.
Note that our vocabulary makes it so that our model can only process a predetermined, fixed set of languages.

\subsubsection{Unsupervised Objective}
\label{sec:baseline_objective}

\begin{figure}[t]
    \centering
    \includegraphics[width=0.6\textwidth]{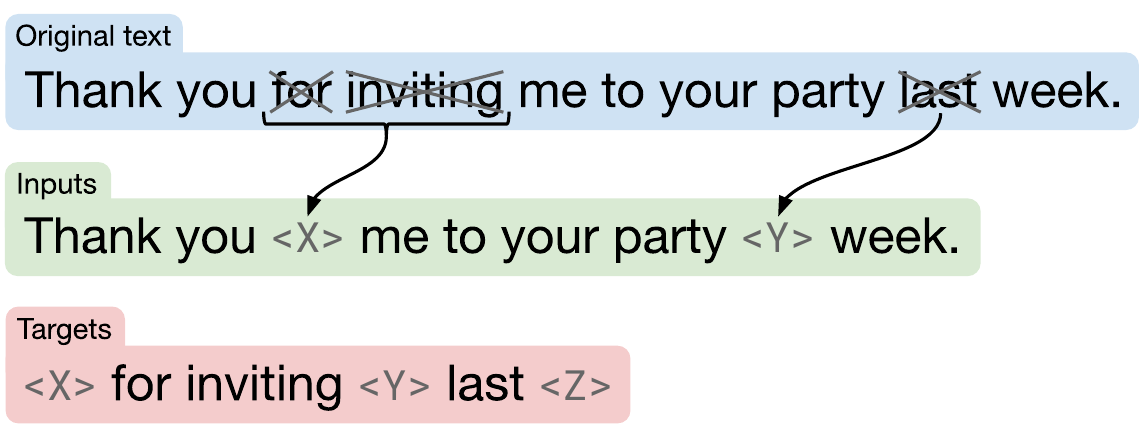}
    \caption{Schematic of the objective we use in our baseline model.
    In this example, we process the sentence ``Thank you for inviting me to your party last week.''
    The words ``for'', ``inviting'' and ``last'' (marked with an $\times$) are randomly chosen for corruption.
    Each consecutive span of corrupted tokens is replaced by a sentinel token (shown as \texttt{<X>} and \texttt{<Y>}) that is unique over the example.
    Since ``for'' and ``inviting'' occur consecutively, they are replaced by a single sentinel \texttt{<X>}.
    The output sequence then consists of the dropped-out spans, delimited by the sentinel tokens used to replace them in the input plus a final sentinel token \texttt{<Z>}.}
    \label{fig:objective}
\end{figure}

Leveraging unlabeled data to pre-train our model necessitates an objective that does not require labels but (loosely speaking) teaches the model generalizable knowledge that will be useful in downstream tasks.
Preliminary work that applied the transfer learning paradigm of pre-training and fine-tuning all of the model's parameters to NLP problems used a causal language modeling objective for pre-training \citep{dai2015semi,peters2018deep,radford2018improving,howard2018universal}.
However, it has recently been shown that ``denoising'' objectives \citep{devlin2018bert,taylor1953cloze} (also called ``masked language modeling'') produce better performance and as a result they have quickly become standard.
In a denoising objective, the model is trained to predict missing or otherwise corrupted tokens in the input.
Inspired by BERT's ``masked language modeling'' objective and the ``word dropout'' regularization technique \citep{bowman2015generating}, we design an objective that randomly samples and then drops out $15\%$ of tokens in the input sequence.
All consecutive spans of dropped-out tokens are replaced by a single sentinel token.
Each sentinel token is assigned a token ID that is unique to the sequence.
The sentinel IDs are special tokens which are added to our vocabulary and do not correspond to any wordpiece.
The target then corresponds to all of the dropped-out spans of tokens, delimited by the same sentinel tokens used in the input sequence plus a final sentinel token to mark the end of the target sequence.
Our choices to mask consecutive spans of tokens and only predict dropped-out tokens were made to reduce the computational cost of pre-training.
We perform thorough investigation into pre-training objectives in \cref{sec:objectives}.
An example of the transformation resulting from applying this objective is shown in \cref{fig:objective}.
We empirically compare this objective to many other variants in \cref{sec:objectives}.

\subsubsection{Baseline Performance}

In this section, we present results using the baseline experimental procedure described above to get a sense of what kind of performance to expect on our suite of downstream tasks.
Ideally, we would repeat every experiment in our study multiple times to get a confidence interval on our results.
Unfortunately, this would be prohibitively expensive due to the large number of experiments we run.
As a cheaper alternative, we train our baseline model $10$ times from scratch (i.e.\ with different random initializations and data set shuffling) and assume that the variance over these runs of the base model also applies to each experimental variant.
We don't expect most of the changes we make to have a dramatic effect on the inter-run variance, so this should provide a reasonable indication of the significance of different changes.
Separately, we also measure the performance of training our model for $2^{18}$ steps (the same number we use for fine-tuning) on all downstream tasks without pre-training.
This gives us an idea of how much pre-training benefits our model in the baseline setting.

When reporting results in the main text, we only report a subset of the scores across all the benchmarks to conserve space and ease interpretation.
For GLUE and SuperGLUE, we report the average score across all subtasks (as stipulated by the official benchmarks) under the headings ``GLUE'' and ``SGLUE''.
For all translation tasks, we report the BLEU score \citep{papineni2002bleu} as provided by SacreBLEU v1.3.0 \citep{post2018call} with ``exp'' smoothing and ``intl'' tokenization.
We refer to scores for WMT English to German, English to French, and English to Romanian as EnDe, EnFr, and EnRo, respectively.
For CNN/Daily Mail, we find the performance of models on the ROUGE-1-F, ROUGE-2-F, and ROUGE-L-F metrics \citep{lin2004rouge} to be highly correlated so we report the ROUGE-2-F score alone under the heading ``CNNDM''.
Similarly, for SQuAD we find the performance of the ``exact match'' and ``F1'' scores to be highly correlated so we report the ``exact match'' score alone.
We provide every score achieved on every task for all experiments in \cref{tab:giant}, \cref{sec:giant}.

Our results tables are all formatted so that each row corresponds to a particular experimental configuration with columns giving the scores for each benchmark.
We will include the mean performance of the baseline configuration in most tables.
Wherever a baseline configuration appears, we will mark it with a $\bigstar$ (as in the first row of \cref{tab:baseline}).
We also will \textbf{boldface} any score that is within two standard deviations of the maximum (best) in a given experiment.

\begin{table}
\footnotesize
\centering
\begin{tabular}{l c c c c c c c}
\toprule
                                & GLUE        & CNNDM       & SQuAD       & SGLUE       & EnDe        & EnFr        & EnRo    \\
\midrule
    \bsl Baseline average       & $\*{83.28}$ & $\*{19.24}$ & $\*{80.88}$ & $\*{71.36}$ & $\*{26.98}$ & $\*{39.82}$ & $\*{27.65}$ \\
    Baseline standard deviation & $0.235$     & $0.065$     & $0.343$     & $0.416$     & $0.112$     & $0.090$     & $0.108$ \\
    No pre-training             & $66.22$     & $17.60$     & $50.31$     & $53.04$     & $25.86$     & $\*{39.77}$ & $24.04$ \\
\bottomrule
\end{tabular}
\caption{
Average and standard deviation of scores achieved by our baseline model and training procedure.
For comparison, we also report performance when training on each task from scratch (i.e.\ without any pre-training) for the same number of steps used to fine-tune the baseline model.
All scores in this table (and every table in our paper except \cref{tab:final}) are reported on the validation sets of each data set.
}
\label{tab:baseline}
\end{table}

Our baseline results are shown in \cref{tab:baseline}.
Overall, our results are comparable to existing models of similar size.
For example, BERT\textsubscript{BASE} achieved an exact match score of $80.8$ on SQuAD and an accuracy of $84.4$ on MNLI-matched, whereas we achieve $80.88$ and $84.24$, respectively (see \cref{tab:giant}).
Note that we cannot directly compare our baseline to BERT\textsubscript{BASE} because ours is an encoder-decoder model and was pre-trained for roughly \sfrac{1}{4} as many steps.
Unsurprisingly, we find that pre-training provides significant gains across almost all benchmarks.
The only exception is WMT English to French, which is a large enough data set that gains from pre-training tend to be marginal.
We include this task in our experiments to test the behavior of transfer learning in the high-resource regime.
Since we perform early stopping by selecting the best-performing checkpoint, the large disparity between our baseline and ``no pre-training'' emphasize how much pre-training improves performance on tasks with limited data.
While we do not explicitly measure improvements in data efficiency in this paper, we emphasize that this is one of the primary benefits of the transfer learning paradigm.

As for inter-run variance, we find that for most tasks the standard deviation across runs is smaller than $1\%$ of the task's baseline score.
Exceptions to this rule include CoLA, CB, and COPA, which are all low-resource tasks from the GLUE and SuperGLUE benchmarks.
For example, on CB our baseline model had an average F1 score of $91.22$ with a standard deviation of $3.237$ (see \cref{tab:giant}), which may be partly due to the fact that CB's validation set contains only $56$ examples.
Note that the GLUE and SuperGLUE scores are computed as the average of scores across the tasks comprising each benchmark.
As a result, we caution that the high inter-run variance of CoLA, CB, and COPA can make it harder to compare models using the GLUE and SuperGLUE scores alone.

\subsection{Architectures}
\label{sec:architectures}

While the Transformer was originally introduced with an encoder-decoder architecture, much modern work on transfer learning for NLP uses alternative architectures.
In this section, we review and compare these architectural variants.

\subsubsection{Model Structures}
\label{sec:structures}

\begin{figure}[t]
    \centering
    \includegraphics[width=0.8\textwidth]{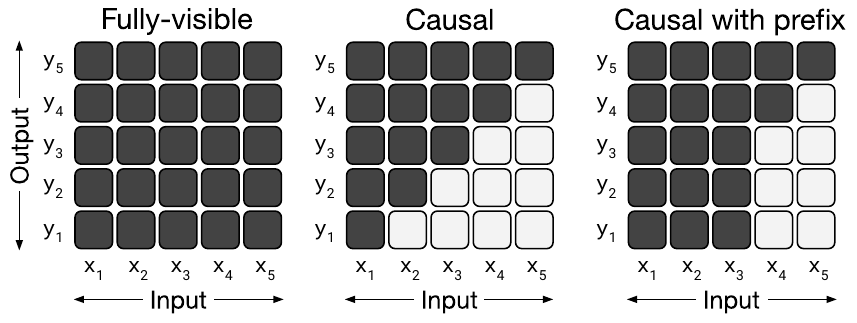}
    \caption{
    Matrices representing different attention mask patterns.
    The input and output of the self-attention mechanism are denoted $x$ and $y$ respectively.
    A dark cell at row $i$ and column $j$ indicates that the self-attention mechanism is allowed to attend to input element $j$ at output timestep $i$.
    A light cell indicates that the self-attention mechanism is \textit{not} allowed to attend to the corresponding $i$ and $j$ combination.
    Left: A fully-visible mask allows the self-attention mechanism to attend to the full input at every output timestep.
    Middle: A causal mask prevents the $i$th output element from depending on any input elements from ``the future''.
    Right: Causal masking with a prefix allows the self-attention mechanism to use fully-visible masking on a portion of the input sequence.
    }
    \label{fig:attention_masks}
\end{figure}

A major distinguishing factor for different architectures is the ``mask'' used by different attention mechanisms in the model.
Recall that the self-attention operation in a Transformer takes a sequence as input and outputs a new sequence of the same length.
Each entry of the output sequence is produced by computing a weighted average of entries of the input sequence.
Specifically, let $y_i$ refer to the $i$th element of the output sequence and $x_j$ refer to the $j$th entry of the input sequence.
$y_i$ is computed as $\sum_j w_{i, j} x_j$, where $w_{i, j}$ is the scalar weight produced by the self-attention mechanism as a function of $x_i$ and $x_j$.
The attention mask is then used to zero out certain weights in order to constrain which entries of the input can be attended to at a given output timestep.
Diagrams of the masks we will consider are shown in \cref{fig:attention_masks}.
For example, the causal mask (\cref{fig:attention_masks}, middle) sets any $w_{i, j}$ to zero if $j > i$.

\begin{figure}[t]
    \centering
    \includegraphics[width=0.8\textwidth]{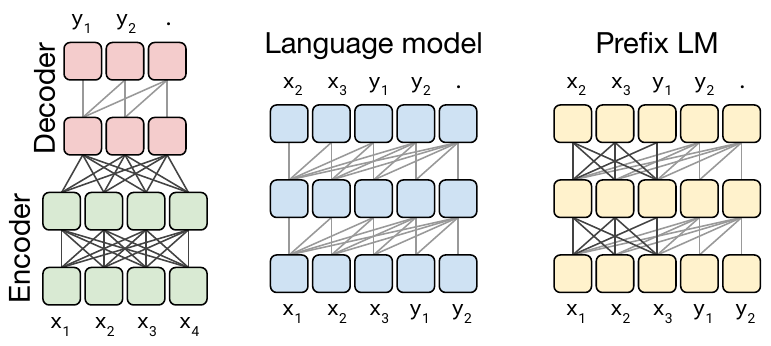}
    \caption{
    Schematics of the Transformer architecture variants we consider.
    In this diagram, blocks represent elements of a sequence and lines represent attention visibility.
    Different colored groups of blocks indicate different Transformer layer stacks.
    Dark grey lines correspond to fully-visible masking and light grey lines correspond to causal masking.
    We use ``.''\ to denote a special end-of-sequence token that represents the end of a prediction.
    The input and output sequences are represented as $x$ and $y$ respectively.
    Left: A standard encoder-decoder architecture uses fully-visible masking in the encoder and the encoder-decoder attention, with causal masking in the decoder.
    Middle: A language model consists of a single Transformer layer stack and is fed the concatenation of the input and target, using a causal mask throughout.
    Right: Adding a prefix to a language model corresponds to allowing fully-visible masking over the input.
    }
    \label{fig:architectures}
\end{figure}

The first model structure we consider is an an encoder-decoder Transformer, which consists of two layer stacks: The encoder, which is fed an input sequence, and the decoder, which produces a new output sequence.
A schematic of this architectural variant is shown in the left panel of \cref{fig:architectures}.

The encoder uses a ``fully-visible'' attention mask.
Fully-visible masking allows a self-attention mechanism to attend to any entry of the input when producing each entry of its output.
We visualize this masking pattern in \cref{fig:attention_masks}, left.
This form of masking is appropriate when attending over a ``prefix'', i.e.\ some context provided to the model that is later used when making predictions.
BERT \citep{devlin2018bert} also uses a fully-visible masking pattern and appends a special ``classification'' token to the input.
BERT's output at the timestep corresponding to the classification token is then used to make a prediction for classifying the input sequence.

The self-attention operations in the Transformer's decoder use a ``causal'' masking pattern.
When producing the $i$th entry of the output sequence, causal masking prevents the model from attending to the $j$th entry of the input sequence for $j > i$.
This is used during training so that the model can't ``see into the future'' as it produces its output.
An attention matrix for this masking pattern is shown in \cref{fig:attention_masks}, middle.

The decoder in an encoder-decoder Transformer is used to autoregressively produce an output sequence.
That is, at each output timestep, a token is sampled from the model's predicted distribution and the sample is fed back into the model to produce a prediction for the next output timestep, and so on.
As such, a Transformer decoder (without an encoder) can be used as a language model (LM), i.e.\ a model trained solely for next-step prediction \citep{liu2018generating,radford2018improving,al2019character}.
This constitutes the second model structure we consider.
A schematic of this architecture is shown in \cref{fig:architectures}, middle.
In fact, early work on transfer learning for NLP used this architecture with a language modeling objective as a pre-training method \citep{radford2018improving}.

Language models are typically used for compression or sequence generation \citep{graves2013generating}.
However, they can also be used in the text-to-text framework simply by concatenating the inputs and targets.
As an example, consider the case of English to German translation: If we have a training datapoint with input sentence ``That is good.'' and target ``Das ist gut.'', we would simply train the model on next-step prediction over the concatenated input sequence ``translate English to German: That is good. target: Das ist gut.''
If we wanted to obtain the model's prediction for this example, the model would be fed the prefix ``translate English to German: That is good. target:'' and would be asked to generate the remainder of the sequence autoregressively.
In this way, the model can predict an output sequence given an input, which satisfies the needs of text-to-text tasks.
This approach was recently used to show that language models can learn to perform some text-to-text tasks without supervision \citep{radford2019language}.

A fundamental and frequently cited drawback of using a language model in the text-to-text setting is that causal masking forces the model's representation of the $i$th entry of the input sequence to only depend on the entries up until $i$.
To see why this is potentially disadvantageous, consider the text-to-text framework where the model is provided with a prefix/context before being asked to make predictions (e.g., the prefix is an English sentence and the model is asked to predict the German translation).
With fully causal masking, the model's representation of a prefix state can only depend on prior entries of the prefix.
So, when predicting an entry of the output, the model will attend to a representation of the prefix that is unnecessarily limited.
Similar arguments have been made against using a unidirectional recurrent neural network encoder in sequence-to-sequence models \citep{bahdanau2014neural}.

This issue can be avoided in a Transformer-based language model simply by changing the masking pattern.
Instead of using a causal mask, we use fully-visible masking during the prefix portion of the sequence.
This masking pattern and a schematic of the resulting ``prefix LM'' (the third model structure we consider) are illustrated in the rightmost panels of \cref{fig:attention_masks,fig:architectures}, respectively.
In the English to German translation example mentioned above, fully-visible masking would be applied to the prefix ``translate English to German: That is good. target:'' and causal masking would be used during training for predicting the target ``Das ist gut.''
Using a prefix LM in the text-to-text framework was originally proposed by \cite{liu2018generating}.
More recently, \cite{dong2019unified} showed that this architecture is effective on a wide variety of text-to-text tasks.
This architecture is similar to an encoder-decoder model with parameters shared across the encoder and decoder and with the encoder-decoder attention replaced with full attention across the input and target sequence.

We note that when following our text-to-text framework, the prefix LM architecture closely resembles BERT \citep{devlin2018bert} for classification tasks.
To see why, consider an example from the MNLI benchmark where the premise is ``I hate pigeons.'', the hypothesis is ``My feelings towards pigeons are filled with animosity.'' and the correct label is ``entailment''.
To feed this example into a language model, we would transform it into the sequence ``mnli premise: I hate pigeons. hypothesis: My feelings towards pigeons are filled with animosity. target: entailment''.
In this case, the fully-visible prefix would correspond to the entire input sequence up to the word ``target:'', which can be seen as being analogous to the ``classification'' token used in BERT.
So, our model would have full visibility over the entire input, and then would be tasked with making a classification by outputting the word ``entailment''.
It is easy for the model to learn to output one of the valid class labels given the task prefix (``mnli'' in this case).
As such, the main difference between a prefix LM and the BERT architecture is that the classifier is simply integrated into the output layer of the Transformer decoder in the prefix LM.

\subsubsection{Comparing Different Model Structures}
\label{sec:architecture_variants}

In the interest of experimentally comparing these architectural variants, we would like each model we consider to be equivalent in some meaningful way.
We might say that two models are equivalent if they either have the same number of parameters or they require roughly the same amount of computation to process a given (input-sequence, target-sequence) pair.
Unfortunately, it is not possible to compare an encoder-decoder model to a language model architecture (comprising a single Transformer stack) according to both of these criteria at the same time.
To see why, first note an encoder-decoder model with $L$ layers in the encoder and $L$ layers in the decoder has approximately the same number of parameters as a language model with $2L$ layers.
However, the same $L + L$ encoder-decoder model will have approximately the same computational cost as a language model with \textit{only} $L$ layers.
This is a consequence of the fact that the $L$ layers in the language model must be applied to \textit{both} the input and output sequence, while the encoder is only applied to the input sequence and the decoder is only applied to the output sequence.
Note that these equivalences are approximate---there are some extra parameters in the decoder due to the encoder-decoder attention and there are also some computational costs in the attention layers that are quadratic in the sequence lengths.
In practice, however, we observed nearly identical step times for $L$-layer language models versus $L + L$-layer encoder-decoder models, suggesting a roughly equivalent computational cost.
Further, for the model sizes we consider, the number of parameters in the encoder-decoder attention layers is about 10\% of the total parameter count, so we make the simplifying assumption that an $L + L$-layer encoder-decoder model has the same number of parameters as an $2L$-layer language model.

To provide a reasonable means of comparison, we consider multiple configurations for our encoder-decoder model.
We will refer to the number of layers and parameters in a BERT\textsubscript{BASE}-sized layer stack as $L$ and $P$, respectively.
We will use $M$ to refer to the number of FLOPs required for an $L + L$-layer encoder-decoder model or $L$-layer decoder-only model to process a given input-target pair.
In total, we will compare:
\begin{itemize}
    \item An encoder-decoder model with $L$ layers in the encoder and $L$ layers in the decoder. This model has $2P$ parameters and a computation cost of $M$ FLOPs.
    \item An equivalent model, but with parameters shared across the encoder and decoder, resulting in $P$ parameters and an $M$-FLOP computational cost.
    \item An encoder-decoder model with $L/2$ layers each in the encoder and decoder, giving $P$ parameters and an $M/2$-FLOP cost.
    \item A decoder-only language model with $L$ layers and $P$ parameters and a resulting computational cost of $M$ FLOPs.
    \item A decoder-only prefix LM with the same architecture (and thus the same number of parameters and computational cost), but with fully-visible self-attention over the input.
\end{itemize}

\subsubsection{Objectives}
\label{sec:architecture_objectives}

As an unsupervised objective, we will consider both a basic language modeling objective as well as our baseline denoising objective described in \cref{sec:baseline_objective}.
We include the language modeling objective due to its historic use as a pre-training objective \citep{dai2015semi,ramachandran2016unsupervised,howard2018universal,radford2018improving,peters2018deep} as well as its natural fit for the language model architectures we consider.
For models that ingest a prefix before making predictions (the encoder-decoder model and prefix LM), we sample a span of text from our unlabeled data set and choose a random point to split it into prefix and target portions.
For the standard language model, we train the model to predict the entire span from beginning to end.
Our unsupervised denoising objective is designed for text-to-text models; to adapt it for use with a language model we concatenate the inputs and targets as described in \cref{sec:structures}.

\subsubsection{Results}

\begin{table}
\footnotesize
\begin{widepage}
\centering
\begin{tabular}{l c c c c c c c c c c}
\toprule
    Architecture         & Objective & Params & Cost  & GLUE        & CNNDM       & SQuAD       & SGLUE       & EnDe        & EnFr        & EnRo    \\
\midrule
    \bsl Encoder-decoder & Denoising & $2P$   & $M$   & $\*{83.28}$ & $\*{19.24}$ & $\*{80.88}$ & $\*{71.36}$ & $\*{26.98}$ & $\*{39.82}$ & $\*{27.65}$ \\
    Enc-dec, shared      & Denoising & $P$    & $M$   & $82.81$     & $18.78$     & $\*{80.63}$ & $\*{70.73}$ & $26.72$     & $39.03$     & $\*{27.46}$ \\
    Enc-dec, 6 layers    & Denoising & $P$    & $M/2$ & $80.88$     & $18.97$     & $77.59$     & $68.42$     & $26.38$     & $38.40$     & $26.95$ \\
    Language model       & Denoising & $P$    & $M$   & $74.70$     & $17.93$     & $61.14$     & $55.02$     & $25.09$     & $35.28$     & $25.86$ \\
    Prefix LM            & Denoising & $P$    & $M$   & $81.82$     & $18.61$     & $78.94$     & $68.11$     & $26.43$     & $37.98$     & $27.39$ \\
\midrule
    Encoder-decoder      & LM        & $2P$   & $M$   & $79.56$     & $18.59$     & $76.02$     & $64.29$     & $26.27$     & $39.17$     & $26.86$ \\
    Enc-dec, shared      & LM        & $P$    & $M$   & $79.60$     & $18.13$     & $76.35$     & $63.50$     & $26.62$     & $39.17$     & $27.05$ \\
    Enc-dec, 6 layers    & LM        & $P$    & $M/2$ & $78.67$     & $18.26$     & $75.32$     & $64.06$     & $26.13$     & $38.42$     & $26.89$ \\
    Language model       & LM        & $P$    & $M$   & $73.78$     & $17.54$     & $53.81$     & $56.51$     & $25.23$     & $34.31$     & $25.38$ \\
    Prefix LM            & LM        & $P$    & $M$   & $79.68$     & $17.84$     & $76.87$     & $64.86$     & $26.28$     & $37.51$     & $26.76$ \\
\bottomrule
\end{tabular}
\end{widepage}
\caption{
Performance of the different architectural variants described in \cref{sec:architecture_variants}.
We use $P$ to refer to the number of parameters in a 12-layer base Transformer layer stack and $M$ to refer to the FLOPs required to process a sequence using the encoder-decoder model.
We evaluate each architectural variant using a denoising objective (described in \cref{sec:baseline_objective}) and an autoregressive objective (as is commonly used to train language models).
}
\label{tab:architectures_results}

\end{table}

The scores achieved by each of the architectures we compare are shown in \cref{tab:architectures_results}.
For all tasks, the encoder-decoder architecture with the denoising objective performed best.
This variant has the highest parameter count ($2P$) but the same computational cost as the $P$-parameter decoder-only models.
Surprisingly, we found that sharing parameters across the encoder and decoder performed nearly as well.
In contrast, halving the number of layers in the encoder and decoder stacks significantly hurt performance.
Concurrent work \citep{lan2019albert} also found that sharing parameters across Transformer blocks can be an effective means of lowering the total parameter count without sacrificing much performance.
XLNet also bears some resemblance to the shared encoder-decoder approach with a denoising objective \citep{yang2019xlnet}.
We also note that the shared parameter encoder-decoder outperforms the decoder-only prefix LM, suggesting that the addition of an explicit encoder-decoder attention is beneficial.
Finally, we confirm the widely-held conception that using a denoising objective always results in better downstream task performance compared to a language modeling objective.
This observation has been previously made by \citet{devlin2018bert}, \citet{voita2019bottom}, and \citet{lample2019cross} among others.
We undertake a more detailed exploration of unsupervised objectives in the following section.

\subsection{Unsupervised Objectives}
\label{sec:objectives}

The choice of unsupervised objective is of central importance as it provides the mechanism through which the model gains general-purpose knowledge to apply to downstream tasks.
This has led to the development of a wide variety of pre-training objectives \citep{dai2015semi,ramachandran2016unsupervised,radford2018improving,devlin2018bert,yang2019xlnet,liu2019multi,wang2019can,song2019mass,dong2019unified,joshi2019spanbert}.
In this section, we perform a procedural exploration of the space of unsupervised objectives.
In many cases, we will not replicate an existing objective exactly---some will be modified to fit our text-to-text encoder-decoder framework and, in other cases, we will use objectives that combine concepts from multiple common approaches.

Overall, all of our objectives ingest a sequence of token IDs corresponding to a tokenized span of text from our unlabeled text data set.
The token sequence is processed to produce a (corrupted) input sequence and a corresponding target.
Then, the model is trained as usual with maximum likelihood to predict the target sequence.
We provide illustrative examples of many of the objectives we consider in \cref{tab:objectives}.

\begin{table}
\footnotesize
\begin{widepage}
\centering
\begin{tabular}{l l l}
\toprule
    Objective & Inputs & Targets   \\
\midrule
    Prefix language modeling & Thank you for inviting & me to your party last week . \\
    BERT-style \cite{devlin2018bert} & Thank you \texttt{<M>} \texttt{<M>} me to your party \textcolor{gray}{apple} week . & \textit{(original text)}  \\
    Deshuffling & party me for your to . last fun you inviting week Thank & \textit{(original text)} \\
    MASS-style \cite{song2019mass} & Thank you \texttt{<M>} \texttt{<M>} me to your party \texttt{<M>} week . & \textit{(original text)} \\
    I.i.d.\ noise, replace spans & Thank you \texttt{<X>} me to your party \texttt{<Y>} week . & \texttt{<X>} for inviting \texttt{<Y>} last \texttt{<Z>} \\
    I.i.d.\ noise, drop tokens & Thank you me to your party week . & for inviting last \\
    Random spans & Thank you \texttt{<X>} to \texttt{<Y>} week . & \texttt{<X>} for inviting me \texttt{<Y>} your party last \texttt{<Z>} \\
\bottomrule
\end{tabular}
\end{widepage}
\caption{
Examples of inputs and targets produced by some of the unsupervised objectives we consider applied to the input text ``Thank you for inviting me to your party last week .''
Note that all of our objectives process \textit{tokenized} text. For this particular sentence, all words were mapped to a single token by our vocabulary.
We write \textit{(original text)} as a target to denote that the model is tasked with reconstructing the entire input text.
\texttt{<M>} denotes a shared mask token and \texttt{<X>}, \texttt{<Y>}, and \texttt{<Z>} denote sentinel tokens that are assigned unique token IDs.
The BERT-style objective (second row) includes a corruption where some tokens are replaced by a random token ID; we show this via the greyed-out word \textcolor{gray}{apple}.
}
\label{tab:objectives}
\end{table}

\subsubsection{Disparate High-Level Approaches}
\label{sec:objectives_highlevel}

To begin with, we compare three techniques that are inspired by commonly-used objectives but differ significantly in their approach.
First, we include a basic ``prefix language modeling'' objective as was used in \cref{sec:architecture_objectives}.
This technique splits a span of text into two components, one to use as inputs to the encoder and the other to use as a target sequence to be predicted by the decoder.
Second, we consider an objective inspired by the ``masked language modeling'' (MLM) objective used in BERT \citep{devlin2018bert}.
MLM takes a span of text and corrupts $15\%$ of the tokens.
$90\%$ of the corrupted tokens are replaced with a special mask token and $10\%$ are replaced with a random token.
Since BERT is an encoder-only model, its goal during pre-training is to reconstruct masked tokens at the output of the encoder.
In the encoder-decoder case, we simply use the entire uncorrupted sequence as the target.
Note that this differs from our baseline objective, which uses only the corrupted tokens as targets; we compare these two approaches in \cref{sec:simplifying_bert}.
Finally, we also consider a basic deshuffling objective as used e.g.\ in \citep{liu2019summae} where it was applied to a denoising sequential autoencoder.
This approach takes a sequence of tokens, shuffles it, and then uses the original deshuffled sequence as a target.
We provide examples of the inputs and targets for these three methods in the first three rows of \cref{tab:objectives}.

\begin{table}
\footnotesize
\begin{widepage}
\centering
\begin{tabular}{l c c c c c c c c c c}
\toprule
    Objective                         & GLUE        & CNNDM       & SQuAD       & SGLUE       & EnDe        & EnFr        & EnRo        \\
\midrule
    Prefix language modeling          & $80.69$     & $18.94$     & $77.99$     & $65.27$     & $\*{26.86}$ & $39.73$     & $\*{27.49}$ \\
    BERT-style \citep{devlin2018bert} & $\*{82.96}$ & $\*{19.17}$ & $\*{80.65}$ & $\*{69.85}$ & $\*{26.78}$ & $\*{40.03}$ & $\*{27.41}$ \\
    Deshuffling                       & $73.17$     & $18.59$     & $67.61$     & $58.47$     & $26.11$     & $39.30$     & $25.62$ \\
\bottomrule
\end{tabular}
\end{widepage}
\caption{
Performance of the three disparate pre-training objectives described in \cref{sec:objectives_highlevel}.
}
\label{tab:objectives_highlevel}
\end{table}

The performance of these three objectives is shown in \cref{tab:objectives_highlevel}.
Overall, we find that the BERT-style objective performs best, though the prefix language modeling objective attains similar performance on the translation tasks.
Indeed, the motivation for the BERT objective was to outperform language model-based pre-training.
The deshuffling objective performs considerably worse than both prefix language modeling and the BERT-style objective.

\subsubsection{Simplifying the BERT Objective}
\label{sec:simplifying_bert}

Based on the results in the prior section, we will now focus on exploring modifications to the BERT-style denoising objective.
This objective was originally proposed as a pre-training technique for an encoder-only model trained for classification and span prediction.
As such, it may be possible to modify it so that it performs better or is more efficient in our encoder-decoder text-to-text setup.

First, we consider a simple variant of the BERT-style objective where we don't include the random token swapping step.
The resulting objective simply replaces $15\%$ of the tokens in the input with a mask token and the model is trained to reconstruct the original uncorrupted sequence.
A similar masking objective was used by \cite{song2019mass} where it was referred to as ``MASS'', so we call this variant the ``MASS-style'' objective.
Second, we were interested to see if it was possible to avoid predicting the entire uncorrupted text span since this requires self-attention over long sequences in the decoder.
We consider two strategies to achieve this:
First, instead of replacing each corrupted token with a mask token, we replace the entirety of each consecutive span of corrupted tokens with a unique mask token.
Then, the target sequence becomes the concatenation of the ``corrupted'' spans, each prefixed by the mask token used to replace it in the input.
This is the pre-training objective we use in our baseline, described in \cref{sec:baseline_objective}.
Second, we also consider a variant where we simply drop the corrupted tokens from the input sequence completely and task the model with reconstructing the dropped tokens in order.
Examples of these approaches are shown in the fifth and sixth rows of \cref{tab:objectives}.

\begin{table}
\footnotesize
\begin{widepage}
\centering
\begin{tabular}{l c c c c c c c c c c}
\toprule
    Objective                         & GLUE        & CNNDM       & SQuAD       & SGLUE       & EnDe        & EnFr        & EnRo    \\
\midrule
    BERT-style \citep{devlin2018bert} & $82.96$     & $19.17$     & $\*{80.65}$ & $69.85$     & $26.78$     & $\*{40.03}$ & $27.41$ \\
    MASS-style \citep{song2019mass}   & $82.32$     & $19.16$     & $80.10$     & $69.28$     & $26.79$     & $\*{39.89}$ & $27.55$ \\
    \bsl Replace corrupted spans      & $83.28$     & $\*{19.24}$ & $\*{80.88}$ & $\*{71.36}$ & $\*{26.98}$ & $39.82$     & $\*{27.65}$ \\
    Drop corrupted tokens             & $\*{84.44}$ & $\*{19.31}$ & $\*{80.52}$ & $68.67$     & $\*{27.07}$ & $39.76$     & $\*{27.82}$ \\
\bottomrule
\end{tabular}
\end{widepage}
\caption{
Comparison of variants of the BERT-style pre-training objective.
In the first two variants, the model is trained to reconstruct the original uncorrupted text segment.
In the latter two, the model only predicts the sequence of corrupted tokens.
}
\label{tab:objectives_bert}
\end{table}

An empirical comparison of the original BERT-style objective to these three alternatives is shown in \cref{tab:objectives_bert}.
We find that in our setting, all of these variants perform similarly.
The only exception was that dropping corrupted tokens completely produced a small improvement in the GLUE score thanks to a significantly higher score on CoLA ($60.04$, compared to our baseline average of $53.84$, see \cref{tab:giant}).
This may be due to the fact that CoLA involves classifying whether a given sentence is grammatically and syntactically acceptable, and being able to determine when tokens are missing is closely related to detecting acceptability.
However, dropping tokens completely performed worse than replacing them with sentinel tokens on SuperGLUE.
The two variants that do not require predicting the full original sequence (``replace corrupted spans'' and ``drop corrupted spans'') are both potentially attractive since they make the target sequences shorter and consequently make training faster.
Going forward, we will explore variants where we replace corrupted spans with sentinel tokens and only predict the corrupted tokens (as in our baseline objective).

\subsubsection{Varying the Corruption Rate}
\label{sec:corruption_rate}

So far, we have been corrupting 15\% of the tokens, the value used in BERT \citep{devlin2018bert}.
Again, since our text-to-text framework differs from BERT's, we are interested to see if a different corruption rate works better for us.
We compare corruption rates of $10\%$, $15\%$, $25\%$, and $50\%$ in \cref{tab:objectives_rate}.
Overall, we find that the corruption rate had a limited effect on the model's performance.
The only exception is that the largest corruption rate we consider ($50\%$) results in a significant degradation of performance on GLUE and SQuAD.
Using a larger corruption rate also results in longer targets, which can potentially slow down training.
Based on these results and the historical precedent set by BERT, we will use a corruption rate of $15\%$ going forward.

\begin{table}
\footnotesize
\begin{widepage}
\centering
\begin{tabular}{l c c c c c c c c c c}
\toprule
    Corruption rate & GLUE        & CNNDM       & SQuAD       & SGLUE       & EnDe        & EnFr        & EnRo    \\
\midrule
    $10\%$          & $\*{82.82}$ & $19.00$     & $\*{80.38}$ & $69.55$     & $\*{26.87}$ & $39.28$     & $\*{27.44}$ \\
    \bsl $15\%$     & $\*{83.28}$ & $19.24$     & $\*{80.88}$ & $\*{71.36}$ & $\*{26.98}$ & $\*{39.82}$ & $\*{27.65}$ \\
    $25\%$          & $\*{83.00}$ & $\*{19.54}$ & $\*{80.96}$ & $70.48$     & $\*{27.04}$ & $\*{39.83}$ & $\*{27.47}$ \\
    $50\%$          & $81.27$     & $19.32$     & $79.80$     & $70.33$     & $\*{27.01}$ & $\*{39.90}$ & $\*{27.49}$ \\
\bottomrule
\end{tabular}
\end{widepage}
\caption{
Performance of the i.i.d.\ corruption objective with different corruption rates.
}
\label{tab:objectives_rate}
\end{table}

\subsubsection{Corrupting Spans}
\label{sec:objective_spans}

We now turn towards the goal of speeding up training by predicting shorter targets.
The approach we have used so far makes an i.i.d.\ decision for each input token as to whether to corrupt it or not.
When multiple consecutive tokens have been corrupted, they are treated as a ``span'' and a single unique mask token is used to replace the entire span.
Replacing entire spans with a single token results in unlabeled text data being processed into shorter sequences.
Since we are using an i.i.d.\ corruption strategy, it is not always the case that a significant number of corrupted tokens appear consecutively.
As a result, we might obtain additional speedup by specifically corrupting spans of tokens rather than corrupting individual tokens in an i.i.d.\ manner.
Corrupting spans was also previously considered as a pre-training objective for BERT, where it was found to improve performance \citep{joshi2019spanbert}.

To test this idea, we consider an objective that specifically corrupts contiguous, randomly-spaced spans of tokens.
This objective can be parametrized by the proportion of tokens to be corrupted and the total number of corrupted spans.
The span lengths are then chosen randomly to satisfy these specified parameters.
For example, if we are processing a sequence of $500$ tokens and we have specified that $15\%$ of tokens should be corrupted and that there should be $25$ total spans, then the total number of corrupted tokens would be $500 \times 0.15 = 75$ and the average span length would be $75 / 25 = 3$.
Note that given the original sequence length and corruption rate, we can equivalently parametrize this objective either by the average span length or the total number of spans.
\begin{table}
\footnotesize
\begin{widepage}
\centering
\begin{tabular}{l c c c c c c c c c c}
\toprule
    Span length            & GLUE        & CNNDM       & SQuAD       & SGLUE       & EnDe        & EnFr        & EnRo    \\
\midrule
    \bsl Baseline (i.i.d.) & $\*{83.28}$ & $19.24$     & $80.88$     & $71.36$     & $\*{26.98}$ & $\*{39.82}$ & $\*{27.65}$ \\
    $2$                    & $\*{83.54}$ & $19.39$     & $\*{82.09}$ & $\*{72.20}$ & $\*{26.76}$ & $\*{39.99}$ & $\*{27.63}$ \\
    $3$                    & $\*{83.49}$ & $\*{19.62}$ & $\*{81.84}$ & $\*{72.53}$ & $\*{26.86}$ & $39.65$     & $\*{27.62}$ \\
    $5$                    & $\*{83.40}$ & $19.24$     & $\*{82.05}$ & $\*{72.23}$ & $\*{26.88}$ & $39.40$     & $\*{27.53}$ \\
    $10$                   & $82.85$     & $19.33$     & $\*{81.84}$ & $70.44$     & $\*{26.79}$ & $39.49$     & $\*{27.69}$ \\
\bottomrule
\end{tabular}
\end{widepage}
\caption{
Performance of the span-corruption objective (inspired by \cite{joshi2019spanbert}) for different average span lengths.
In all cases, we corrupt 15\% of the original text sequence.
}
\label{tab:objectives_span}
\end{table}

We compare the span-corruption objective to the i.i.d-corruption objective in \cref{tab:objectives_span}.
We use a corruption rate of $15\%$ in all cases and compare using average span lengths of $2$, $3$, $5$ and $10$.
Again, we find a limited difference between these objectives, though the version with an average span length of $10$ slightly underperforms the other values in some cases.
We also find in particular that using an average span length of $3$ slightly (but significantly) outperforms the i.i.d.\ objective on most non-translation benchmarks.
Fortunately, the span-corruption objective also provides some speedup during training compared to the i.i.d.\ noise approach because span corruption produces shorter sequences on average.

\begin{figure}[t]
    \centering
    \includegraphics[width=0.6\textwidth]{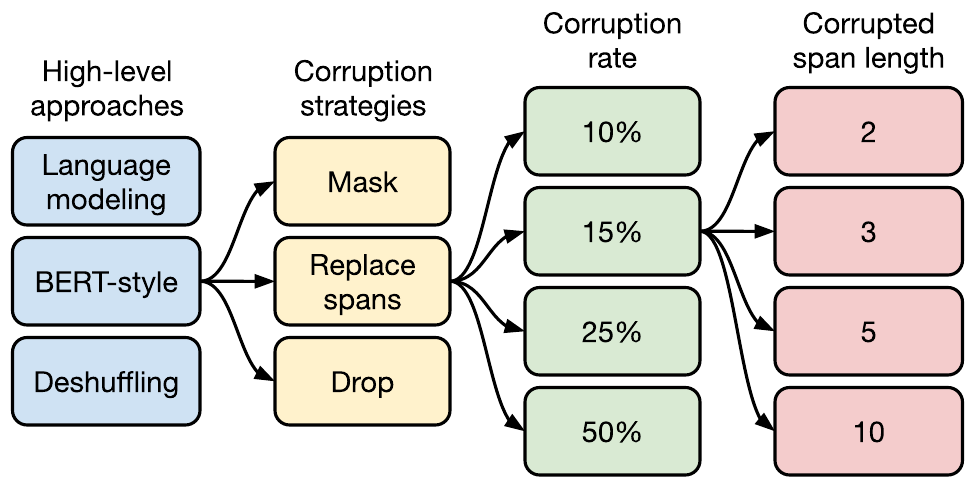}
    \caption{
    A flow chart of our exploration of unsupervised objectives.
    We first consider a few disparate approaches in \cref{sec:objectives_highlevel} and find that a BERT-style denoising objective performs best.
    Then, we consider various methods for simplifying the BERT objective so that it produces shorter target sequences in \cref{sec:simplifying_bert}.
    Given that replacing dropped-out spans with sentinel tokens performs well and results in short target sequences, in \cref{sec:corruption_rate} we experiment with different corruption rates.
    Finally, we evaluate an objective that intentionally corrupts contiguous spans of tokens in \cref{sec:objective_spans}.
    }
    \label{fig:objectives_flow}
\end{figure}

\subsubsection{Discussion}

\Cref{fig:objectives_flow} shows a flow chart of the choices made during our exploration of unsupervised objectives.
Overall, the most significant difference in performance we observed was that denoising objectives outperformed language modeling and deshuffling for pre-training.
We did not observe a remarkable difference across the many variants of the denoising objectives we explored.
However, different objectives (or parameterizations of objectives) can lead to different sequence lengths and thus different training speeds.
This implies that choosing among the denoising objectives we considered here should mainly be done according to their computational cost.
Our results also suggest that additional exploration of objectives similar to the ones we consider here may not lead to significant gains for the tasks and model we consider.
Instead, it may be fortuitous to explore entirely different ways of leveraging unlabeled data.

\subsection{Pre-training Data set}
\label{sec:datasets}

Like the unsupervised objective, the pre-training data set itself is a crucial component of the transfer learning pipeline.
However, unlike objectives and benchmarks, new pre-training data sets are usually not treated as significant contributions on their own and are often not released alongside pre-trained models and code.
Instead, they are typically introduced in the course of presenting a new method or model.
As a result, there has been relatively little comparison of different pre-training data sets as well as a lack of a ``standard'' data set used for pre-training.
Some recent notable exceptions \citep{baevski2019cloze,liu2019roberta,yang2019xlnet} have compared pre-training on a new large (often Common Crawl-sourced) data set to using a smaller preexisting data set (often Wikipedia).
To probe more deeply into the impact of the pre-training data set on performance, in this section we compare variants of our C4 data set and other potential sources of pre-training data.
We release all of the C4 data set variants we consider as part of TensorFlow Datasets.\footnote{\url{https://www.tensorflow.org/datasets/catalog/c4}}

\subsubsection{Unlabeled Data Sets}
\label{sec:datasets_comparison}

In creating C4, we developed various heuristics to filter the web-extracted text from Common Crawl (see \cref{sec:dataset} for a description).
We are interested in measuring whether this filtering results in improved performance on downstream tasks, in addition to comparing it to other filtering approaches and common pre-training data sets.
Towards this end, we compare the performance of our baseline model after pre-training on the following data sets:

\begin{description}
\item[C4] As a baseline, we first consider pre-training on our proposed unlabeled data set as described in \cref{sec:dataset}.

\item[Unfiltered C4] To measure the effect of the heuristic filtering we used in creating C4 (deduplication, removing bad words, only retaining sentences, etc.), we also generate an alternate version of C4 that forgoes this filtering.
Note that we still use \texttt{langdetect} to extract English text.
As a result, our ``unfiltered'' variant still includes some filtering because \texttt{langdetect} sometimes assigns a low probability to non-natural English text.

\item[RealNews-like] Recent work has used text data extracted from news websites \citep{zellers2019defending,baevski2019cloze}.
To compare to this approach, we generate another unlabeled data set by additionally filtering C4 to only include content from one of the domains used in the ``RealNews'' data set \citep{zellers2019defending}.
Note that for ease of comparison, we retain the heuristic filtering methods used in C4; the only difference is that we have ostensibly omitted any non-news content.

\item[WebText-like] Similarly, the WebText data set \citep{radford2019language} only uses content from webpages that were submitted to the content aggregation website Reddit and received a ``score'' of at least 3.
The score for a webpage submitted to Reddit is computed based on the proportion of users who endorse (upvote) or oppose (downvote) the webpage.
The idea behind using the Reddit score as a quality signal is that users of the site would only upvote high-quality text content.
To generate a comparable data set, we first tried removing all content from C4 that did not originate from a URL that appeared in the list prepared by the OpenWebText effort.\footnote{\url{https://github.com/jcpeterson/openwebtext}}
However, this resulted in comparatively little content---only about 2 GB---because most pages never appear on Reddit.
Recall that C4 was created based on a single month of Common Crawl data.
To avoid using a prohibitively small data set, we therefore downloaded 12 months of data from Common Crawl from August 2018 to July 2019, applied our heuristic filtering for C4, then applied the Reddit filter.
This produced a 17 GB WebText-like data set, which is of comparable size to the original 40GB WebText data set \citep{radford2019language}.

\item[Wikipedia] The website Wikipedia consists of millions of encyclopedia articles written collaboratively.
The content on the site is subject to strict quality guidelines and therefore has been used as a reliable source of clean and natural text.
We use the English Wikipedia text data from TensorFlow Datasets,\footnote{\url{https://www.tensorflow.org/datasets/catalog/wikipedia}} which omits any markup or reference sections from the articles.

\item[Wikipedia + Toronto Books Corpus] A drawback of using pre-training data from Wikipedia is that it represents only one possible domain of natural text (encyclopedia articles).
To mitigate this, BERT \citep{devlin2018bert} combined data from Wikipedia with the Toronto Books Corpus (TBC) \citep{zhu2015aligning}.
TBC contains text extracted from eBooks, which represents a different domain of natural language.
BERT's popularity has led to the Wikipedia + TBC combination being used in many subsequent works.

\end{description}

\begin{table}
\footnotesize
\begin{widepage}
\centering
\begin{tabular}{l c c c c c c c c c c}
\toprule
    Data set         & Size  & GLUE        & CNNDM       & SQuAD       & SGLUE       & EnDe        & EnFr        & EnRo    \\
\midrule
    \bsl C4         & 745GB & $83.28$     & $\*{19.24}$ & $80.88$     & $71.36$     & $\*{26.98}$ & $\*{39.82}$ & $\*{27.65}$ \\
    C4, unfiltered  & 6.1TB & $81.46$     & $19.14$     & $78.78$     & $68.04$     & $26.55$     & $39.34$     & $27.21$ \\
    RealNews-like   & 35GB  & $\*{83.83}$ & $\*{19.23}$ & $80.39$     & $72.38$     & $\*{26.75}$ & $\*{39.90}$ & $\*{27.48}$ \\
    WebText-like    & 17GB  & $\*{84.03}$ & $\*{19.31}$ & $\*{81.42}$ & $71.40$     & $\*{26.80}$ & $\*{39.74}$ & $\*{27.59}$ \\
    Wikipedia       & 16GB  & $81.85$     & $\*{19.31}$ & $81.29$     & $68.01$     & $\*{26.94}$ & $39.69$     & $\*{27.67}$ \\
    Wikipedia + TBC & 20GB  & $83.65$     & $\*{19.28}$ & $\*{82.08}$ & $\*{73.24}$ & $\*{26.77}$ & $39.63$     & $\*{27.57}$ \\
\bottomrule
\end{tabular}
\end{widepage}
\caption{
Performance resulting from pre-training on different data sets.
The first four variants are based on our new C4 data set.
}
\label{tab:datasets}
\end{table}

The results achieved after pre-training on each of these data sets is shown in \cref{tab:datasets}.
A first obvious takeaway is that removing the heuristic filtering from C4 uniformly degrades performance and makes the unfiltered variant perform the worst in every task.
Beyond this, we found that in some cases a pre-training data set with a more constrained domain outperformed the diverse C4 data set.
For example, using the Wikipedia + TBC corpus produced a SuperGLUE score of $73.24$, beating our baseline's score (using C4) of $71.36$.
This is almost entirely attributable to a boost in performance from $25.78$ (baseline, C4) to $50.93$ (Wikipedia + TBC) on the Exact Match score for MultiRC (see \cref{tab:giant}).
MultiRC is a reading comprehension data set whose largest source of data comes from fiction books, which is exactly the domain covered by TBC.
Similarly, using the RealNews-like data set for pre-training conferred an increase from $68.16$ to $73.72$ on the Exact Match score for ReCoRD, a data set that measures reading comprehension on news articles.
As a final example, using data from Wikipedia produced significant (but less dramatic) gains on SQuAD, which is a question-answering data set with passages sourced from Wikipedia.
Similar observations have been made in prior work, e.g.\ \citet{beltagy2019scibert} found that pre-training BERT on text from research papers improved its performance on scientific tasks.
The main lesson behind these findings is that \textit{pre-training on in-domain unlabeled data can improve performance on downstream tasks}.
This is unsurprising but also unsatisfying if our goal is to pre-train a model that can rapidly adapt to language tasks from arbitrary domains.
\citet{liu2019roberta} also observed that pre-training on a more diverse data set yielded improvements on downstream tasks.
This observation also motivates the parallel line of research on domain adaptation for natural language processing; for surveys of this field see e.g.\ \cite{ruder2019neural,li2012literature}.

A drawback to only pre-training on a single domain is that the resulting data sets are often substantially smaller.
Similarly, while the WebText-like variant performed as well or better than the C4 data set in our baseline setting, the Reddit-based filtering produced a data set that was about $40\times$ smaller than C4 despite being based on $12\times$ more data from Common Crawl.
Note, however, that in our baseline setup we only pre-train on $2^{35} \approx 34\mathrm{B}$ tokens, which is only about $8$ times larger than the smallest pre-training data set we consider.
We investigate at what point using a smaller pre-training data sets poses an issue in the following section.

\subsubsection{Pre-training Data set Size}
\label{sec:datasets_take}

The pipeline we use to create C4 was designed to be able to create extremely large pre-training data sets.
The access to so much data allows us to pre-train our models without repeating examples.
It is not clear whether repeating examples during pre-training would be helpful or harmful to downstream performance because our pre-training objective is itself stochastic and can help prevent the model from seeing the same exact data multiple times.

To test the effect of limited unlabeled data set sizes, we pre-trained our baseline model on artificially truncated versions of C4.
Recall that we pre-train our baseline model on $2^{35} \approx 34\mathrm{B}$ tokens (a small fraction of the total size of C4).
We consider training on truncated variants of C4 consisting of $2^{29}$, $2^{27}$, $2^{25}$ and $2^{23}$ tokens.
These sizes correspond to repeating the data set $64$, $256$, $1{,}024$, and $4{,}096$ times respectively over the course of pre-training.

\begin{table}
\footnotesize
\begin{widepage}
\centering
\begin{tabular}{l c c c c c c c c c c}
\toprule
    Number of tokens   & Repeats   & GLUE        & CNNDM       & SQuAD       & SGLUE       & EnDe        & EnFr        & EnRo    \\
\midrule
    \bsl Full data set  & $0$       & $\*{83.28}$ & $\*{19.24}$ & $\*{80.88}$ & $\*{71.36}$ & $\*{26.98}$ & $\*{39.82}$ & $\*{27.65}$ \\
    $2^{29}$           & $64$      & $\*{82.87}$ & $\*{19.19}$ & $\*{80.97}$ & $\*{72.03}$ & $\*{26.83}$ & $\*{39.74}$ & $\*{27.63}$ \\
    $2^{27}$           & $256$     & $82.62$     & $\*{19.20}$ & $79.78$     & $69.97$     & $\*{27.02}$ & $\*{39.71}$ & $27.33$ \\
    $2^{25}$           & $1{,}024$ & $79.55$     & $18.57$     & $76.27$     & $64.76$     & $26.38$     & $39.56$     & $26.80$ \\
    $2^{23}$           & $4{,}096$ & $76.34$     & $18.33$     & $70.92$     & $59.29$     & $26.37$     & $38.84$     & $25.81$ \\
\bottomrule
\end{tabular}
\end{widepage}
\caption{
Measuring the effect of repeating data during pre-training.
In these experiments, we only use the first $N$ tokens from C4 (with varying values of $N$ shown in the first column) but still pre-train over $2^{35}$ tokens.
This results in the data set being repeated over the course of pre-training (with the number of repeats for each experiment shown in the second column), which may result in memorization (see \cref{fig:datasets_take_loss}).
}
\label{tab:datasets_take}
\end{table}

\begin{figure}[t]
    \centering
    \includegraphics[width=0.6\textwidth]{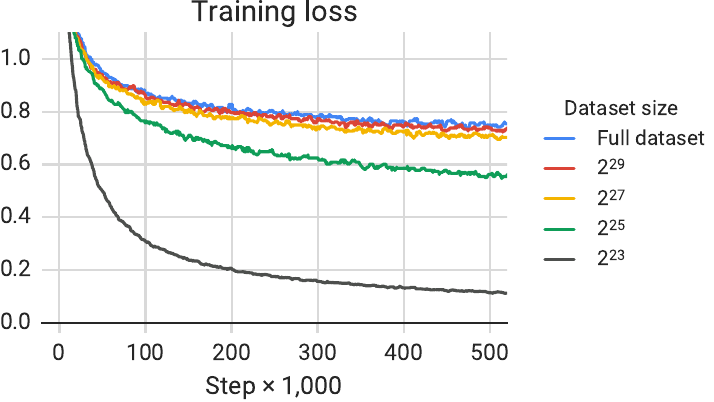}
    \caption{
    Pre-training loss for our original C4 data set as well as $4$ artificially truncated versions.
    The sizes listed refer to the number of tokens in each data set.
    The four sizes considered correspond to repeating the data set between $64$ and $4{,}096$ times over the course of pre-training.
    Using a smaller data set size results in smaller training loss values, which may suggest some memorization of the unlabeled data set.
}
    \label{fig:datasets_take_loss}
\end{figure}

The resulting downstream performance is shown in \cref{tab:datasets_take}.
As expected, performance degrades as the data set size shrinks.
We suspect this may be due to the fact that the model begins to memorize the pre-training data set.
To measure if this is true, we plot the training loss for each of these data set sizes in \cref{fig:datasets_take_loss}.
Indeed, the model attains significantly smaller training losses as the size of the pre-training data set shrinks, suggesting possible memorization.
\citet{baevski2019cloze} similarly observed that truncating the pre-training data set size can degrade downstream task performance.

We note that these effects are limited when the pre-training data set is repeated only $64$ times.
This suggests that some amount of repetition of pre-training data might not be harmful.
However, given that additional pre-training can be beneficial (as we will show in \cref{sec:scaling}) and that obtaining additional unlabeled data is cheap and easy, we suggest using large pre-training data sets whenever possible.
We also note that this effect may be more pronounced for larger model sizes, i.e.\ a bigger model may be more prone to overfitting to a smaller pre-training data set.

\subsection{Training Strategy}
\label{sec:transfer}

So far we have considered the setting where all parameters of a model are pre-trained on an unsupervised task before being fine-tuned on individual supervised tasks.
While this approach is straightforward, various alternative methods for training the model on downstream/supervised tasks have been proposed.
In this section, we compare different schemes for fine-tuning the model in addition to the approach of training the model simultaneously on multiple tasks.

\subsubsection{Fine-tuning Methods}

It has been argued that fine-tuning all of the model's parameters can lead to suboptimal results, particularly on low-resource tasks \citep{peters2019tune}.
Early results on transfer learning for text classification tasks advocated fine-tuning only the parameters of a small classifier that was fed sentence embeddings produced by a fixed pre-trained model \citep{subramanian2018learning,kiros2015skip,logeswaran2018efficient,hill2016learning,conneau2017supervised}.
This approach is less applicable to our encoder-decoder model because the entire decoder must be trained to output the target sequences for a given task.
Instead, we focus on two alternative fine-tuning approaches that update only a subset of the parameters of our encoder-decoder model.

The first, ``adapter layers'' \citep{houlsby2019parameter,bapna2019simple}, is motivated by the goal of keeping most of the original model fixed while fine-tuning.
Adapter layers are additional dense-ReLU-dense blocks that are added after each of the preexisting feed-forward networks in each block of the Transformer.
These new feed-forward networks are designed so that their output dimensionality matches their input.
This allows them to be inserted into the network with no additional changes to the structure or parameters.
When fine-tuning, only the adapter layer and layer normalization parameters are updated.
The main hyperparameter of this approach is the inner dimensionality $d$ of the feed-forward network, which changes the number of new parameters added to the model.
We experiment with various values for $d$.

The second alternative fine-tuning method we consider is ``gradual unfreezing'' \citep{howard2018universal}.
In gradual unfreezing, more and more of the model's parameters are fine-tuned over time.
Gradual unfreezing was originally applied to a language model architecture consisting of a single stack of layers.
In this setting, at the start of fine-tuning only the parameters of the final layer are updated, then after training for a certain number of updates the parameters of the second-to-last layer are also included, and so on until the entire network's parameters are being fine-tuned.
To adapt this approach to our encoder-decoder model, we gradually unfreeze layers in the encoder and decoder in parallel, starting from the top in both cases.
Since the parameters of our input embedding matrix and output classification matrix are shared, we update them throughout fine-tuning.
Recall that our baseline model consists of $12$ layers each in the encoder and decoder and is fine-tuned for $2^{18}$ steps.
As such, we subdivide the fine-tuning process into $12$ episodes of $\sfrac{2^{18}}{12}$ steps each and train from layers $12 - n$ to $12$ in the $n$th episode.
We note that \cite{howard2018universal} suggested fine-tuning an additional layer after each epoch of training.
However, since our supervised data sets vary so much in size and since some of our downstream tasks are actually mixtures of many tasks (GLUE and SuperGLUE), we instead adopt the simpler strategy of fine-tuning an additional layer after every $\sfrac{2^{18}}{12}$ steps.

A comparison of the performance of these fine-tuning approaches is shown in \cref{tab:finetuning}.
For adapter layers, we report the performance using an inner dimensionality $d$ of $32$, $128$, $512$, $2048$.
Pursuant with past results \citep{houlsby2019parameter,bapna2019simple} we find that lower-resource tasks like SQuAD work well with a small value of $d$ whereas higher resource tasks require a large dimensionality to achieve reasonable performance.
This suggests that adapter layers could be a promising technique for fine-tuning on fewer parameters as long as the dimensionality is scaled appropriately to the task size.
Note that in our case we treat GLUE and SuperGLUE each as a single ``task'' by concatenating their constituent data sets, so although they comprise some low-resource data sets the combined data set is large enough that it necessitates a large value of $d$.
We found that gradual unfreezing caused a minor degradation in performance across all tasks, though it did provide some speedup during fine-tuning.
Better results may be attainable by more carefully tuning the unfreezing schedule.

\begin{table}
\footnotesize
\begin{widepage}
\centering
\begin{tabular}{l c c c c c c c c c c}
\toprule
    Fine-tuning method       & GLUE        & CNNDM       & SQuAD       & SGLUE       & EnDe        & EnFr        & EnRo    \\
\midrule
    \bsl All parameters      & $\*{83.28}$ & $\*{19.24}$ & $\*{80.88}$ & $\*{71.36}$ & $\*{26.98}$ & $\*{39.82}$ & $\*{27.65}$ \\
    Adapter layers, $d=32$    & $80.52$ & $15.08$ & $79.32$ & $60.40$ & $13.84$ & $17.88$ & $15.54$ \\
    Adapter layers, $d=128$   & $81.51$ & $16.62$ & $79.47$ & $63.03$ & $19.83$ & $27.50$ & $22.63$ \\
    Adapter layers, $d=512$   & $81.54$ & $17.78$ & $79.18$ & $64.30$ & $23.45$ & $33.98$ & $25.81$ \\
    Adapter layers, $d=2048$  & $81.51$ & $16.62$ & $79.47$ & $63.03$ & $19.83$ & $27.50$ & $22.63$ \\
    Gradual unfreezing       & $82.50$ & $18.95$ & $79.17$ & $\*{70.79}$ & $26.71$ & $39.02$ & $26.93$ \\
\bottomrule
\end{tabular}
\end{widepage}
\caption{
Comparison of different alternative fine-tuning methods that only update a subset of the model's parameters.
For adapter layers, $d$ refers to the inner dimensionality of the adapters.
}
\label{tab:finetuning}
\end{table}

\subsubsection{Multi-task Learning}
\label{sec:multitask}

So far, we have been pre-training our model on a single unsupervised learning task before fine-tuning it individually on each downstream task.
An alternative approach, called ``multi-task learning'' \citep{ruder2017overview,caruana1997multitask}, is to train the model on multiple tasks at a time.
This approach typically has the goal of training a single model that can simultaneously perform many tasks at once, i.e.\ the model and most of its parameters are shared across all tasks.
We relax this goal somewhat and instead investigate methods for training on multiple tasks at once in order to eventually produce separate parameter settings that perform well on each individual task.
For example, we might train a single model on many tasks, but when reporting performance we are allowed to select a different checkpoint for each task.
This loosens the multi-task learning framework and puts it on more even footing compared to the pre-train-then-fine-tune approach we have considered so far.
We also note that in our unified text-to-text framework, ``multi-task learning'' simply corresponds to mixing data sets together.
It follows that we can still train on unlabeled data when using multi-task learning by treating the unsupervised task as one of the tasks being mixed together.
In contrast, most applications of multi-task learning to NLP add task-specific classification networks or use different loss functions for each task \citep{liu2019multi}.

As pointed out by \cite{arivazhagan2019massively}, an extremely important factor in multi-task learning is how much data from each task the model should be trained on.
Our goal is to not under- or over-train the model---that is, we want the model to see enough data from a given task that it can perform the task well, but not to see so much data that it memorizes the training set.
How exactly to set the proportion of data coming from each task can depend on various factors including data set sizes, the ``difficulty'' of learning the task (i.e.\ how much data the model must see before being able to perform the task effectively), regularization, etc.
An additional issue is the potential for ``task interference'' or ``negative transfer'', where achieving good performance on one task can hinder performance on another.
Given these concerns, we begin by exploring various strategies for setting the proportion of data coming from each task.
A similar exploration was performed by \cite{wang2019can}.

\begin{description}
\item[Examples-proportional mixing] A major factor in how quickly a model will overfit to a given task is the task's data set size.
As such, a natural way to set the mixing proportions is to sample in proportion to the size of each task's data set.
This is equivalent to concatenating the data sets for all tasks and randomly sampling examples from the combined data set.
Note, however, that we are including our unsupervised denoising task, which uses a data set that is orders of magnitude larger than every other task's.
It follows that if we simply sample in proportion to each data set's size, the vast majority of the data the model sees will be unlabeled, and it will undertrain on all of the supervised tasks.
Even without the unsupervised task, some tasks (e.g.\ WMT English to French) are so large that they would similarly crowd out most of the batches.
To get around this issue, we set an artificial ``limit'' on the data set sizes before computing the proportions.
Specifically, if the number of examples in each of our $N$ task's data sets is $e_n, n \in \{1, \ldots, N\}$ then we set probability of sampling an example from the $m$th task during training to $r_m = \min(e_m, K)/\sum \min(e_n, K)$ where $K$ is the artificial data set size limit.

\item[Temperature-scaled mixing] An alternative way of mitigating the huge disparity between data set sizes is to adjust the ``temperature'' of the mixing rates.
This approach was used by multilingual BERT to ensure that the model was sufficiently trained on low-resource languages.\footnote{\url{https://github.com/google-research/bert/blob/master/multilingual.md}}
To implement temperature scaling with temperature $T$, we raise each task's mixing rate $r_m$ to the power of \sfrac{1}{$T$} and renormalize the rates so that they sum to 1.
When $T = 1$, this approach is equivalent to examples-proportional mixing and as $T$ increases the proportions become closer to equal mixing.
We retain the data set size limit $K$ (applied to obtain $r_m$ before temperature scaling) but set it to a large value of $K = 2^{21}$.
We use a large value of $K$ because increasing the temperature will decrease the mixing rate of the largest data sets.

\item[Equal mixing] In this case, we sample examples from each task with equal probability.
Specifically, each example in each batch is sampled uniformly at random from one of the data sets we train on.
This is most likely a suboptimal strategy, as the model will overfit quickly on low-resource tasks and underfit on high-resource tasks.
We mainly include it as a point of reference of what might go wrong when the proportions are set suboptimally.

\end{description}

% TODO: Potentially make a table or bar chart showing some examples of proportions
To compare these mixing strategies on equal footing with our baseline pre-train-then-fine-tune results, we train multi-task models for the same total number of steps: $2^{19} + 2^{18} = 786{,}432$.
The results are shown in \cref{tab:multitask}.

In general, we find that multi-task training underperforms pre-training followed by fine-tuning on most tasks.
The ``equal'' mixing strategy in particular results in dramatically degraded performance, which may be because the low-resource tasks have overfit, the high-resource tasks have not seen enough data, or the model has not seen enough unlabeled data to learn general-purpose language capabilities.
For examples-proportional mixing, we find that for most tasks there is a ``sweet spot'' for $K$ where the model obtains the best performance, and larger or smaller values of $K$ tend to result in worse performance.
The exception (for the range of $K$ values we considered) was WMT English to French translation, which is such a high-resource task that it always benefits from a higher mixing proportion.
Finally, we note that temperature-scaled mixing also provides a means of obtaining reasonable performance from most tasks, with $T = 2$ performing the best in most cases.
The finding that a multi-task model is outperformed by separate models trained on each individual task has previously been observed e.g.\ by \citet{arivazhagan2019massively} and \citet{mccann2018natural}, though it has been shown that the multi-task setup can confer benefits across very similar tasks \cite{liu2019multi,ratner2018snorkel}.
In the following section, we explore ways to close the gap between multi-task training and the pre-train-then-fine-tune approach.

\begin{table}
\footnotesize
\begin{widepage}
\centering
\begin{tabular}{l c c c c c c c c c c}
\toprule
    Mixing strategy                     & GLUE        & CNNDM       & SQuAD       & SGLUE       & EnDe        & EnFr        & EnRo    \\
\midrule
    \bsl Baseline (pre-train/fine-tune) & $\*{83.28}$ & $\*{19.24}$ & $\*{80.88}$ & $\*{71.36}$ & $\*{26.98}$ & $\*{39.82}$ & $\*{27.65}$ \\
    Equal                               & $76.13$     & $19.02$     & $76.51$     & $63.37$     & $23.89$     & $34.31$     & $26.78$ \\
    Examples-proportional, $K = 2^{16}$ & $80.45$     & $19.04$     & $77.25$     & $69.95$     & $24.35$     & $34.99$     & $27.10$ \\
    Examples-proportional, $K = 2^{17}$ & $81.56$     & $19.12$     & $77.00$     & $67.91$     & $24.36$     & $35.00$     & $27.25$ \\
    Examples-proportional, $K = 2^{18}$ & $81.67$     & $19.07$     & $78.17$     & $67.94$     & $24.57$     & $35.19$     & $27.39$ \\
    Examples-proportional, $K = 2^{19}$ & $81.42$     & $\*{19.24}$ & $79.78$     & $67.30$     & $25.21$     & $36.30$     & $\*{27.76}$ \\
    Examples-proportional, $K = 2^{20}$ & $80.80$     & $\*{19.24}$ & $\*{80.36}$ & $67.38$     & $25.66$     & $36.93$     & $\*{27.68}$ \\
    Examples-proportional, $K = 2^{21}$ & $79.83$     & $18.79$     & $79.50$     & $65.10$     & $25.82$     & $37.22$     & $27.13$ \\
    Temperature-scaled, $T = 2$         & $81.90$     & $\*{19.28}$ & $79.42$     & $69.92$     & $25.42$     & $36.72$     & $27.20$ \\
    Temperature-scaled, $T = 4$         & $80.56$     & $\*{19.22}$ & $77.99$     & $69.54$     & $25.04$     & $35.82$     & $27.45$ \\
    Temperature-scaled, $T = 8$         & $77.21$     & $19.10$     & $77.14$     & $66.07$     & $24.55$     & $35.35$     & $27.17$ \\
\bottomrule
\end{tabular}
\end{widepage}
\caption{
Comparison of multi-task training using different mixing strategies.
Examples-proportional mixing refers to sampling examples from each data set according to the total size of each data set, with an artificial limit ($K$) on the maximum data set size.
Temperature-scaled mixing re-scales the sampling rates by a temperature $T$.
For temperature-scaled mixing, we use an artificial data set size limit of $K = 2^{21}$.
}
\label{tab:multitask}
\end{table}

\subsubsection{Combining Multi-Task Learning with Fine-Tuning}
\label{sec:mtft}

Recall that we are studying a relaxed version of multi-task learning where we train a single model on a mixture of tasks but are allowed to evaluate performance using different parameter settings (checkpoints) for the model.
We can extend this approach by considering the case where the model is pre-trained on all tasks at once but is then fine-tuned on the individual supervised tasks.
This is the method used by the ``MT-DNN'' \citep{liu2015representation,liu2019multi}, which achieved state-of-the-art performance on GLUE and other benchmarks when it was introduced.
We consider three variants of this approach:
In the first, we simply pre-train the model on an examples-proportional mixture with an artificial data set size limit of $K=2^{19}$ before fine-tuning it on each individual downstream task.
This helps us measure whether including the supervised tasks alongside the unsupervised objective during pre-training gives the model some beneficial early exposure to the downstream tasks.
We might also hope that mixing in many sources of supervision could help the pre-trained model obtain a more general set of ``skills'' (loosely speaking) before it is adapted to an individual task.
To measure this directly, we consider a second variant where we pre-train the model on the same examples-proportional mixture (with $K=2^{19}$) except that we omit one of the downstream tasks from this pre-training mixture.
Then, we fine-tune the model on the task that was left out during pre-training.
We repeat this for each of the downstream tasks we consider.
We call this approach ``leave-one-out'' multi-task training.
This simulates the real-world setting where a pre-trained model is fine-tuned on a task it had not seen during pre-training.
Note that multi-task pre-training provides a diverse mixture of supervised tasks.
Since other fields (e.g.\ computer vision \citep{oquab2014learning,jia2014caffe,huh2016makes,yosinski2014transferable}) use a supervised data set for pre-training, we were interested to see whether omitting the unsupervised task from the multi-task pre-training mixture still produced good results.
For our third variant we therefore pre-train on an examples-proportional mixture of all of the supervised tasks we consider with $K = 2^{19}$.
In all of these variants, we follow our standard procedure of pre-training for $2^{19}$ steps before fine-tuning for $2^{18}$ steps.

\begin{table}
\footnotesize
\begin{widepage}
\centering
\begin{tabular}{l c c c c c c c c c c}
\toprule
    Training strategy                             & GLUE        & CNNDM       & SQuAD       & SGLUE       & EnDe        & EnFr        & EnRo    \\
\midrule
    \bsl Unsupervised pre-training + fine-tuning  & $\*{83.28}$ & $\*{19.24}$ & $\*{80.88}$ & $\*{71.36}$ & $\*{26.98}$ & $39.82$     & $27.65$ \\
    Multi-task training                            & $81.42$     & $\*{19.24}$ & $79.78$     & $67.30$     & $25.21$     & $36.30$     & $27.76$ \\
    Multi-task pre-training + fine-tuning          & $\*{83.11}$ & $\*{19.12}$ & $\*{80.26}$ & $\*{71.03}$ & $\*{27.08}$ & $39.80$     & $\*{28.07}$ \\
    Leave-one-out multi-task training              & $81.98$     & $19.05$     & $79.97$     & $\*{71.68}$ & $\*{26.93}$ & $39.79$     & $\*{27.87}$      \\
    Supervised multi-task pre-training             & $79.93$     & $18.96$     & $77.38$     & $65.36$     & $26.81$     & $\*{40.13}$ & $\*{28.04}$ \\
\bottomrule
\end{tabular}
\end{widepage}
\caption{
Comparison of unsupervised pre-training, multi-task learning, and various forms of multi-task pre-training.
}
\label{tab:multitask_ft}
\end{table}

We compare the results of these approaches in \cref{tab:multitask_ft}.
For comparison, we also include results for our baseline (pre-train then fine-tune) and for standard multi-task learning (without fine-tuning) on an examples-proportional mixture with $K=2^{19}$.
We find that fine-tuning after multi-task pre-training results in comparable performance to our baseline.
This suggests that using fine-tuning after multi-task learning can help mitigate some of the trade-offs between different mixing rates described in \cref{sec:multitask}.
Interestingly, the performance of ``leave-one-out'' training was only slightly worse, suggesting that a model that was trained on a variety of tasks can still adapt to new tasks (i.e.\ multi-task pre-training might not result in a dramatic task interference).
Finally, supervised multi-task pre-training performed significantly worse in every case except for the translation tasks.
This could suggest that the translation tasks benefit less from (English) pre-training, whereas unsupervised pre-training is an important factor in the other tasks.

\subsection{Scaling}
\label{sec:scaling}

The ``bitter lesson'' of machine learning research argues that general methods that can leverage additional computation ultimately win out against methods that rely on human expertise \citep{sutton2019bitter,hestness2017deep,shazeer2017outrageously,jozefowicz2016exploring,mahajan2018exploring,shazeer2018mesh,shazeer2017outrageously,huang2018gpipe,keskar2019ctrl}.
Recent results suggest that this may hold true for transfer learning in NLP \citep{liu2019roberta,radford2019language,yang2019xlnet,lan2019albert}, i.e.\ it has repeatedly been shown that scaling up produces improved performance compared to more carefully-engineered methods.
However, there are a variety of possible ways to scale, including using a bigger model, training the model for more steps, and ensembling.
In this section, we compare these different approaches by addressing the following premise: ``You were just given $4\times$ more compute. How should you use it?''

We start with our baseline model, which has $220\mathrm{M}$ parameters and is pre-trained and fine-tuned for $2^{19}$ and $2^{18}$ steps respectively.
The encoder and decoder are both sized similarly to ``BERT\textsubscript{BASE}''.
To experiment with increased model size, we follow the guidelines of ``BERT\textsubscript{LARGE}'' \cite{devlin2018bert} and use $d_{\mathrm{ff}} = 4096$, $d_{\mathrm{model}} = 1024$, $d_{\mathrm{kv}} = 64$ and $16$-head attention mechanisms.
We then generate two variants with $16$ and $32$ layers each in the encoder and decoder, producing models with $2\times$ and $4\times$ as many parameters as our original model.
These two variants also have a roughly $2\times$ and $4\times$ the computational cost.
Using our baseline and these two larger models, we consider three ways of using $4\times$ as much computation: Training for $4\times$ as many steps, training for $2\times$ as many steps with the $2\times$ bigger model, and training the $4\times$ bigger model for the ``baseline'' number of training steps.
When we increase the training steps, we scale both the pre-train and fine-tune steps for simplicity.
Note that when increasing the number of pre-training steps, we are effectively including more pre-training data as C4 is so large that we do not complete one pass over the data even when training for $2^{23}$ steps.

An alternative way for the model to see $4\times$ as much data is to increase the batch size by a factor of $4$.
This can potentially result in faster training due to more efficient parallelization.
However, training with a $4\times$ larger batch size can yield a different outcome than training for $4\times$ as many steps \citep{shallue2018measuring}.
We include an additional experiment where we train our baseline model with a $4\times$ larger batch size to compare these two cases.

It is common practice on many of the benchmarks we consider to eke out additional performance by training and evaluating using an ensemble of models.
This provides an orthogonal way of using additional computation.
To compare other scaling methods to ensembling, we also measure the performance of an ensemble of $4$ separately pre-trained and fine-tuned models.
We average the logits across the ensemble before feeding them into the output $\mathrm{softmax}$ nonlinearity to obtain an aggregate prediction.
Instead of pre-training $4$ separate models, a cheaper alternative is to take a single pre-trained model and produce $4$ separate fine-tuned versions.
While this does not use our entire $4\times$ computational budget, we also include this method to see if it produces competitive performance to the other scaling methods.

\begin{table}
\footnotesize
\begin{widepage}
\centering
\begin{tabular}{l c c c c c c c c c c}
\toprule
    Scaling strategy                         & GLUE        & CNNDM       & SQuAD       & SGLUE       & EnDe        & EnFr        & EnRo    \\
\midrule
    \bsl Baseline                            & $83.28$     & $19.24$     & $80.88$     & $71.36$     & $26.98$     & $39.82$     & $27.65$ \\
    $1\times$ size, $4\times$ training steps & $85.33$     & $19.33$     & $82.45$     & $74.72$     & $27.08$     & $40.66$     & $27.93$ \\
    $1\times$ size, $4\times$ batch size     & $84.60$     & $19.42$     & $82.52$     & $74.64$     & $27.07$     & $40.60$     & $27.84$ \\
    $2\times$ size, $2\times$ training steps & $\*{86.18}$ & $19.66$     & $\*{84.18}$ & $77.18$     & $27.52$     & $\*{41.03}$ & $28.19$ \\
    $4\times$ size, $1\times$ training steps & $\*{85.91}$ & $19.73$     & $\*{83.86}$ & $\*{78.04}$ & $27.47$     & $40.71$     & $28.10$ \\
    $4\times$ ensembled                      & $84.77$     & $\*{20.10}$ & $83.09$     & $71.74$     & $\*{28.05}$ & $40.53$     & $\*{28.57}$  \\
    $4\times$ ensembled, fine-tune only      & $84.05$     & $19.57$     & $82.36$     & $71.55$     & $27.55$     & $40.22$     & $28.09$  \\
\bottomrule
\end{tabular}
\end{widepage}
\caption{
Comparison of different methods of scaling up our baseline model.
All methods except ensembling fine-tuned models use $4\times$ the computation as the baseline.
``Size'' refers to the number of parameters in the model and ``training time'' refers to the number of steps used for both pre-training and fine-tuning.
}
\label{tab:scaling}
\end{table}

The performance achieved after applying these various scaling methods is shown in \cref{tab:scaling}.
Unsurprisingly, increasing the training time and/or model size consistently improves the baseline.
There was no clear winner between training for $4\times$ as many steps or using a $4\times$ larger batch size, though both were beneficial.
In general, increasing the model size resulted in an additional bump in performance compared to solely increasing the training time or batch size.
We did not observe a large difference between training a $2\times$ bigger model for $2\times$ as long and training a $4\times$ bigger model on any of the tasks we studied.
This suggests that increasing the training time and increasing the model size can be complementary means of improving performance.
Our results also suggest that ensembling provides an orthogonal and effective means of improving performance through scale.
In some tasks (CNN/DM, WMT English to German, and WMT English to Romanian), ensembling $4$ completely separately trained models significantly outperformed every other scaling approach.
Ensembling models that were pre-trained together but fine-tuned separately also gave a substantial performance increase over the baseline, which suggests a cheaper means of improving performance.
The only exception was SuperGLUE, where neither ensembling approach significantly improved over the baseline.

We note that different scaling methods have different trade-offs that are separate from their performance.
For example, using a larger model can make downstream fine-tuning and inference more expensive.
In contrast, the cost of pre-training a small model for longer is effectively amortized if it is applied to many downstream tasks.
Separately, we note that ensembling $N$ separate models has a similar cost to using a model that has an $N\times$ higher computational cost.
As a result, some consideration for the eventual use of the model is important when choosing between scaling methods.

\subsection{Putting It All Together}
\label{sec:together}

We now leverage the insights from our systematic study to determine how far we can push performance on popular NLP benchmarks.
We are also interested in exploring the current limits of transfer learning for NLP by training larger models on large amounts of data.
We start with our baseline training approach and make the following changes:

\begin{description}

\item[Objective] We swap out the i.i.d.\ denoising objective in our baseline for the span-corruption objective described in \cref{sec:objective_spans}, which was loosely inspired by SpanBERT \citep{joshi2019spanbert}.
Specifically, we use a mean span length of $3$ and corrupt $15\%$ of the original sequence.
We found that this objective produced marginally better performance (\cref{tab:objectives_span}) while being slightly more computationally efficient due to shorter target sequence lengths.

\item[Longer training] Our baseline model uses a relatively small amount of pre-training (\sfrac{1}{4} as much as BERT \citep{devlin2018bert}, \sfrac{1}{16} as much as XLNet \citep{yang2019xlnet}, \sfrac{1}{64} as much as RoBERTa \citep{liu2019roberta}, etc.).
Fortunately, C4 is big enough that we can train for substantially longer without repeating data (which can be detrimental, as shown in \cref{sec:datasets_take}).
We found in \cref{sec:scaling} that additional pre-training can indeed be helpful, and that both increasing the batch size and increasing the number of training steps can confer this benefit.
We therefore pre-train our models for $1$ million steps on a batch size of $2^{11}$ sequences of length $512$, corresponding to a total of about $1$ trillion pre-training tokens (about $32\times$ as many as our baseline).
In \cref{sec:datasets_comparison}, we showed that pre-training on the RealNews-like, WebText-like, and Wikipedia + TBC data sets outperformed pre-training on C4 on a few downstream tasks.
However, these data set variants are sufficiently small that they would be repeated hundreds of times over the course of pre-training on $1$ trillion tokens.
Since we showed in \cref{sec:datasets_take} that this repetition could be harmful, we opted instead to continue using the C4 data set.

\item[Model sizes] In \cref{sec:scaling} we also showed how scaling up the baseline model size improved performance.
However, using smaller models can be helpful in settings where limited computational resources are available for fine-tuning or inference.
Based on these factors, we train models with a wide range of sizes:
\begin{itemize}
  \item \textbf{Base.} This is our baseline model, whose hyperparameters are described in \cref{sec:model_hparams}. It has roughly $220$ million parameters.
  \item \textbf{Small.} We consider a smaller model, which scales the baseline down by using $d_{\mathrm{model}} = 512$, $d_{\mathrm{ff}} = 2{,}048$, $8$-headed attention, and only $6$ layers each in the encoder and decoder. This variant has about $60$ million parameters.
  \item \textbf{Large.} Since our baseline uses a BERT\textsubscript{BASE}-sized encoder and decoder, we also consider a variant where the encoder and decoder are both similar in size and structure to BERT\textsubscript{LARGE}. Specifically, this variant uses $d_{\mathrm{model}} = 1{,}024$, $d_{\mathrm{ff}} = 4{,}096$, $d_{\mathrm{kv}} = 64$, $16$-headed attention, and $24$ layers each in the encoder and decoder, resulting in around $770$ million parameters.
  \item \textbf{3B and 11B.} To further explore what kind of performance is possible when using larger models, we consider two additional variants. In both cases, we use $d_{\mathrm{model}} = 1024$, a $24$ layer encoder and decoder, and $d_{\mathrm{kv}} = 128$. For the ``3B'' variant, we use $d_{\mathrm{ff}} = 16{,}384$ with $32$-headed attention, which results in around $2.8$ billion parameters; for ``11B'' we use $d_{\mathrm{ff}} = 65{,}536$ with $128$-headed attention producing a model with about $11$ billion parameters. We chose to scale up $d_{\mathrm{ff}}$ specifically because modern accelerators (such as the TPUs we train our models on) are most efficient for large dense matrix multiplications like those in the Transformer's feed-forward networks.
\end{itemize}

\item[Multi-task pre-training] In \cref{sec:mtft}, we showed that pre-training on a multi-task mixture of unsupervised and supervised tasks before fine-tuning worked as well as pre-training on the unsupervised task alone.
This is the approach advocated by the ``MT-DNN'' \citep{liu2015representation,liu2019multi}.
It also has the practical benefit of being able to monitor ``downstream'' performance for the entire duration of training, rather than just during fine-tuning.
We therefore used multi-task pre-training in our final set of experiments.
We hypothesize that larger models trained for longer might benefit from a larger proportion of unlabeled data because they are more likely to overfit to smaller training data sets.
However, we also note that the results of \cref{sec:mtft} suggest that fine-tuning after multi-task pre-training can mitigate some of the issues that might arise from choosing a suboptimal proportion of unlabeled data.
Based on these ideas, we substitute the following artificial data set sizes for our unlabeled data before using standard example-proportional mixing (described in \cref{sec:multitask}): $710{,}000$ for Small, $2{,}620{,}000$ for Base, $8{,}660{,}000$ for Large, $33{,}500{,}000$ for 3B, and $133{,}000{,}000$ for 11B.
For all model variants, we also capped the effective data set size of the WMT English to French and WMT English to German data sets to $1\mathrm{M}$ examples during pre-training.

\item[Fine-tuning on individual GLUE and SuperGLUE tasks] So far, when fine-tuning on GLUE and SuperGLUE, we have concatenated all of the data sets in each benchmark so that we only fine-tune models once for GLUE and once for SuperGLUE.
This approach makes our study logistically simpler, but we found that this sacrifices a small amount of performance on some tasks compared to fine-tuning on the task separately.
A potential issue with fine-tuning on individual tasks, which would otherwise be mitigated by training on all tasks at once, is that we might overfit quickly to low-resource tasks.
For example, our large batch size of $2^{11}$ length-$512$ sequences would result in the entire data set appearing multiple times in each batch for many of the low-resource GLUE and SuperGLUE tasks.
We therefore use a smaller batch size of $8$ length-$512$ sequences during fine-tuning for each GLUE and SuperGLUE task.
We also save checkpoints every $1{,}000$ steps rather than every $5{,}000$ steps to ensure we have access to the model's parameters before it overfits.

\item[Beam search] All of our previous results were reported using greedy decoding.
For tasks with long output sequences, we found improved performance from using beam search \citep{sutskever2014sequence}.
Specifically, we use a beam width of $4$ and a length penalty of $\alpha = 0.6$ \citep{wu2016google} for the WMT translation and CNN/DM summarization tasks.

\item[Test set] Since this is our final set of experiments, we report results on the test set rather than the validation set.
For CNN/Daily Mail, we use the standard test set distributed with the data set.
For the WMT tasks, this corresponds to using \texttt{newstest2014} for English-German, \texttt{newstest2015} for English-French, and \texttt{newstest2016} for English-Romanian.
For GLUE and SuperGLUE, we used the benchmark evaluation servers to compute official test set scores.\footnote{\url{http://gluebenchmark.com}}\textsuperscript{,}\footnote{\url{http://super.gluebenchmark.com}}
For SQuAD, evaluating on the test set requires running inference on a benchmark server.
Unfortunately, the computational resources on this server are insufficient for obtaining predictions from our largest models.
As a result, we instead continue to report performance on the SQuAD validation set.
Fortunately, the model with the highest performance on the SQuAD test set also reported results on the validation set, so we can still compare to what is ostensibly the state-of-the-art.

\end{description}

Apart from those changes mentioned above, we use the same training procedure and hyperparameters as our baseline (AdaFactor optimizer, inverse square root learning rate schedule for pre-training, constant learning rate for fine-tuning, dropout regularization, vocabulary, etc.).
For reference, these details are described in \cref{sec:setup}.

\begin{table}
\footnotesize
\begin{widepage}
\centering

\begin{tabular}{l c c c c c c c}
\toprule
                  & GLUE       & CoLA       & SST-2         & MRPC          & MRPC          & STS-B         & STS-B    \\
    Model         & Average    & Matthew's  & Accuracy      & F1            & Accuracy      & Pearson       & Spearman \\
\cmidrule(r){1-1}
\cmidrule(l){2-8}
    Previous best & $89.4$\xa  & $69.2$\xb     & $97.1$\xa    & $\*{93.6}$\xb & $\*{91.5}$\xb  & $92.7$\xb  & $92.3$\xb   \\
    % \bsl Baseline & $77.8$     & $44.9$        & $92.6$        & $89.4$        & $85.9$        & $81.3$     & $81.0$      \\
    % Baseline-1T   & $81.4$     & $52.9$        & $94.3$        & $90.5$        & $87.3$        & $82.2$     & $81.9$      \\
    T5-Small      & $77.4$     & $41.0$        & $91.8$        & $89.7$        & $86.6$        & $85.6$     & $85.0$      \\
    T5-Base       & $82.7$     & $51.1$        & $95.2$        & $90.7$        & $87.5$        & $89.4$     & $88.6$      \\
    T5-Large      & $86.4$     & $61.2$        & $96.3$        & $92.4$        & $89.9$        & $89.9$     & $89.2$      \\
    T5-3B         & $88.5$     & $67.1$        & $97.4$        & $92.5$        & $90.0$        & $90.6$     & $89.8$      \\
    T5-11B        & $\*{90.3}$ & $\*{71.6}$    & $\*{97.5}$    & $92.8$        & $90.4$        & $\*{93.1}$ & $\*{92.8}$  \\
\bottomrule
\end{tabular}

\vspace{0.2em}

\begin{tabular}{l c c c c c c c}
                  & QQP           & QQP           & MNLI-m     & MNLI-mm     & QNLI         & RTE       & WNLI     \\
    Model         & F1            & Accuracy      & Accuracy   & Accuracy    & Accuracy     & Accuracy  & Accuracy \\
\cmidrule(r){1-1}
\cmidrule(l){2-8}
    Previous best & $74.8$\xc     & $\*{90.7}$\xb & $91.3$\xa  & $91.0$\xa  & $\*{99.2}$\xa & $89.2$\xa  & $91.8$\xa \\
    % \bsl Baseline & $70.6$        & $89.1$        & $83.7$     & $82.4$     & $89.4$       & $70.9$     & $69.9$        \\
    % Baseline-1T   & $71.6$        & $88.8$        & $86.9$     & $86.1$     & $92.3$       & $77.2$     & $78.1$        \\
    T5-Small      & $70.0$        & $88.0$        & $82.4$     & $82.3$     & $90.3$       & $69.9$     & $69.2$        \\
    T5-Base       & $72.6$        & $89.4$        & $87.1$     & $86.2$     & $93.7$       & $80.1$     & $78.8$        \\
    T5-Large      & $73.9$        & $89.9$        & $89.9$     & $89.6$     & $94.8$       & $87.2$     & $85.6$        \\
    T5-3B         & $74.4$        & $89.7$        & $91.4$     & $91.2$     & $96.3$       & $91.1$     & $89.7$        \\
    T5-11B        & $\*{75.1}$    & $90.6$        & $\*{92.2}$ & $\*{91.9}$ & $96.9$       & $\*{92.8}$ & $\*{94.5}$    \\
\bottomrule
\end{tabular}

\vspace{0.2em}

\begin{tabular}{l c c c c c c c}
                  & SQuAD       & SQuAD       & SuperGLUE & BoolQ       & CB          & CB        & COPA     \\
    Model         & EM          & F1          & Average   & Accuracy    & F1          & Accuracy  & Accuracy \\
\cmidrule(r){1-1}
\cmidrule(lr){2-3}
\cmidrule(l){4-8}
    Previous best & $90.1$\xa   & $95.5$\xa   & $84.6$\xd  & $87.1$\xd  & $90.5$\xd  & $95.2$\xd  & $90.6$\xd \\
    % \bsl Baseline & $81.01$     & $88.86$     & $66.6$     & $75.0$     & $83.8$     & $88.8$     & $57.0$     \\
    % Baseline-1T   & $83.01$     & $90.67$     & $72.5$     & $78.7$     & $83.9$     & $90.0$     & $68.6$     \\
    T5-Small      & $79.10$     & $87.24$     & $63.3$     & $76.4$     & $56.9$     & $81.6$     & $46.0$     \\
    T5-Base       & $85.44$     & $92.08$     & $76.2$     & $81.4$     & $86.2$     & $94.0$     & $71.2$     \\
    T5-Large      & $86.66$     & $93.79$     & $82.3$     & $85.4$     & $91.6$     & $94.8$     & $83.4$     \\
    T5-3B         & $88.53$     & $94.95$     & $86.4$     & $89.9$     & $90.3$     & $94.4$     & $92.0$     \\
    T5-11B        & $\*{91.26}$ & $\*{96.22}$ & $\*{88.9}$ & $\*{91.2}$ & $\*{93.9}$ & $\*{96.8}$ & $\*{94.8}$ \\
\bottomrule
\end{tabular}

\vspace{0.2em}

\begin{tabular}{l c c c c c c c}
                  & MultiRC     & MultiRC   & ReCoRD     & ReCoRD      & RTE       & WiC        & WSC \\
    Model         & F1a         & EM        & F1         & Accuracy    & Accuracy  & Accuracy   & Accuracy\\
\cmidrule(r){1-1}
\cmidrule(l){2-8}
    Previous best & $84.4$\xd  & $52.5$\xd  & $90.6$\xd  & $90.0$\xd  & $88.2$\xd  & $69.9$\xd  & $89.0$\xd \\
    % \bsl Baseline & $69.7$     & $22.9$     & $65.6$     & $64.8$     & $67.4$     & $64.9$     & $70.5$     \\
    % Baseline-1T   & $73.1$     & $31.6$     & $75.1$     & $74.2$     & $72.8$     & $66.1$     & $79.5$     \\
    T5-Small      & $69.3$     & $26.3$     & $56.3$     & $55.4$     & $73.3$     & $66.9$     & $70.5$     \\
    T5-Base       & $79.7$     & $43.1$     & $75.0$     & $74.2$     & $81.5$     & $68.3$     & $80.8$     \\
    T5-Large      & $83.3$     & $50.7$     & $86.8$     & $85.9$     & $87.8$     & $69.3$     & $86.3$     \\
    T5-3B         & $86.8$     & $58.3$     & $91.2$     & $90.4$     & $90.7$     & $72.1$     & $90.4$     \\
    T5-11B        & $\*{88.1}$ & $\*{63.3}$ & $\*{94.1}$ & $\*{93.4}$ & $\*{92.5}$ & $\*{76.9}$ & $\*{93.8}$ \\
\bottomrule
\end{tabular}

\vspace{0.2em}

\begin{tabular}{l c c c c c c}
                  & WMT EnDe      & WMT EnFr      & WMT EnRo      & CNN/DM      & CNN/DM      & CNN/DM  \\
    Model         & BLEU          & BLEU          & BLEU          & ROUGE-1     & ROUGE-2     & ROUGE-L \\
\cmidrule(r){1-1}
\cmidrule(lr){2-4}
\cmidrule(l){5-7}
    Previous best & $\*{33.8}$\xe & $\*{43.8}$\xe & $\*{38.5}$\xf & $43.47$\xg  & $20.30$\xg  & $40.63$\xg \\
    % \bsl Baseline & $29.3$        & $40.0$        & $26.8$        & $40.91$     & $19.31$     & $38.15$     \\
    % Baseline-1T   & $30.1$        & $40.7$        & $27.4$        & $41.66$     & $19.91$     & $38.93$     \\
    T5-Small      & $26.7$        & $36.0$        & $26.8$        & $41.12$     & $19.56$     & $38.35$     \\
    T5-Base       & $30.9$        & $41.2$        & $28.0$        & $42.05$     & $20.34$     & $39.40$     \\
    T5-Large      & $32.0$        & $41.5$        & $28.1$        & $42.50$     & $20.68$     & $39.75$     \\
    T5-3B         & $31.8$        & $42.6$        & $28.2$        & $42.72$     & $21.02$     & $39.94$     \\
    T5-11B        & $32.1$        & $43.4$        & $28.1$        & $\*{43.52}$ & $\*{21.55}$ & $\*{40.69}$ \\
\bottomrule
\end{tabular}

\end{widepage}
\caption{
Performance of our T5 variants on every task we study.
Small, Base, Large, 3B, and 11B refer to model configurations with $60$ million, $220$ million, $770$ million, $3$ billion, and $11$ billion parameters, respectively.
In the first row of each table, we report the state-of-the-art for the task (as of October 24th, 2019), with the superscript denoting its source with references listed at the end of this caption.
All results are reported on the test set except for SQuAD where we use the validation set.
$^a$\citep{lan2019albert}
$^b$\citep{wang2019structbert}
$^c$\citep{zhu2019freelb}
$^d$\citep{liu2019roberta}
$^e$\citep{edunov2018understanding}
$^f$\citep{lample2019cross}
$^g$\citep{dong2019unified}
}
\label{tab:final}
\end{table}

The results of this final set of experiments are shown in \cref{tab:final}.
Overall, we achieved state-of-the-art performance on $18$ out of the $24$ tasks we consider.
As expected, our largest ($11$ billion parameter) model performed best among our model size variants across all tasks.
Our T5-3B model variant did beat the previous state of the art in a few tasks, but scaling the model size to $11$ billion parameters was the most important ingredient for achieving our best performance.
We now analyze the results for each individual benchmark.

We achieved a state-of-the-art average GLUE score of $90.3$.
Notably, our performance was substantially better than the previous state-of-the-art for the natural language inference tasks MNLI, RTE, and WNLI.
RTE and WNLI are two of the tasks where machine performance has historically lagged behind human performance, which is $93.6$ and $95.9$ respectively \citep{wang2018glue}.
In terms of parameter count, our 11B model variant is the largest model that has been submitted to the GLUE benchmark.
However, most of the best-scoring submissions use a large amount of ensembling and computation to produce predictions.
For example, the best-performing variant of ALBERT \citep{lan2019albert} uses a model similar in size and architecture to our 3B variant (though it has dramatically fewer parameters due to clever parameter sharing).
To produce its impressive performance on GLUE, the ALBERT authors ensembled ``from 6 to 17'' models depending on the task.
This likely results in it being more computationally expensive to produce predictions with the ALBERT ensemble than it is with T5-11B.

For SQuAD, we outperformed the previous state-of-the-art (ALBERT \citep{lan2019albert}) by over one point on the Exact Match score.
SQuAD is a long-standing benchmark that was created over three years ago, and most recent improvements have only increased the state-of-the-art by a fraction of a percentage point.
We note that when results are reported on the test set, they are typically based on an ensemble of models and/or leverage external data sets (e.g.\ TriviaQA \citep{joshi2017triviaqa} or NewsQA \citep{trischler2016newsqa}) to augment the small SQuAD training set.
Human performance on SQuAD is estimated at $82.30$ and $91.22$ for the Exact Match and F1 metric respectively \citep{rajpurkar2016squad}, so it is not clear if further improvements on this benchmark are meaningful.

For SuperGLUE, we improved upon the state-of-the-art by a large margin (from an average score of $84.6$ \citep{liu2019roberta} to $88.9$).
SuperGLUE was designed to include tasks that were ``beyond the scope of current state-of-the-art systems, but solvable by most college-educated English speakers'' \citep{wang2019superglue}.
We nearly match the human performance of $89.8$ \citep{wang2019superglue}.
Interestingly, on the reading comprehension tasks (MultiRC and ReCoRD) we exceed human performance by a large margin, suggesting the evaluation metrics used for these tasks may be biased towards machine-made predictions.
On the other hand, humans achieve $100\%$ accuracy on both COPA and WSC, which is significantly better than our model's performance.
This suggests that there remain linguistic tasks that are hard for our model to perfect, particularly in the low-resource setting.

We did not achieve state-of-the-art performance on any of the WMT translation tasks.
This may be in part due to our use of an English-only unlabeled data set.
We also note that most of the best results on these tasks use backtranslation \citep{edunov2018understanding,lample2019cross}, which is a sophisticated data augmentation scheme.
The state of the art on the low-resource English to Romanian benchmark also uses additional forms of cross-lingual unsupervised training \citep{lample2019cross}.
Our results suggest that scale and English-language pre-training may be insufficient to match the performance of these more sophisticated methods.
On a more specific note, the best results on English to German \texttt{newstest2014} set use the much larger training set from WMT 2018 \citep{edunov2018understanding}, making direct comparison to our results difficult.

Finally, on CNN/Daily Mail we attain state-of-the-art performance, though only by a significant amount on the ROUGE-2-F score.
It has been shown that improvements to the ROUGE score do not necessarily correspond to more coherent summaries \citep{paulus2017deep}.
Furthermore, while CNN/Daily Mail is posed as an abstractive summarization benchmark, purely extractive approaches have been shown to work well \citep{liu2019fine}.
It has also been argued that generative models trained with maximum likelihood are prone to producing repetitive summaries \citep{see2017get}.
Despite these potential issues, we find that our models do generate coherent and largely correct summaries.
We provide some non-cherry-picked validation set examples in \cref{sec:cnndm_decodes}.

To achieve its strong results, T5 combines insights from our experimental study with unprecedented scale.
Note that in \cref{sec:scaling} we found that scaling up the pre-training amount or size of our baseline model produced substantial gains.
Given this, we were interested to measure how much the ``non-scaling'' changes we introduced into T5 contributed to its strong performance.
We therefore carried out a final experiment where we compared the following three configurations:
First, the standard baseline model, which was pre-trained on $2^{35} \approx 34\mathrm{B}$ tokens;
second, the baseline trained instead for about 1 trillion tokens (i.e.\ the same amount of pre-training used for T5), which we refer to as ``baseline-1T'';
and third, T5-Base.
Note that the differences between baseline-1T and T5-Base comprise the ``non-scaling'' changes we made when designing T5.
As such, comparing the performance of these two models gives us a concrete measurement of the impact of the insights from our systematic study.

The performance of these three model configurations is shown in \cref{tab:insight_impact}.
Consistent with the findings in \cref{sec:scaling}, we find that additional pre-training improves performance over the baseline.
Nevertheless, T5-Base substantially outperforms baseline-1T on all downstream tasks.
This suggests that scale is not the only factor that contributes to T5's success.
We hypothesize that the larger models benefit not only from their increased size but also from these non-scaling factors.

\begin{table}
\footnotesize
\begin{widepage}
\centering
\begin{tabular}{l c c c c c c c c c c}
\toprule
    Model          & GLUE        & CNNDM       & SQuAD       & SGLUE       & EnDe        & EnFr        & EnRo    \\
\midrule
    \bsl Baseline  & $83.28$     & $19.24$     & $80.88$     & $71.36$     & $26.98$     & $39.82$     & $27.65$ \\
    Baseline-1T    & $84.80$     & $19.62$     & $83.01$     & $73.90$     & $27.46$     & $40.30$     & $28.34$ \\
    T5-Base        & $\*{85.97}$ & $\*{20.90}$ & $\*{85.44}$ & $\*{75.64}$ & $\*{28.37}$ & $\*{41.37}$ & $\*{28.98}$ \\
\bottomrule
\end{tabular}
\end{widepage}
\caption{
Performance comparison of T5-Base to our baseline experimental setup used in the rest of the paper.
Results are reported on the validation set.
``Baseline-1T'' refers to the performance achieved by pre-training the baseline model on 1 trillion tokens (the same number used for the T5 model variants) instead of $2^{35} \approx 34\mathrm{B}$ tokens (as was used for the baseline).
}
\label{tab:insight_impact}
\end{table}

\section{Reflection}
\label{sec:conclusion}

Having completed our systematic study, we wrap up by first recapping some of our most significant findings.
Our results provide some high-level perspective on which avenues of research might be more or less promising.
To conclude, we outline some topics we think might provide effective approaches for further progressing the field.

\subsection{Takeaways}

\begin{description}

\item[Text-to-text] Our text-to-text framework provides a simple way to train a single model on a wide variety of text tasks using the same loss function and decoding procedure.
We showed how this approach can be successfully applied to generative tasks like abstractive summarization, classification tasks like natural language inference, and even regression tasks like STS-B.
In spite of its simplicity, we found the text-to-text framework obtained comparable performance to task-specific architectures and ultimately produced state-of-the-art results when combined with scale.

\item[Architectures] While some work on transfer learning for NLP has considered architectural variants of the Transformer, we found the original encoder-decoder form worked best in our text-to-text framework.
Though an encoder-decoder model uses twice as many parameters as ``encoder-only'' (e.g.\ BERT) or ``decoder-only'' (language model) architectures, it has a similar computational cost.
We also showed that sharing the parameters in the encoder and decoder did not result in a substantial performance drop while halving the total parameter count.

\item[Unsupervised objectives] Overall, we found that most ``denoising'' objectives, which train the model to reconstruct randomly corrupted text, performed similarly in the text-to-text setup.
As a result, we suggest using objectives that produce short target sequences so that unsupervised pre-training is more computationally efficient.

\item[Data sets] We introduced the ``Colossal Clean Crawled Corpus'' (C4), which comprises heuristically-cleaned text from the Common Crawl web dump.
When comparing C4 to data sets that use additional filtering, we found that training on in-domain unlabeled data could boost performance in a few downstream tasks.
However, constraining to a single domain typically results in a smaller data set.
We separately showed that performance can degrade when an unlabeled data set is small enough that it is repeated many times over the course of pre-training.
This motivates the use of a large and diverse data set like C4 for generic language understanding tasks.

\item[Training strategies] We found that the basic approach of updating all of a pre-trained model's parameters during fine-tuning outperformed methods that are designed to update fewer parameters, although updating all parameters is most expensive.
We also experimented with various approaches for training the model on multiple tasks at once, which in our text-to-text setting simply corresponds to mixing examples from different data sets when constructing batches.
The primary concern in multi-task learning is setting the proportion of each task to train on.
We ultimately did not find a strategy for setting mixing proportions that matched the performance of the basic approach of unsupervised pre-training followed by supervised fine-tuning.
However, we found that fine-tuning after pre-training on a mixture of tasks produced comparable performance to unsupervised pre-training.

\item[Scaling] We compared various strategies for taking advantage of additional compute, including training the model on more data, training a larger model, and using an ensemble of models.
We found each approach conferred a significant boost in performance, though training a smaller model on more data was often outperformed by training a larger model for fewer steps.
We also showed an ensemble of models can provide substantially better results than a single model, which provides an orthogonal means of leveraging additional computation.
Ensembling models that were fine-tuned from the same base pre-trained model performed worse than pre-training and fine-tuning all models completely separately, though fine-tune-only ensembling still substantially outperformed a single model.

\item[Pushing the limits] We combined our above insights and trained substantially larger models (up to $11$ billion parameters) to achieve state-of-the-art results across many of the benchmarks we considered.
For unsupervised training, we extracted text from our C4 data set and applied a denoising objective that corrupts contiguous spans of tokens.
We pre-trained on a multi-task mixture before fine-tuning on individual tasks.
Overall, our models were trained on over $1$ trillion tokens.
In the interest of facilitating the replication, extension, and application of our results, we release our code, the C4 data set, and pre-trained model weights for each T5 variant.\footnoteref{fn:oss}

\end{description}

\subsection{Outlook}

\begin{description}

\item[The inconvenience of large models] An unsurprising but important result from our study is that larger models tend to perform better.
The fact that the hardware used for running these models is continually getting cheaper and more powerful suggests that scaling up may continue to be a promising way to achieve better performance \citep{sutton2019bitter}.
However, it will always be the case that there are applications and scenarios where using a smaller or less expensive model is helpful, for example when performing client-side inference or federated learning \citep{konevcny2015federated,konevcny2016federated}.
Relatedly, one beneficial use of transfer learning is the possibility of attaining good performance on low-resource tasks.
Low-resource tasks often occur (by definition) in settings where one lacks the assets to label more data.
It follows that low-resource applications often also have limited access to computational resources which can incur additional costs.
As a result, we advocate for research on methods that achieve stronger performance with cheaper models so that transfer learning can be applied where it will have the most impact.
Some current work along these lines include distillation \citep{hinton2015distilling,sanh2019distilbert,jiao2019tinybert}, parameter sharing \citep{lan2019albert}, and conditional computation \citep{shazeer2017outrageously}.

\item[More efficient knowledge extraction] Recall that one of the goals of pre-training is (loosely speaking) to provide the model with general-purpose ``knowledge'' that improves its performance on downstream tasks.
The method we use in this work, which is currently common practice, is to train the model to denoise corrupted spans of text.
We suspect that this simplistic technique may not be a very efficient way to teach the model general-purpose knowledge.
More concretely, it would be useful to be able to attain good fine-tuning performance without needing to train our models on $1$ trillion tokens of text first.
Some concurrent work along these lines improves efficiency by pre-training a model to distinguish between real and machine-generated text \citep{clark2020electra}.

\item[Formalizing the similarity between tasks] We observed that pre-training on unlabeled in-domain data can improve performance on downstream tasks (\cref{sec:datasets}).
This finding mostly relies on basic observations like the fact that SQuAD was created using data from Wikipedia.
It would be useful to formulate a more rigorous notion of the ``similarity'' between the pre-training and downstream tasks, so that we could make more principled choices about what source of unlabeled data to use.
There is some early empirical work along these lines in the field of computer vision \citep{huh2016makes,kornblith2018better,he2018rethinking}.
A better notion of the relatedness of tasks could also help choose \textit{supervised} pre-training tasks, which has been shown to be helpful for the GLUE benchmark \citep{phang2018sentence}.

\item[Language-agnostic models] We were disappointed to find that English-only pre-training did not achieve state-of-the-art results on the translation tasks we studied.
We also are interested in avoiding the logistical difficulty of needing to specify which languages a vocabulary can encode ahead of time.
To address these issues, we are interested in further investigating language-agnostic models, i.e.\ models that can perform a given NLP task with good performance regardless of the text's language.
This is an especially pertinent issue given that English is not the native language for the majority of the world's population.

The motivation for this paper was the flurry of recent work on transfer learning for NLP.
Before we began this work, these advances had already enabled breakthroughs in settings where learning-based methods had not yet been shown to be effective.
We are happy to be able to continue this trend, for example by nearly matching human-level performance on the SuperGLUE benchmark, a task specifically designed to be difficult for modern transfer-learning pipelines.
Our results stem from the combination of a straightforward and unified text-to-text framework, our new C4 data set, and insights from our systematic study.
Additionally, we provided an empirical overview of the field and a perspective on where it stands.
We are excited to see continued work using transfer learning towards the goal of general language understanding.

\end{description}

\acks{We thank Grady Simon, Noah Fiedel, Samuel R.\ Bowman, Augustus Odena, Daphne Ippolito, Noah Constant, Orhan Firat, Ankur Bapna, and Sebastian Ruder for their comments on this manuscript; Zak Stone and the TFRC team for their support; Austin Tarango for his guidance on data set creation; Melvin Johnson, Dima Lepikhin, Katrin Tomanek, Jeff Klingner, and Naveen Arivazhagan for insight into multi-task machine translation; Neil Houlsby for comments on adapter layers; Olga Wichowska, Ola Spyra, Michael Banfield, Yi Lin, and Frank Chen for assistance with infrastructure; Etienne Pot, Ryan Sepassi, and Pierre Ruyssen for collaboration on TensorFlow Datasets; Rohan Anil for help with our download pipeline for Common Crawl; Robby Neale and Taku Kudo for their work on SentencePiece; Jeffrey Li for pointing out missing details about the creation of C4; and many other members of the Google Brain team for their discussion and insight.}

\clearpage

\appendix

\section{Contributions}
\label{sec:contributions}

Colin designed the scope of this project and wrote this paper, ran all the experiments in \cref{sec:baseline,sec:architectures,sec:objectives,sec:datasets,sec:transfer,sec:scaling}, and contributed a large portion of our codebase.
Noam contributed many of the ideas, including the text-to-text framework, unsupervised objectives, and data set mixing strategies; implemented our base Transformer model and its architectural variants; and ran the experiments in \cref{sec:together}.
Adam oversaw all engineering aspects for this project, created the C4 data set, implemented our data set pipeline, and added various benchmark data sets.
Katherine coordinated experiments, wrote and updated documentation, ran experiments to help design our baseline, and contributed to many parts of our codebase.
Sharan contributed some of the required data sets and preprocessors, and ran assorted preliminary experiments, in addition to co-leading the open-sourcing of our codebase.
Michael owned all aspects of the Winograd data sets, ingested many of the data sets we used, contributed various improvements and fixes to our infrastructure, and ran some preliminary experiments.
Yanqi ran experiments and implemented methods to help settle on a reasonable baseline and helped with the final fine-tuning of the models in \cref{sec:together}.
Wei also helped with final fine-tuning and improved some of our preprocessors.
Peter prototyped an early version of the pre-training data set and resolved issues pertaining to the SQuAD and CNN/DM tasks.
All authors helped set the scope and research direction we followed in this work.

\section{Converting WNLI to Our Text-to-Text Format}
\label{sec:wnli_preprocessing}

Note that as discussed in \cref{sec:format}, we do not train on any of the data from WNLI.
Instead, when evaluating on the WNLI test set (for the results in \cref{sec:together}), we convert the WNLI test set to the ``referent noun prediction'' text-to-text format so that we can evaluate using a model trained on WSC and DPR.
Our WNLI preprocessor is inspired by the one proposed by \cite{he2019hybrid}.
Recall that examples from WNLI consist of a premise, a hypothesis, and a label that indicates whether the hypothesis is True or False.
Using the example from \cref{sec:format}, the hypothesis would be ``The city councilmen refused the demonstrators a permit because they
feared violence.'' with the premise ``The demonstrators feared violence.'' and the label False.
We first find the location of all pronouns in the premise (``they'' in our example).
Then, we find the maximum number of words that precede or follow each pronoun that are a substring in the hypothesis (``feared violence'' in our example), ignoring case and punctuation.
When the premise contains multiple candidate pronouns, we choose the pronoun that is preceded or followed by the largest substring of the hypothesis.
We then highlight the pronoun in the premise by surrounding it with asterisks.
For the candidate noun (which is compared to our model's prediction to obtain a True or False label), we remove the matching substring from the hypothesis and optionally make it non-possessive (resulting in ``the demonstrators'').

\section{Example Predictions on CNN/Daily Mail}
\label{sec:cnndm_decodes}

To show that our model is generating fluent summaries, we include a few example decodes from our best model (T5-11B) on the validation set along with the ground-truth summaries.
These examples selected at random and were not cherry-picked.

\begin{enumerate}
\item \textbf{Ground-truth:} \texttt{leopard gave up after spiky creature refused to back down in fight in kruger national park, south africa . wildlife enthusiast lisl moolman, 41, caught the bizarre battle while out on the road and armed with her camera . the leopard and porcupine tumbled out of the bushes and began to fight by roadside - watched by ms moolman .}

\textbf{Prediction:} \texttt{leopard tried to make lunch out of a plucky porcupine in kruger national park, south africa . but the predator was put firmly in its place after the spiky creature refused to back down during a fight . wildlife enthusiast lisl moolman, 41, caught the bizarre battle while out on the road and armed with her camera .}

\item \textbf{Ground-truth:} \texttt{researchers say homes are a hotbed of undiscovered species . study of 40 north carolina homes found 8,000 bacterial and archaeal taxa . 11 houses in california revealed a great variety of fungus among us .}

\textbf{Prediction:} \texttt{researchers estimate that the indoor biome makes up about 0.5 percent of ice-free land, or about 247,000 square miles, almost the size of texas . a study of 40 houses in north carolina used genomic technologies to document more than 8,000 bacterial and archaeal taxa . another study of 50 houses in the same state described more than 750 types of arthropods, including insects and spiders . an examination of 11 houses in california found a great variety of fungus among us .}

\item \textbf{Ground-truth:} \texttt{executions have been temporarily halted in the state after the drug used for lethal injections appeared faulty on monday night . kelly gissendaner had her death delayed for a second time when it was found that the lethal pentobarbital dose appeared cloudy . state officials dithered over the decision , changing their mind three times before deciding not to take the risk that it wouldn't work . it is 'essential that executions are carried out in a constitutional manner,' said georgia attorney general sam olens .}

\textbf{Prediction:} \texttt{the execution of kelly gissendaner, the only woman on georgia's death row, was delayed for a second time on monday . the execution team found that the lethal pentobarbital dose appeared cloudy . the cloudy drug bolstered death penalty opponents, who have been vocal in their opposition after three botched executions in other parts of the country .}

\item \textbf{Ground-truth:} \texttt{dani alves was not selected for the brazil squad to face france and chile . barcelona defender was filmed serving up hot dogs to people on saturday . this week alves released a charity single with former team-mat jose pinto . alves looks set to leave barcelona on a free transfer this summer . reports in spanish press claim he will sign for paris saint-germain .}

\textbf{Prediction:} \texttt{dani alves was not selected for brazil's friendlies against france and chile . the barcelona right back has released a charity single with jose pinto . alves was filmed serving up snacks from behind the counter of a hot dog bar on saturday . alves is yet to be offered a new deal at the nou camp . click here for all the latest barcelona news .}

\end{enumerate}

\section{Preprocessed Examples}
\label{sec:preprocessing}

% ODO: Fill in this section.

In this section, we provide examples of our preprocessing for each of the data sets we consider.

\subsection{CoLA}
\begin{description}[leftmargin=0.5cm]
\item[Original input:] ~
\begin{description}[leftmargin=0.5cm]
  \item[Sentence:] \texttt{John made Bill master of himself.}
\end{description}
\item[Processed input:] \texttt{cola sentence:\ John made Bill master of himself.}
\item[Original target:] \texttt{1}
\item[Processed target:] \texttt{acceptable}
\end{description}

\subsection{RTE}
\begin{description}[leftmargin=0.5cm]
\item[Original input:] ~
\begin{description}[leftmargin=0.5cm]
  \item[Sentence 1:] \texttt{A smaller proportion of Yugoslavia's Italians were settled in Slovenia (at the 1991 national census, some 3000 inhabitants of Slovenia declared themselves as ethnic Italians).}
  \item[Sentence 2:] \texttt{Slovenia has 3,000 inhabitants.}
\end{description}
\item[Processed input:] \texttt{rte sentence1:\ A smaller proportion of Yugoslavia's Italians were settled in Slovenia (at the 1991 national census, some 3000 inhabitants of Slovenia declared themselves as ethnic Italians). sentence2:\ Slovenia has 3,000 inhabitants.}
\item[Original target:] \texttt{1}
\item[Processed target:] \texttt{not\_entailment}
\end{description}

\subsection{MNLI}
\begin{description}[leftmargin=0.5cm]
\item[Original input:] ~
\begin{description}[leftmargin=0.5cm]
  \item[Hypothesis:] \texttt{The St.\ Louis Cardinals have always won.}
  \item[Premise:] \texttt{yeah well losing is i mean i'm i'm originally from Saint Louis and Saint Louis Cardinals when they were there were uh a mostly a losing team but}
\end{description}
\item[Processed input:] \texttt{mnli hypothesis:\ The St. Louis Cardinals have always won. premise:\ yeah well losing is i mean i'm i'm originally from Saint Louis and Saint Louis Cardinals when they were there were uh a mostly a losing team but}
\item[Original target:] \texttt{2}
\item[Processed target:] \texttt{contradiction}
\end{description}

\subsection{MRPC}
\begin{description}[leftmargin=0.5cm]
\item[Original input:] ~
\begin{description}[leftmargin=0.5cm]
  \item[Sentence 1:] \texttt{We acted because we saw the existing evidence in a new light , through the prism of our experience on 11 September , " Rumsfeld said .}
  \item[Sentence 2:] \texttt{Rather , the US acted because the administration saw " existing evidence in a new light , through the prism of our experience on September 11 " .}
\end{description}
\item[Processed input:] \texttt{mrpc sentence1:\ We acted because we saw the existing evidence in a new light , through the prism of our experience on 11 September , " Rumsfeld said . sentence2:\ Rather , the US acted because the administration saw " existing evidence in a new light , through the prism of our experience on September 11 " .}
\item[Original target:] \texttt{1}
\item[Processed target:] \texttt{equivalent}
\end{description}

\subsection{QNLI}
\begin{description}[leftmargin=0.5cm]
\item[Original input:] ~
\begin{description}[leftmargin=0.5cm]
  \item[Question:] \texttt{Where did Jebe die?}
  \item[Sentence:] \texttt{Genghis Khan recalled Subutai back to Mongolia soon afterwards, and Jebe died on the road back to Samarkand.}
\end{description}
\item[Processed input:] \texttt{qnli question:\ Where did Jebe die? sentence:\ Genghis Khan recalled Subutai back to Mongolia soon afterwards, and Jebe died on the road back to Samarkand.}
\item[Original target:] \texttt{0}
\item[Processed target:] \texttt{entailment}
\end{description}

\subsection{QQP}
\begin{description}[leftmargin=0.5cm]
\item[Original input:] ~
\begin{description}[leftmargin=0.5cm]
  \item[Question 1:] \texttt{What attributes would have made you highly desirable in ancient Rome?}
  \item[Question 2:] \texttt{How I GET OPPERTINUTY TO JOIN IT COMPANY AS A FRESHER?}
\end{description}
\item[Processed input:] \texttt{qqp question1:\ What attributes would have made you highly desirable in ancient Rome? question2:\ How I GET OPPERTINUTY TO JOIN IT COMPANY AS A FRESHER?}
\item[Original target:] \texttt{0}
\item[Processed target:] \texttt{not\_duplicate}
\end{description}

\subsection{SST2}
\begin{description}[leftmargin=0.5cm]
\item[Original input:] ~
\begin{description}[leftmargin=0.5cm]
  \item[Sentence:] \texttt{it confirms fincher 's status as a film maker who artfully bends technical know-how to the service of psychological insight . }
\end{description}
\item[Processed input:] \texttt{sst2 sentence:\ it confirms fincher 's status as a film maker who artfully bends technical know-how to the service of psychological insight . }
\item[Original target:] \texttt{1}
\item[Processed target:] \texttt{positive}
\end{description}

\subsection{STSB}
\begin{description}[leftmargin=0.5cm]
\item[Original input:] ~
\begin{description}[leftmargin=0.5cm]
  \item[Sentence 1:] \texttt{Representatives for Puretunes could not immediately be reached for comment Wednesday.}
  \item[Sentence 2:] \texttt{Puretunes representatives could not be located Thursday to comment on the suit.}
\end{description}
\item[Processed input:] \texttt{stsb sentence1:\ Representatives for Puretunes could not immediately be reached for comment Wednesday. sentence2:\ Puretunes representatives could not be located Thursday to comment on the suit.}
\item[Original target:] \texttt{3.25}
\item[Processed target:] \texttt{3.2}
\end{description}

\subsection{CB}
\begin{description}[leftmargin=0.5cm]
\item[Original input:] ~
\begin{description}[leftmargin=0.5cm]
  \item[Hypothesis:] \texttt{Valence was helping}
  \item[Premise:] \texttt{Valence the void-brain, Valence the virtuous valet. Why couldn't the figger choose his own portion of titanic anatomy to shaft? Did he think he was helping?}
\end{description}
\item[Processed input:] \texttt{cb hypothesis:\ Valence was helping premise:\ Valence the void-brain, Valence the virtuous valet. Why couldn't the figger choose his own portion of titanic anatomy to shaft? Did he think he was helping?}
\item[Original target:] \texttt{1}
\item[Processed target:] \texttt{contradiction}
\end{description}

\subsection{COPA}
\begin{description}[leftmargin=0.5cm]
\item[Original input:] ~
\begin{description}[leftmargin=0.5cm]
  \item[Question:] \texttt{effect}
  \item[Premise:] \texttt{Political violence broke out in the nation.}
  \item[Choice 1:] \texttt{Many citizens relocated to the capitol.}
  \item[Choice 2:] \texttt{Many citizens took refuge in other territories.}
\end{description}
\item[Processed input:] \texttt{copa choice1:\ Many citizens relocated to the capitol. choice2:\ Many citizens took refuge in other territories. premise:\ Political violence broke out in the nation. question:\ effect}
\item[Original target:] \texttt{1}
\item[Processed target:] \texttt{True}
\end{description}

\subsection{MultiRC}
\begin{description}[leftmargin=0.5cm]
\item[Original input:] ~
\begin{description}[leftmargin=0.5cm]
  \item[Answer:] \texttt{There was only pie to eat, rather than traditional breakfast foods}
  \item[Paragraph:] \texttt{<b>Sent 1:\ </b>Once upon a time, there was a squirrel named Joey.<br><b>Sent 2:\ </b>Joey loved to go outside and play with his cousin Jimmy.<br><b>Sent 3:\ </b>Joey and Jimmy played silly games together, and were always laughing.<br><b>Sent 4:\ </b>One day, Joey and Jimmy went swimming together at their Aunt Julie's pond.<br><b>Sent 5:\ </b>Joey woke up early in the morning to eat some food before they left.<br><b>Sent 6:\ </b>He couldn't find anything to eat except for pie!<br><b>Sent 7:\ </b>Usually, Joey would eat cereal, fruit (a pear), or oatmeal for breakfast.<br><b>Sent 8:\ </b>After he ate, he and Jimmy went to the pond.<br><b>Sent 9:\ </b>On their way there they saw their friend Jack Rabbit.<br><b>Sent 10:\ </b>They dove into the water and swam for several hours.<br><b>Sent 11:\ </b>The sun was out, but the breeze was cold.<br><b>Sent 12:\ </b>Joey and Jimmy got out of the water and started walking home.<br><b>Sent 13:\ </b>Their fur was wet, and the breeze chilled them.<br><b>Sent 14:\ </b>When they got home, they dried off, and Jimmy put on his favorite purple shirt.<br><b>Sent 15:\ </b>Joey put on a blue shirt with red and green dots.<br><b>Sent 16:\ </b>The two squirrels ate some food that Joey's mom, Jasmine, made and went off to bed.<br>}
  \item[Question:] \texttt{Why was Joey surprised the morning he woke up for breakfast?}
\end{description}
\item[Processed input:] \texttt{multirc question:\ Why was Joey surprised the morning he woke up for breakfast? answer:\ There was only pie to eat, rather than traditional breakfast foods paragraph:\ <b>Sent 1:\ </b>Once upon a time, there was a squirrel named Joey.<br><b>Sent 2:\ </b>Joey loved to go outside and play with his cousin Jimmy.<br><b>Sent 3:\ </b>Joey and Jimmy played silly games together, and were always laughing.<br><b>Sent 4:\ </b>One day, Joey and Jimmy went swimming together at their Aunt Julie's pond.<br><b>Sent 5:\ </b>Joey woke up early in the morning to eat some food before they left.<br><b>Sent 6:\ </b>He couldn't find anything to eat except for pie!<br><b>Sent 7:\ </b>Usually, Joey would eat cereal, fruit (a pear), or oatmeal for breakfast.<br><b>Sent 8:\ </b>After he ate, he and Jimmy went to the pond.<br><b>Sent 9:\ </b>On their way there they saw their friend Jack Rabbit.<br><b>Sent 10:\ </b>They dove into the water and swam for several hours.<br><b>Sent 11:\ </b>The sun was out, but the breeze was cold.<br><b>Sent 12:\ </b>Joey and Jimmy got out of the water and started walking home.<br><b>Sent 13:\ </b>Their fur was wet, and the breeze chilled them.<br><b>Sent 14:\ </b>When they got home, they dried off, and Jimmy put on his favorite purple shirt.<br><b>Sent 15:\ </b>Joey put on a blue shirt with red and green dots.<br><b>Sent 16:\ </b>The two squirrels ate some food that Joey's mom, Jasmine, made and went off to bed.<br>}
\item[Original target:] \texttt{1}
\item[Processed target:] \texttt{True}
\end{description}

\subsection{WiC}
\begin{description}[leftmargin=0.5cm]
\item[Original input:] ~
\begin{description}[leftmargin=0.5cm]
  \item[POS:] \texttt{N}
  \item[Sentence 1:] \texttt{It was the deliberation of his act that was insulting .}
  \item[Sentence 2:] \texttt{The deliberations of the jury .}
  \item[Word:] \texttt{deliberation}
\end{description}
\item[Processed input:] \texttt{wic pos:\ N sentence1:\ It was the deliberation of his act that was insulting . sentence2:\ The deliberations of the jury . word:\ deliberation}
\item[Original target:] \texttt{0}
\item[Processed target:] \texttt{False}
\end{description}

\subsection{WSC and DPR}
\begin{description}[leftmargin=0.5cm]
\item[Original input:] ~
\begin{description}[leftmargin=0.5cm]
  \item[Span 2 text:] \texttt{it}
  \item[Span 1 text:] \texttt{stable}
  \item[Span 2 index:] \texttt{20}
  \item[Span 1 index:] \texttt{1}
  \item[Text:] \texttt{The stable was very roomy, with four good stalls; a large swinging window opened into the yard , which made it pleasant and airy.}
\end{description}
\item[Processed input:] \texttt{wsc: The stable was very roomy, with four good stalls; a large swinging window opened into the yard , which made *it* pleasant and airy.}
\item[Original target:] \texttt{1}
\item[Processed target:] \texttt{stable}
\end{description}

\subsection{CNN/Daily Mail}
\begin{description}[leftmargin=0.5cm]
\item[Original input:] \texttt{marouane fellaini and adnan januzaj continue to show the world they are not just teammates but also best mates. the manchester united and belgium duo both posted pictures of themselves out at a restaurant on monday night ahead of their game against newcastle on wednesday . januzaj poses in the middle of fellaini and a friend looking like somebody who failed to receive the memo about it being a jackson 5 themed night. premier league duo adnan januzaj and marouane fellaini pose with a friend on the dance floor . manchester united and belgium duo fellaini and januzaj are good friends both on and off the pitch . manchester united ace fellaini runs over to the bench to celebrate his goal against qpr with friend januzaj . the disco effect in the background adds to the theory, but januzaj doesn’t seem to mind as they later pose on the dance floor with other friends. united haven’t had too many reasons to have a song and dance this season so it seems they may be hitting the discotheques as another form of release. however, victory against newcastle on wednesday would leave manager louis van gaal at least tapping his toes as they continue to fight for a champions league spot this season. januzaj and robin van persie join fellaini in celebrating in front of the manchester united fans at west brom . januzaj receives some words of wisdom from manchester united's dutch manager louis van gaal . januzaj and fellaini are joined by some friends as they take to the dance floor ahead of the newcastle game .}
\item[Processed input:] \texttt{summarize:\ marouane fellaini and adnan januzaj continue to show the world they are not just teammates but also best mates. the manchester united and belgium duo both posted pictures of themselves out at a restaurant on monday night ahead of their game against newcastle on wednesday . januzaj poses in the middle of fellaini and a friend looking like somebody who failed to receive the memo about it being a jackson 5 themed night. premier league duo adnan januzaj and marouane fellaini pose with a friend on the dance floor . manchester united and belgium duo fellaini and januzaj are good friends both on and off the pitch . manchester united ace fellaini runs over to the bench to celebrate his goal against qpr with friend januzaj . the disco effect in the background adds to the theory, but januzaj doesn’t seem to mind as they later pose on the dance floor with other friends. united haven’t had too many reasons to have a song and dance this season so it seems they may be hitting the discotheques as another form of release. however, victory against newcastle on wednesday would leave manager louis van gaal at least tapping his toes as they continue to fight for a champions league spot this season. januzaj and robin van persie join fellaini in celebrating in front of the manchester united fans at west brom . januzaj receives some words of wisdom from manchester united's dutch manager louis van gaal . januzaj and fellaini are joined by some friends as they take to the dance floor ahead of the newcastle game .}
\item[Original target:] \texttt{the belgian duo took to the dance floor on monday night with some friends . manchester united face newcastle in the premier league on wednesday  . red devils will be looking for just their second league away win in seven  . louis van gaal's side currently sit two points clear of liverpool in fourth   .}
\item[Processed target:] \texttt{the belgian duo took to the dance floor on monday night with some friends . manchester united face newcastle in the premier league on wednesday  . red devils will be looking for just their second league away win in seven  . louis van gaal's side currently sit two points clear of liverpool in fourth   .}
\end{description}

\subsection{SQuAD}
\begin{description}[leftmargin=0.5cm]
\item[Original input:] ~
\begin{description}[leftmargin=0.5cm]
  \item[Question:] \texttt{What does increased oxygen concentrations in the patient's lungs displace?}
  \item[Context:] \texttt{Hyperbaric (high-pressure) medicine uses special oxygen chambers to increase the partial pressure of O
2 around the patient and, when needed, the medical staff. Carbon monoxide poisoning, gas gangrene, and decompression sickness (the 'bends') are sometimes treated using these devices. Increased O
2 concentration in the lungs helps to displace carbon monoxide from the heme group of hemoglobin. Oxygen gas is poisonous to the anaerobic bacteria that cause gas gangrene, so increasing its partial pressure helps kill them. Decompression sickness occurs in divers who decompress too quickly after a dive, resulting in bubbles of inert gas, mostly nitrogen and helium, forming in their blood. Increasing the pressure of O
2 as soon as possible is part of the treatment.}
\end{description}
\item[Processed input:] \texttt{question:\ What does increased oxygen concentrations in the patient's lungs displace? context:\ Hyperbaric (high-pressure) medicine uses special oxygen chambers to increase the partial pressure of O
2 around the patient and, when needed, the medical staff. Carbon monoxide poisoning, gas gangrene, and decompression sickness (the 'bends') are sometimes treated using these devices. Increased O
2 concentration in the lungs helps to displace carbon monoxide from the heme group of hemoglobin. Oxygen gas is poisonous to the anaerobic bacteria that cause gas gangrene, so increasing its partial pressure helps kill them. Decompression sickness occurs in divers who decompress too quickly after a dive, resulting in bubbles of inert gas, mostly nitrogen and helium, forming in their blood. Increasing the pressure of O
2 as soon as possible is part of the treatment.}
\item[Original target:] \texttt{carbon monoxide}
\item[Processed target:] \texttt{carbon monoxide}
\end{description}

\subsection{WMT English to German}
\begin{description}[leftmargin=0.5cm]
\item[Original input:] \texttt{"Luigi often said to me that he never wanted the brothers to end up in court," she wrote.}
\item[Processed input:] \texttt{translate English to German:\ "Luigi often said to me that he never wanted the brothers to end up in court," she wrote.}
\item[Original target:] \texttt{"Luigi sagte oft zu mir, dass er nie wollte, dass die Brüder vor Gericht landen", schrieb sie.}
\item[Processed target:] \texttt{"Luigi sagte oft zu mir, dass er nie wollte, dass die Brüder vor Gericht landen", schrieb sie.}
\end{description}

\subsection{WMT English to French}
\begin{description}[leftmargin=0.5cm]
\item[Original input:] \texttt{This image section from an infrared recording by the Spitzer telescope shows a "family portrait" of countless generations of stars:\ the oldest stars are seen as blue dots, while more difficult to identify are the pink-coloured "new-borns" in the star delivery room.}
\item[Processed input:] \texttt{translate English to French:\ This image section from an infrared recording by the Spitzer telescope shows a "family portrait" of countless generations of stars:\ the oldest stars are seen as blue dots, while more difficult to identify are the pink-coloured "new-borns" in the star delivery room.}
\item[Original target:] \texttt{Ce détail d'une photographie infrarouge prise par le télescope Spitzer montre un "portrait de famille" des innombrables générations d'étoiles:\ les plus vieilles étoiles sont en bleu et les points roses, plus difficiles à identifier, sont les "nouveau-nés" dans la salle d'accouchement de l'univers.}
\item[Processed target:] \texttt{Ce détail d'une photographie infrarouge prise par le télescope Spitzer montre un "portrait de famille" des innombrables générations d'étoiles:\ les plus vieilles étoiles sont en bleu et les points roses, plus difficiles à identifier, sont les "nouveau-nés" dans la salle d'accouchement de l'univers.}
\end{description}

\subsection{WMT English to Romanian}
\begin{description}[leftmargin=0.5cm]
\item[Original input:] \texttt{Taco Bell said it plans to add 2,000 locations in the US by 2022.}
\item[Processed input:] \texttt{translate English to Romanian:\ Taco Bell said it plans to add 2,000 locations in the US by 2022.}
\item[Original target:] \texttt{Taco Bell a afirmat că, până în 2022, intenționează să deschidă 2000 de restaurante în SUA.}
\item[Processed target:] \texttt{Taco Bell a afirmat că, până în 2022, intenționează să deschidă 2000 de restaurante în SUA.}
\end{description}

\clearpage
%\KOMAoptions{paper=A2,paper=landscape,pagesize,DIV=16}
%\recalctypearea

\section{Scores on Every Task for All Experiments}
\label{sec:giant}

The following table lists the scores achieved on every task in the experiments described in \cref{sec:architectures,sec:objectives,sec:datasets,sec:transfer,sec:scaling}.

\clearpage
\eject \pdfpagewidth=58cm \pdfpageheight=39cm
\thispagestyle{empty}

\begin{table}[!ht]
\centering
\begin{minipage}{0.85\pdfpagewidth}
\footnotesize
\begin{tabular}{llccccccccccccccccccccccccccccccccc}
\toprule
& & \multicolumn{13}{c}{\textbf{GLUE}} & & & & & & \multicolumn{12}{c}{\textbf{SuperGLUE}} & \multicolumn{3}{c}{\textbf{WMT}} \\
& & Score & CoLA & SST-2 & MRPC & MRPC & STSB & STSB & QQP & QQP & MNLI\textsubscript{m} & MNLI\textsubscript{mm} & QNLI & RTE & \multicolumn{3}{c}{\textbf{CNN/DM}} & \multicolumn{2}{c}{\textbf{SQuAD}} & Score & BoolQ & CB & CB & COPA & MultiRC & MultiRC & ReCoRD & ReCoRD & RTE & WiC & WSC & EnDe & EnFr & EnRo \\
Table & Experiment & Average & MCC & Acc & F1 & Acc & PCC & SCC & F1 & Acc & Acc & Acc & Acc & Acc & R-1-F & R-2-F & R-L-F & EM & F1 & Average & Acc & F1 & Acc & Acc & F1 & EM & F1 & EM & Acc & Acc & Acc & BLEU & BLEU & BLEU \\
\cmidrule(r){1-2} \cmidrule(lr){3-15} \cmidrule(lr){16-18} \cmidrule(lr){19-20} \cmidrule(lr){21-32} \cmidrule(l){33-35}
\ref{tab:baseline} & \bsl Baseline average & $83.28$ & $53.84$ & $92.68$ & $92.07$ & $88.92$ & $88.02$ & $87.94$ & $88.67$ & $91.56$ & $84.24$ & $84.57$ & $90.48$ & $76.28$ & $41.33$ & $19.24$ & $38.77$ & $80.88$ & $88.81$ & $71.36$ & $76.62$ & $91.22$ & $91.96$ & $66.20$ & $66.13$ & $25.78$ & $69.05$ & $68.16$ & $75.34$ & $68.04$ & $78.56$ & $26.98$ & $39.82$ & $27.65$ \\
\ref{tab:baseline} & Baseline standard deviation & $0.235$ & $1.111$ & $0.569$ & $0.729$ & $1.019$ & $0.374$ & $0.418$ & $0.108$ & $0.070$ & $0.291$ & $0.231$ & $0.361$ & $1.393$ & $0.065$ & $0.065$ & $0.058$ & $0.343$ & $0.226$ & $0.416$ & $0.365$ & $3.237$ & $2.560$ & $2.741$ & $0.716$ & $1.011$ & $0.370$ & $0.379$ & $1.228$ & $0.850$ & $2.029$ & $0.112$ & $0.090$ & $0.108$ \\
\ref{tab:baseline} & No pre-training & $66.22$ & $12.29$ & $80.62$ & $81.42$ & $73.04$ & $72.58$ & $72.97$ & $81.94$ & $86.62$ & $68.02$ & $67.98$ & $75.69$ & $58.84$ & $39.19$ & $17.60$ & $36.69$ & $50.31$ & $61.97$ & $53.04$ & $65.38$ & $71.61$ & $76.79$ & $62.00$ & $59.10$ & $0.84$ & $20.33$ & $17.95$ & $54.15$ & $54.08$ & $65.38$ & $25.86$ & $39.77$ & $24.04$ \\
\cmidrule(r){1-2} \cmidrule(lr){3-15} \cmidrule(lr){16-18} \cmidrule(lr){19-20} \cmidrule(lr){21-32} \cmidrule(l){33-35}
\ref{tab:architectures_results} & \bsl Enc/dec, denoising & $83.28$ & $53.84$ & $92.68$ & $92.07$ & $88.92$ & $88.02$ & $87.94$ & $88.67$ & $91.56$ & $84.24$ & $84.57$ & $90.48$ & $76.28$ & $41.33$ & $19.24$ & $38.77$ & $80.88$ & $88.81$ & $71.36$ & $76.62$ & $91.22$ & $91.96$ & $66.20$ & $66.13$ & $25.78$ & $69.05$ & $68.16$ & $75.34$ & $68.04$ & $78.56$ & $26.98$ & $39.82$ & $27.65$ \\
\ref{tab:architectures_results} & Enc/dec, shared, denoising & $82.81$ & $55.24$ & $91.86$ & $91.58$ & $88.24$ & $87.43$ & $87.58$ & $88.69$ & $91.60$ & $83.88$ & $84.01$ & $90.23$ & $73.65$ & $41.11$ & $18.78$ & $38.48$ & $80.63$ & $88.49$ & $70.73$ & $77.13$ & $95.04$ & $96.43$ & $65.00$ & $66.16$ & $22.98$ & $68.95$ & $68.09$ & $70.76$ & $68.18$ & $75.96$ & $26.72$ & $39.03$ & $27.46$ \\
\ref{tab:architectures_results} & Enc/dec, 6 layers, denoising & $80.88$ & $46.26$ & $92.09$ & $91.51$ & $87.99$ & $87.01$ & $86.76$ & $87.93$ & $90.97$ & $82.20$ & $82.41$ & $88.83$ & $71.48$ & $40.83$ & $18.97$ & $38.31$ & $77.59$ & $86.07$ & $68.42$ & $73.79$ & $91.70$ & $92.86$ & $67.00$ & $61.02$ & $19.62$ & $61.26$ & $60.33$ & $72.20$ & $65.99$ & $75.00$ & $26.38$ & $38.40$ & $26.95$ \\
\ref{tab:architectures_results} & Language model, denoising & $74.70$ & $24.50$ & $90.60$ & $86.08$ & $78.92$ & $85.22$ & $85.42$ & $85.40$ & $88.99$ & $76.72$ & $77.05$ & $86.02$ & $64.62$ & $39.49$ & $17.93$ & $36.91$ & $61.14$ & $71.37$ & $55.02$ & $65.47$ & $60.08$ & $71.43$ & $58.00$ & $43.03$ & $2.94$ & $53.35$ & $52.31$ & $53.07$ & $58.62$ & $63.46$ & $25.09$ & $35.28$ & $25.86$ \\
\ref{tab:architectures_results} & Prefix LM, denoising & $81.82$ & $49.99$ & $92.43$ & $91.43$ & $88.24$ & $87.20$ & $86.98$ & $88.41$ & $91.39$ & $82.32$ & $82.93$ & $88.71$ & $74.01$ & $40.46$ & $18.61$ & $37.90$ & $78.94$ & $87.31$ & $68.11$ & $75.50$ & $93.37$ & $91.07$ & $60.00$ & $63.43$ & $21.20$ & $65.03$ & $64.11$ & $71.48$ & $65.67$ & $73.08$ & $26.43$ & $37.98$ & $27.39$ \\
\ref{tab:architectures_results} & Enc/dec, LM & $79.56$ & $42.03$ & $91.86$ & $91.64$ & $88.24$ & $87.13$ & $87.00$ & $88.21$ & $91.15$ & $81.68$ & $81.66$ & $88.54$ & $65.70$ & $40.67$ & $18.59$ & $38.13$ & $76.02$ & $84.85$ & $64.29$ & $72.23$ & $85.74$ & $89.29$ & $57.00$ & $60.53$ & $16.26$ & $59.28$ & $58.30$ & $65.34$ & $64.89$ & $70.19$ & $26.27$ & $39.17$ & $26.86$ \\
\ref{tab:architectures_results} & Enc/dec, shared, LM & $79.60$ & $44.83$ & $92.09$ & $90.20$ & $85.78$ & $86.03$ & $85.87$ & $87.77$ & $91.02$ & $81.74$ & $82.29$ & $89.16$ & $65.34$ & $40.16$ & $18.13$ & $37.59$ & $76.35$ & $84.86$ & $63.50$ & $70.49$ & $91.41$ & $87.50$ & $55.00$ & $60.21$ & $16.89$ & $57.83$ & $56.73$ & $63.54$ & $63.48$ & $70.19$ & $26.62$ & $39.17$ & $27.05$ \\
\ref{tab:architectures_results} & Enc/dec, 6 layers, LM & $78.67$ & $38.72$ & $91.40$ & $90.40$ & $86.52$ & $86.82$ & $86.49$ & $87.87$ & $91.03$ & $80.99$ & $80.92$ & $88.05$ & $65.70$ & $40.29$ & $18.26$ & $37.70$ & $75.32$ & $84.06$ & $64.06$ & $71.38$ & $85.25$ & $89.29$ & $60.00$ & $57.56$ & $16.79$ & $55.22$ & $54.30$ & $66.79$ & $63.95$ & $71.15$ & $26.13$ & $38.42$ & $26.89$ \\
\ref{tab:architectures_results} & Language model, LM & $73.78$ & $28.53$ & $89.79$ & $85.23$ & $78.68$ & $84.22$ & $84.00$ & $84.88$ & $88.70$ & $74.94$ & $75.77$ & $84.84$ & $58.84$ & $38.97$ & $17.54$ & $36.37$ & $53.81$ & $64.55$ & $56.51$ & $64.22$ & $59.92$ & $71.43$ & $64.00$ & $53.04$ & $1.05$ & $46.81$ & $45.78$ & $58.84$ & $56.74$ & $69.23$ & $25.23$ & $34.31$ & $25.38$ \\
\ref{tab:architectures_results} & Prefix LM, LM & $79.68$ & $41.26$ & $92.09$ & $90.11$ & $86.27$ & $86.82$ & $86.32$ & $88.35$ & $91.35$ & $81.71$ & $82.02$ & $89.04$ & $68.59$ & $39.66$ & $17.84$ & $37.13$ & $76.87$ & $85.39$ & $64.86$ & $71.47$ & $93.37$ & $91.07$ & $57.00$ & $58.67$ & $16.89$ & $59.25$ & $58.16$ & $64.26$ & $66.30$ & $71.15$ & $26.28$ & $37.51$ & $26.76$ \\
\cmidrule(r){1-2} \cmidrule(lr){3-15} \cmidrule(lr){16-18} \cmidrule(lr){19-20} \cmidrule(lr){21-32} \cmidrule(l){33-35}
\ref{tab:objectives_highlevel} & Language modeling with prefix & $80.69$ & $44.22$ & $93.00$ & $91.68$ & $88.48$ & $87.20$ & $87.18$ & $88.39$ & $91.41$ & $82.66$ & $83.09$ & $89.29$ & $68.95$ & $40.71$ & $18.94$ & $38.15$ & $77.99$ & $86.43$ & $65.27$ & $73.55$ & $83.95$ & $87.50$ & $55.00$ & $59.65$ & $18.89$ & $61.76$ & $60.76$ & $68.59$ & $65.67$ & $73.08$ & $26.86$ & $39.73$ & $27.49$ \\
\ref{tab:objectives_highlevel} & BERT-style \citep{devlin2018bert} & $82.96$ & $52.49$ & $92.55$ & $92.79$ & $89.95$ & $87.68$ & $87.66$ & $88.47$ & $91.44$ & $83.60$ & $84.05$ & $90.33$ & $75.45$ & $41.27$ & $19.17$ & $38.72$ & $80.65$ & $88.24$ & $69.85$ & $76.48$ & $94.37$ & $94.64$ & $61.00$ & $63.29$ & $25.08$ & $66.76$ & $65.85$ & $72.20$ & $69.12$ & $75.00$ & $26.78$ & $40.03$ & $27.41$ \\
\ref{tab:objectives_highlevel} & Deshuffling & $73.17$ & $22.82$ & $87.16$ & $86.88$ & $81.13$ & $84.03$ & $83.82$ & $86.38$ & $89.90$ & $76.30$ & $76.34$ & $84.18$ & $58.84$ & $40.75$ & $18.59$ & $38.10$ & $67.61$ & $76.76$ & $58.47$ & $69.17$ & $63.70$ & $78.57$ & $56.00$ & $59.85$ & $12.70$ & $45.52$ & $44.36$ & $57.04$ & $64.89$ & $68.27$ & $26.11$ & $39.30$ & $25.62$ \\
\cmidrule(r){1-2} \cmidrule(lr){3-15} \cmidrule(lr){16-18} \cmidrule(lr){19-20} \cmidrule(lr){21-32} \cmidrule(l){33-35}
\ref{tab:objectives_bert} & BERT-style \citep{devlin2018bert} & $82.96$ & $52.49$ & $92.55$ & $92.79$ & $89.95$ & $87.68$ & $87.66$ & $88.47$ & $91.44$ & $83.60$ & $84.05$ & $90.33$ & $75.45$ & $41.27$ & $19.17$ & $38.72$ & $80.65$ & $88.24$ & $69.85$ & $76.48$ & $94.37$ & $94.64$ & $61.00$ & $63.29$ & $25.08$ & $66.76$ & $65.85$ & $72.20$ & $69.12$ & $75.00$ & $26.78$ & $40.03$ & $27.41$ \\
\ref{tab:objectives_bert} & MASS-style \citep{song2019mass} & $82.32$ & $47.01$ & $91.63$ & $92.53$ & $89.71$ & $88.21$ & $88.18$ & $88.58$ & $91.44$ & $82.96$ & $83.67$ & $90.02$ & $77.26$ & $41.16$ & $19.16$ & $38.55$ & $80.10$ & $88.07$ & $69.28$ & $75.08$ & $84.98$ & $89.29$ & $63.00$ & $64.46$ & $23.50$ & $66.71$ & $65.91$ & $72.20$ & $67.71$ & $78.85$ & $26.79$ & $39.89$ & $27.55$ \\
\ref{tab:objectives_bert} & \bsl Replace corrupted spans & $83.28$ & $53.84$ & $92.68$ & $92.07$ & $88.92$ & $88.02$ & $87.94$ & $88.67$ & $91.56$ & $84.24$ & $84.57$ & $90.48$ & $76.28$ & $41.33$ & $19.24$ & $38.77$ & $80.88$ & $88.81$ & $71.36$ & $76.62$ & $91.22$ & $91.96$ & $66.20$ & $66.13$ & $25.78$ & $69.05$ & $68.16$ & $75.34$ & $68.04$ & $78.56$ & $26.98$ & $39.82$ & $27.65$ \\
\ref{tab:objectives_bert} & Drop corrupted tokens & $84.44$ & $60.04$ & $92.89$ & $92.79$ & $89.95$ & $87.28$ & $86.85$ & $88.56$ & $91.54$ & $83.94$ & $83.92$ & $90.74$ & $79.42$ & $41.27$ & $19.31$ & $38.70$ & $80.52$ & $88.28$ & $68.67$ & $75.90$ & $96.02$ & $94.64$ & $56.00$ & $65.06$ & $23.92$ & $65.54$ & $64.60$ & $71.12$ & $67.40$ & $74.04$ & $27.07$ & $39.76$ & $27.82$ \\
\cmidrule(r){1-2} \cmidrule(lr){3-15} \cmidrule(lr){16-18} \cmidrule(lr){19-20} \cmidrule(lr){21-32} \cmidrule(l){33-35}
\ref{tab:objectives_rate} & Corruption rate = $10\%$ & $82.82$ & $52.71$ & $92.09$ & $91.55$ & $88.24$ & $88.19$ & $88.15$ & $88.47$ & $91.40$ & $83.50$ & $84.51$ & $90.33$ & $75.45$ & $41.05$ & $19.00$ & $38.53$ & $80.38$ & $88.36$ & $69.55$ & $74.98$ & $92.37$ & $92.86$ & $62.00$ & $66.04$ & $24.66$ & $67.93$ & $67.09$ & $70.76$ & $67.24$ & $75.96$ & $26.87$ & $39.28$ & $27.44$ \\
\ref{tab:objectives_rate} & \bsl Corruption rate = $15\%$ & $83.28$ & $53.84$ & $92.68$ & $92.07$ & $88.92$ & $88.02$ & $87.94$ & $88.67$ & $91.56$ & $84.24$ & $84.57$ & $90.48$ & $76.28$ & $41.33$ & $19.24$ & $38.77$ & $80.88$ & $88.81$ & $71.36$ & $76.62$ & $91.22$ & $91.96$ & $66.20$ & $66.13$ & $25.78$ & $69.05$ & $68.16$ & $75.34$ & $68.04$ & $78.56$ & $26.98$ & $39.82$ & $27.65$ \\
\ref{tab:objectives_rate} & Corruption rate = $25\%$ & $83.00$ & $53.47$ & $93.00$ & $92.44$ & $89.46$ & $87.36$ & $87.36$ & $88.68$ & $91.53$ & $84.44$ & $84.15$ & $90.77$ & $74.01$ & $41.69$ & $19.54$ & $39.14$ & $80.96$ & $88.61$ & $70.48$ & $76.39$ & $93.02$ & $92.86$ & $68.00$ & $65.46$ & $24.66$ & $68.20$ & $67.39$ & $73.65$ & $67.87$ & $72.12$ & $27.04$ & $39.83$ & $27.47$ \\
\ref{tab:objectives_rate} & Corruption rate = $50\%$ & $81.27$ & $46.26$ & $91.63$ & $91.11$ & $87.99$ & $87.87$ & $87.64$ & $88.70$ & $91.57$ & $83.64$ & $84.10$ & $90.24$ & $70.76$ & $41.51$ & $19.32$ & $38.89$ & $79.80$ & $87.76$ & $70.33$ & $75.02$ & $93.05$ & $92.86$ & $68.00$ & $62.97$ & $24.13$ & $64.94$ & $64.13$ & $72.20$ & $68.50$ & $77.88$ & $27.01$ & $39.90$ & $27.49$ \\
\cmidrule(r){1-2} \cmidrule(lr){3-15} \cmidrule(lr){16-18} \cmidrule(lr){19-20} \cmidrule(lr){21-32} \cmidrule(l){33-35}
\ref{tab:objectives_span} & \bsl Baseline (i.i.d.) & $83.28$ & $53.84$ & $92.68$ & $92.07$ & $88.92$ & $88.02$ & $87.94$ & $88.67$ & $91.56$ & $84.24$ & $84.57$ & $90.48$ & $76.28$ & $41.33$ & $19.24$ & $38.77$ & $80.88$ & $88.81$ & $71.36$ & $76.62$ & $91.22$ & $91.96$ & $66.20$ & $66.13$ & $25.78$ & $69.05$ & $68.16$ & $75.34$ & $68.04$ & $78.56$ & $26.98$ & $39.82$ & $27.65$ \\
\ref{tab:objectives_span}& Average span length = $2$ & $83.54$ & $53.82$ & $92.20$ & $93.05$ & $90.44$ & $87.85$ & $87.71$ & $88.42$ & $91.40$ & $84.28$ & $84.46$ & $90.88$ & $77.62$ & $41.23$ & $19.39$ & $38.69$ & $82.09$ & $89.69$ & $72.20$ & $77.06$ & $90.43$ & $91.07$ & $70.00$ & $66.28$ & $26.13$ & $71.34$ & $70.61$ & $75.45$ & $68.34$ & $78.85$ & $26.76$ & $39.99$ & $27.63$ \\
\ref{tab:objectives_span}& Average span length = $3$ & $83.49$ & $53.90$ & $92.43$ & $92.25$ & $89.46$ & $87.49$ & $87.53$ & $88.72$ & $91.51$ & $84.85$ & $84.84$ & $90.99$ & $77.26$ & $41.50$ & $19.62$ & $38.94$ & $81.84$ & $89.66$ & $72.53$ & $76.85$ & $94.37$ & $94.64$ & $70.00$ & $67.64$ & $28.75$ & $70.84$ & $69.90$ & $74.73$ & $67.71$ & $77.88$ & $26.86$ & $39.65$ & $27.62$ \\
\ref{tab:objectives_span}& Average span length = $5$ & $83.40$ & $52.12$ & $93.12$ & $92.63$ & $89.71$ & $88.70$ & $88.47$ & $88.84$ & $91.64$ & $84.32$ & $84.29$ & $90.79$ & $76.90$ & $41.39$ & $19.24$ & $38.82$ & $82.05$ & $89.79$ & $72.23$ & $77.06$ & $83.06$ & $89.29$ & $69.00$ & $68.16$ & $30.12$ & $71.36$ & $70.53$ & $75.81$ & $69.91$ & $79.81$ & $26.88$ & $39.40$ & $27.53$ \\
\ref{tab:objectives_span}& Average span length = $10$ & $82.85$ & $50.11$ & $92.09$ & $91.95$ & $88.97$ & $88.45$ & $88.22$ & $88.86$ & $91.63$ & $84.34$ & $84.28$ & $91.07$ & $76.17$ & $41.38$ & $19.33$ & $38.80$ & $81.84$ & $89.39$ & $70.44$ & $76.45$ & $87.40$ & $89.29$ & $65.00$ & $66.87$ & $29.59$ & $69.82$ & $68.94$ & $72.56$ & $67.55$ & $75.96$ & $26.79$ & $39.49$ & $27.69$ \\
\cmidrule(r){1-2} \cmidrule(lr){3-15} \cmidrule(lr){16-18} \cmidrule(lr){19-20} \cmidrule(lr){21-32} \cmidrule(l){33-35}
\ref{tab:datasets} & \bsl C4 & $83.28$ & $53.84$ & $92.68$ & $92.07$ & $88.92$ & $88.02$ & $87.94$ & $88.67$ & $91.56$ & $84.24$ & $84.57$ & $90.48$ & $76.28$ & $41.33$ & $19.24$ & $38.77$ & $80.88$ & $88.81$ & $71.36$ & $76.62$ & $91.22$ & $91.96$ & $66.20$ & $66.13$ & $25.78$ & $69.05$ & $68.16$ & $75.34$ & $68.04$ & $78.56$ & $26.98$ & $39.82$ & $27.65$ \\
\ref{tab:datasets} & C4, unfiltered & $81.46$ & $48.01$ & $91.63$ & $92.72$ & $89.95$ & $87.79$ & $87.60$ & $88.31$ & $91.27$ & $82.30$ & $82.34$ & $88.71$ & $72.20$ & $41.09$ & $19.14$ & $38.54$ & $78.78$ & $87.04$ & $68.04$ & $75.75$ & $89.17$ & $91.07$ & $62.00$ & $65.52$ & $25.60$ & $62.42$ & $61.58$ & $69.68$ & $67.08$ & $72.12$ & $26.55$ & $39.34$ & $27.21$ \\
\ref{tab:datasets} & RealNews-like & $83.83$ & $56.55$ & $92.66$ & $92.06$ & $88.97$ & $87.71$ & $87.37$ & $88.51$ & $91.49$ & $84.35$ & $84.46$ & $90.61$ & $78.34$ & $41.38$ & $19.23$ & $38.84$ & $80.39$ & $88.50$ & $72.38$ & $77.00$ & $93.09$ & $94.64$ & $66.00$ & $65.92$ & $23.82$ & $74.56$ & $73.72$ & $75.81$ & $66.61$ & $80.77$ & $26.75$ & $39.90$ & $27.48$ \\
\ref{tab:datasets} & WebText-like & $84.03$ & $56.38$ & $93.12$ & $92.31$ & $89.22$ & $88.69$ & $88.68$ & $88.65$ & $91.56$ & $84.70$ & $84.84$ & $90.83$ & $77.62$ & $41.23$ & $19.31$ & $38.70$ & $81.42$ & $89.15$ & $71.40$ & $76.88$ & $83.08$ & $89.29$ & $66.00$ & $64.10$ & $24.24$ & $72.24$ & $71.36$ & $75.45$ & $68.03$ & $82.69$ & $26.80$ & $39.74$ & $27.59$ \\
\ref{tab:datasets} & Wikipedia & $81.85$ & $45.53$ & $92.32$ & $91.67$ & $88.24$ & $85.62$ & $86.40$ & $88.37$ & $91.34$ & $82.61$ & $83.25$ & $90.96$ & $77.26$ & $41.39$ & $19.31$ & $38.81$ & $81.29$ & $89.18$ & $68.01$ & $76.12$ & $56.03$ & $80.36$ & $67.00$ & $65.01$ & $25.92$ & $69.03$ & $68.06$ & $74.73$ & $67.08$ & $76.92$ & $26.94$ & $39.69$ & $27.67$ \\
\ref{tab:datasets} & Wikipedia + TBC & $83.65$ & $55.53$ & $92.78$ & $92.41$ & $89.22$ & $86.67$ & $86.27$ & $89.47$ & $92.29$ & $84.38$ & $83.45$ & $91.94$ & $76.90$ & $41.22$ & $19.28$ & $38.67$ & $82.08$ & $89.70$ & $73.24$ & $76.22$ & $95.40$ & $92.86$ & $69.00$ & $51.59$ & $50.93$ & $69.53$ & $68.51$ & $77.62$ & $66.93$ & $81.73$ & $26.77$ & $39.63$ & $27.57$ \\
\cmidrule(r){1-2} \cmidrule(lr){3-15} \cmidrule(lr){16-18} \cmidrule(lr){19-20} \cmidrule(lr){21-32} \cmidrule(l){33-35}
\ref{tab:datasets_take} & \bsl Full data set & $83.28$ & $53.84$ & $92.68$ & $92.07$ & $88.92$ & $88.02$ & $87.94$ & $88.67$ & $91.56$ & $84.24$ & $84.57$ & $90.48$ & $76.28$ & $41.33$ & $19.24$ & $38.77$ & $80.88$ & $88.81$ & $71.36$ & $76.62$ & $91.22$ & $91.96$ & $66.20$ & $66.13$ & $25.78$ & $69.05$ & $68.16$ & $75.34$ & $68.04$ & $78.56$ & $26.98$ & $39.82$ & $27.65$ \\
\ref{tab:datasets_take} & $2^{29}$ ($64$ repeats) & $82.87$ & $53.82$ & $92.78$ & $91.79$ & $88.73$ & $87.56$ & $87.58$ & $88.73$ & $91.54$ & $84.07$ & $84.21$ & $90.59$ & $73.65$ & $41.18$ & $19.19$ & $38.67$ & $80.97$ & $88.90$ & $72.03$ & $76.76$ & $92.96$ & $92.86$ & $66.00$ & $65.11$ & $26.76$ & $69.35$ & $68.49$ & $75.81$ & $67.24$ & $82.69$ & $26.83$ & $39.74$ & $27.63$ \\
\ref{tab:datasets_take} & $2^{27}$ ($256$ repeats) & $82.62$ & $50.60$ & $92.32$ & $92.07$ & $88.73$ & $87.83$ & $87.60$ & $88.65$ & $91.54$ & $83.43$ & $84.37$ & $90.12$ & $75.81$ & $41.24$ & $19.20$ & $38.70$ & $79.78$ & $87.63$ & $69.97$ & $75.29$ & $93.42$ & $91.07$ & $63.00$ & $61.82$ & $23.61$ & $66.27$ & $65.39$ & $73.65$ & $66.30$ & $80.77$ & $27.02$ & $39.71$ & $27.33$ \\
\ref{tab:datasets_take} & $2^{25}$ ($1{,}024$ repeats) & $79.55$ & $43.84$ & $91.28$ & $89.32$ & $85.05$ & $85.92$ & $85.74$ & $88.05$ & $91.09$ & $81.29$ & $81.72$ & $87.90$ & $69.31$ & $40.66$ & $18.57$ & $38.13$ & $76.27$ & $84.58$ & $64.76$ & $72.63$ & $83.97$ & $82.14$ & $64.00$ & $59.39$ & $17.94$ & $56.94$ & $56.04$ & $64.98$ & $65.20$ & $73.08$ & $26.38$ & $39.56$ & $26.80$ \\
\ref{tab:datasets_take} & $2^{23}$ ($4{,}096$ repeats) & $76.34$ & $32.68$ & $89.45$ & $89.84$ & $86.03$ & $83.49$ & $83.42$ & $87.18$ & $90.61$ & $77.80$ & $78.69$ & $85.47$ & $64.62$ & $40.16$ & $18.33$ & $37.66$ & $70.92$ & $80.20$ & $59.29$ & $69.85$ & $73.48$ & $73.21$ & $56.00$ & $57.66$ & $14.38$ & $46.69$ & $45.79$ & $59.57$ & $65.05$ & $68.27$ & $26.37$ & $38.84$ & $25.81$ \\
\cmidrule(r){1-2} \cmidrule(lr){3-15} \cmidrule(lr){16-18} \cmidrule(lr){19-20} \cmidrule(lr){21-32} \cmidrule(l){33-35}
\ref{tab:finetuning} & \bsl All parameters & $83.28$ & $53.84$ & $92.68$ & $92.07$ & $88.92$ & $88.02$ & $87.94$ & $88.67$ & $91.56$ & $84.24$ & $84.57$ & $90.48$ & $76.28$ & $41.33$ & $19.24$ & $38.77$ & $80.88$ & $88.81$ & $71.36$ & $76.62$ & $91.22$ & $91.96$ & $66.20$ & $66.13$ & $25.78$ & $69.05$ & $68.16$ & $75.34$ & $68.04$ & $78.56$ & $26.98$ & $39.82$ & $27.65$ \\
\ref{tab:finetuning} & Adapter layers, $d=32$ & $80.52$ & $45.33$ & $91.63$ & $90.59$ & $86.76$ & $88.38$ & $88.06$ & $86.99$ & $90.26$ & $83.63$ & $83.94$ & $90.72$ & $67.15$ & $34.50$ & $15.08$ & $32.15$ & $79.32$ & $87.70$ & $60.40$ & $65.32$ & $50.87$ & $73.21$ & $52.00$ & $58.61$ & $19.41$ & $65.50$ & $64.58$ & $62.09$ & $64.58$ & $73.08$ & $13.84$ & $17.88$ & $15.54$ \\
\ref{tab:finetuning} &  Adapter layers, $d=128$ & $81.51$ & $45.35$ & $92.89$ & $91.49$ & $88.24$ & $87.73$ & $87.65$ & $87.73$ & $90.93$ & $83.64$ & $84.09$ & $90.52$ & $72.56$ & $36.71$ & $16.62$ & $34.37$ & $79.47$ & $87.61$ & $63.03$ & $69.20$ & $52.21$ & $75.00$ & $56.00$ & $61.08$ & $18.05$ & $67.94$ & $66.97$ & $68.59$ & $66.77$ & $73.08$ & $19.83$ & $27.50$ & $22.63$ \\
\ref{tab:finetuning} & Adapter layers, $d=512$ & $81.54$ & $44.25$ & $93.35$ & $91.00$ & $87.25$ & $88.74$ & $88.44$ & $88.02$ & $91.15$ & $83.08$ & $83.80$ & $89.62$ & $74.37$ & $38.63$ & $17.78$ & $36.25$ & $79.18$ & $87.32$ & $64.30$ & $73.18$ & $59.86$ & $71.43$ & $56.00$ & $62.94$ & $18.57$ & $66.56$ & $65.74$ & $70.76$ & $67.87$ & $74.04$ & $23.45$ & $33.98$ & $25.81$ \\
\ref{tab:finetuning} & Adapter layers, $d=2048$ & $82.62$ & $49.86$ & $92.55$ & $91.30$ & $87.99$ & $88.46$ & $88.35$ & $88.36$ & $91.40$ & $83.63$ & $83.18$ & $90.66$ & $76.53$ & $39.44$ & $18.30$ & $37.06$ & $79.40$ & $87.36$ & $68.61$ & $74.53$ & $88.00$ & $91.07$ & $58.00$ & $61.10$ & $18.89$ & $66.73$ & $66.06$ & $73.29$ & $71.16$ & $75.96$ & $25.64$ & $36.92$ & $26.93$ \\
\ref{tab:finetuning} & Gradual Unfreezing & $82.50$ & $51.74$ & $91.97$ & $92.61$ & $89.71$ & $87.27$ & $86.90$ & $88.26$ & $91.35$ & $83.42$ & $83.49$ & $89.71$ & $75.09$ & $40.88$ & $18.95$ & $38.40$ & $79.17$ & $87.30$ & $70.79$ & $75.51$ & $93.09$ & $94.64$ & $70.00$ & $62.03$ & $21.51$ & $65.69$ & $64.79$ & $72.92$ & $69.12$ & $77.89$ & $26.71$ & $39.02$ & $26.93$ \\
\cmidrule(r){1-2} \cmidrule(lr){3-15} \cmidrule(lr){16-18} \cmidrule(lr){19-20} \cmidrule(lr){21-32} \cmidrule(l){33-35}
\ref{tab:multitask} & \bsl Baseline (pre-train/fine-tune) & $83.28$ & $53.84$ & $92.68$ & $92.07$ & $88.92$ & $88.02$ & $87.94$ & $88.67$ & $91.56$ & $84.24$ & $84.57$ & $90.48$ & $76.28$ & $41.33$ & $19.24$ & $38.77$ & $80.88$ & $88.81$ & $71.36$ & $76.62$ & $91.22$ & $91.96$ & $66.20$ & $66.13$ & $25.78$ & $69.05$ & $68.16$ & $75.34$ & $68.04$ & $78.56$ & $26.98$ & $39.82$ & $27.65$ \\
\ref{tab:multitask} & Equal & $76.13$ & $39.47$ & $90.94$ & $82.90$ & $75.74$ & $78.83$ & $78.44$ & $86.45$ & $89.71$ & $82.08$ & $82.92$ & $90.13$ & $59.93$ & $40.95$ & $19.02$ & $38.39$ & $76.51$ & $85.61$ & $63.37$ & $73.06$ & $82.37$ & $83.93$ & $65.00$ & $60.89$ & $17.52$ & $60.51$ & $59.70$ & $61.01$ & $60.03$ & $65.38$ & $23.89$ & $34.31$ & $26.78$ \\
\ref{tab:multitask} & Examples-proportional, $K=2^{16}$ & $80.45$ & $42.07$ & $91.97$ & $90.97$ & $87.50$ & $85.41$ & $85.04$ & $86.89$ & $90.10$ & $83.01$ & $83.66$ & $90.74$ & $72.56$ & $41.16$ & $19.04$ & $38.59$ & $77.25$ & $85.72$ & $69.95$ & $76.67$ & $86.38$ & $89.29$ & $70.00$ & $65.93$ & $27.91$ & $62.78$ & $61.95$ & $76.90$ & $65.83$ & $73.08$ & $24.35$ & $34.99$ & $27.10$ \\
\ref{tab:multitask} & Examples-proportional, $K=2^{17}$ & $81.56$ & $47.35$ & $91.40$ & $91.55$ & $88.24$ & $86.15$ & $85.93$ & $86.94$ & $90.06$ & $82.76$ & $84.12$ & $90.79$ & $75.09$ & $41.06$ & $19.12$ & $38.47$ & $77.00$ & $85.87$ & $67.91$ & $77.89$ & $77.54$ & $85.71$ & $57.00$ & $67.78$ & $27.07$ & $61.51$ & $60.54$ & $79.06$ & $65.20$ & $74.04$ & $24.36$ & $35.00$ & $27.25$ \\
\ref{tab:multitask} & Examples-proportional, $K=2^{18}$ & $81.67$ & $46.85$ & $91.63$ & $91.99$ & $88.73$ & $87.68$ & $87.20$ & $86.93$ & $90.35$ & $83.30$ & $84.01$ & $91.47$ & $73.29$ & $40.96$ & $19.07$ & $38.43$ & $78.17$ & $86.74$ & $67.94$ & $76.57$ & $78.88$ & $87.50$ & $62.00$ & $67.70$ & $30.85$ & $63.43$ & $62.54$ & $76.53$ & $65.67$ & $67.31$ & $24.57$ & $35.19$ &
 $27.39$ \\
\ref{tab:multitask} & Examples-proportional, $K=2^{19}$ & $81.42$ & $45.94$ & $91.63$ & $92.20$ & $89.22$ & $88.44$ & $88.32$ & $86.84$ & $90.10$ & $83.73$ & $84.29$ & $91.84$ & $70.40$ & $41.26$ & $19.24$ & $38.71$ & $79.78$ & $88.15$ & $67.30$ & $75.66$ & $75.59$ & $87.50$ & $59.00$ & $68.22$ & $30.64$ & $65.32$ & $64.29$ & $73.65$ & $65.05$ & $69.23$ & $25.21$ & $36.30$ & $27.76$ \\
\ref{tab:multitask} & Examples-proportional, $K=2^{20}$ & $80.80$ & $42.55$ & $92.78$ & $91.27$ & $87.99$ & $88.36$ & $88.10$ & $86.10$ & $89.62$ & $84.15$ & $84.26$ & $92.20$ & $68.95$ & $41.05$ & $19.24$ & $38.46$ & $80.36$ & $88.27$ & $67.38$ & $73.21$ & $76.18$ & $83.93$ & $62.00$ & $67.57$ & $26.86$ & $66.12$ & $65.22$ & $76.90$ & $64.73$ & $69.23$ & $25.66$ & $36.93$ & $27.68$ \\
\ref{tab:multitask} & Examples-proportional, $K=2^{21}$ & $79.83$ & $44.45$ & $91.28$ & $89.00$ & $84.31$ & $87.54$ & $87.40$ & $84.93$ & $88.53$ & $82.54$ & $84.16$ & $90.85$ & $67.87$ & $40.51$ & $18.79$ & $37.92$ & $79.50$ & $87.48$ & $65.10$ & $71.16$ & $68.88$ & $85.71$ & $57.00$ & $62.75$ & $23.40$ & $64.50$ & $63.65$ & $72.92$ & $64.11$ & $71.15$ & $25.82$ & $37.22$ & $27.13$ \\
\ref{tab:multitask} & Temperature-scaled, $T = 2$ & $81.90$ & $54.00$ & $91.74$ & $90.56$ & $86.76$ & $85.11$ & $84.60$ & $86.40$ & $89.74$ & $83.47$ & $84.15$ & $91.51$ & $72.56$ & $41.09$ & $19.28$ & $38.54$ & $79.42$ & $87.77$ & $69.92$ & $76.73$ & $92.37$ & $92.86$ & $57.00$ & $69.80$ & $31.90$ & $66.65$ & $65.74$ & $72.92$ & $67.08$ & $75.96$ & $25.42$ & $36.72$ & $27.20$ \\
\ref{tab:multitask} & Temperature-scaled, $T = 4$ & $80.56$ & $45.38$ & $91.97$ & $89.68$ & $85.78$ & $83.13$ & $82.76$ & $86.39$ & $90.00$ & $82.78$ & $84.19$ & $91.16$ & $73.65$ & $41.09$ & $19.22$ & $38.51$ & $77.99$ & $86.81$ & $69.54$ & $76.76$ & $97.36$ & $96.43$ & $59.00$ & $68.10$ & $31.48$ & $64.26$ & $63.27$ & $74.73$ & $64.26$ & $71.15$ & $25.04$ & $35.82$ & $27.45$ \\
\ref{tab:multitask} & Temperature-scaled, $T = 8$ & $77.21$ & $40.07$ & $91.06$ & $88.11$ & $83.33$ & $79.20$ & $79.06$ & $86.60$ & $89.90$ & $83.05$ & $83.56$ & $90.21$ & $59.93$ & $41.01$ & $19.10$ & $38.40$ & $77.14$ & $85.99$ & $66.07$ & $73.94$ & $93.70$ & $94.64$ & $60.00$ & $66.36$ & $26.86$ & $63.46$ & $62.60$ & $62.09$ & $63.32$ & $65.38$ & $24.55$ & $35.35$ & $27.17$ \\
\cmidrule(r){1-2} \cmidrule(lr){3-15} \cmidrule(lr){16-18} \cmidrule(lr){19-20} \cmidrule(lr){21-32} \cmidrule(l){33-35}
\ref{tab:multitask_ft} & \bsl Unsupervised pre-training + fine-tuning & $83.28$ & $53.84$ & $92.68$ & $92.07$ & $88.92$ & $88.02$ & $87.94$ & $88.67$ & $91.56$ & $84.24$ & $84.57$ & $90.48$ & $76.28$ & $41.33$ & $19.24$ & $38.77$ & $80.88$ & $88.81$ & $71.36$ & $76.62$ & $91.22$ & $91.96$ & $66.20$ & $66.13$ & $25.78$ & $69.05$ & $68.16$ & $75.34$ & $68.04$ & $78.56$ & $26.98$ & $39.82$ & $27.65$ \\
\ref{tab:multitask_ft} & Multi-task training & $81.42$ & $45.94$ & $91.63$ & $92.20$ & $89.22$ & $88.44$ & $88.32$ & $86.84$ & $90.10$ & $83.73$ & $84.29$ & $91.84$ & $70.40$ & $41.26$ & $19.24$ & $38.71$ & $79.78$ & $88.15$ & $67.30$ & $75.66$ & $75.59$ & $87.50$ & $59.00$ & $68.22$ & $30.64$ & $65.32$ & $64.29$ & $73.65$ & $65.05$ & $69.23$ & $25.21$ & $36.30$ & $27.76$ \\
\ref{tab:multitask_ft} & Multi-task pre-training + fine-tuning & $83.11$ & $51.42$ & $92.66$ & $91.73$ & $88.73$ & $88.06$ & $87.70$ & $88.61$ & $91.61$ & $84.09$ & $84.31$ & $91.85$ & $76.53$ & $41.15$ & $19.12$ & $38.59$ & $80.26$ & $88.50$ & $71.03$ & $79.54$ & $81.69$ & $87.50$ & $65.00$ & $70.72$ & $31.48$ & $65.94$ & $65.03$ & $81.23$ & $68.18$ & $73.08$ & $27.08$ & $39.80$ & $28.07$ \\
\ref{tab:multitask_ft} & Leave-one-out multi-task training & $81.98$ & $48.00$ & $93.23$ & $91.72$ & $88.24$ & $87.76$ & $87.32$ & $88.61$ & $91.44$ & $84.00$ & $84.11$ & $90.79$ & $72.20$ & $41.34$ & $19.05$ & $38.77$ & $79.97$ & $88.10$ & $71.68$ & $78.35$ & $86.76$ & $89.29$ & $66.00$ & $68.09$ & $29.49$ & $66.23$ & $65.27$ & $79.06$ & $68.65$ & $78.85$ & $26.93$ & $39.79$ & $27.87$ \\
\ref{tab:multitask_ft} & Supervised multi-task pre-training & $79.93$ & $36.60$ & $92.43$ & $91.58$ & $88.24$ & $87.03$ & $86.78$ & $88.15$ & $91.20$ & $82.87$ & $83.16$ & $90.13$ & $70.76$ & $41.12$ & $18.96$ & $38.49$ & $77.38$ & $85.65$ & $65.36$ & $75.66$ & $68.87$ & $83.93$ & $58.00$ & $64.81$ & $21.93$ & $55.37$ & $54.61$ & $71.12$ & $67.40$ & $75.96$ & $26.81$ & $40.13$ & $28.04$ \\
\cmidrule(r){1-2} \cmidrule(lr){3-15} \cmidrule(lr){16-18} \cmidrule(lr){19-20} \cmidrule(lr){21-32} \cmidrule(l){33-35}
\ref{tab:scaling} & \bsl Baseline & $83.28$ & $53.84$ & $92.68$ & $92.07$ & $88.92$ & $88.02$ & $87.94$ & $88.67$ & $91.56$ & $84.24$ & $84.57$ & $90.48$ & $76.28$ & $41.33$ & $19.24$ & $38.77$ & $80.88$ & $88.81$ & $71.36$ & $76.62$ & $91.22$ & $91.96$ & $66.20$ & $66.13$ & $25.78$ & $69.05$ & $68.16$ & $75.34$ & $68.04$ & $78.56$ & $26.98$ & $39.82$ & $27.65$ \\
\ref{tab:scaling} & $1\times$ size, $4\times$ training steps & $85.33$ & $60.29$ & $93.81$ & $94.06$ & $91.67$ & $89.42$ & $89.25$ & $89.15$ & $91.87$ & $86.01$ & $85.70$ & $91.63$ & $78.34$ & $41.52$ & $19.33$ & $38.96$ & $82.45$ & $90.19$ & $74.72$ & $79.17$ & $94.75$ & $92.86$ & $71.00$ & $67.34$ & $29.70$ & $72.63$ & $71.59$ & $78.34$ & $72.10$ & $82.69$ & $27.08$ & $40.66$ & $27.93$ \\
\ref{tab:scaling} & $1\times$ size, $4\times$ batch size & $84.60$ & $56.08$ & $93.12$ & $92.31$ & $89.22$ & $88.85$ & $88.84$ & $89.35$ & $92.07$ & $85.98$ & $86.13$ & $91.07$ & $80.14$ & $41.70$ & $19.42$ & $39.08$ & $82.52$ & $90.21$ & $74.64$ & $78.78$ & $93.69$ & $94.64$ & $72.00$ & $68.09$ & $30.95$ & $74.73$ & $73.90$ & $76.53$ & $70.06$ & $81.73$ & $27.07$ & $40.60$ & $27.84$ \\
\ref{tab:scaling} & $2\times$ size, $2\times$ training steps & $86.18$ & $62.04$ & $93.69$ & $93.36$ & $90.69$ & $89.18$ & $89.23$ & $89.35$ & $92.05$ & $87.23$ & $87.05$ & $92.68$ & $81.95$ & $41.74$ & $19.66$ & $39.14$ & $84.18$ & $91.29$ & $77.18$ & $80.98$ & $97.36$ & $96.43$ & $74.00$ & $71.34$ & $35.68$ & $77.11$ & $76.34$ & $80.51$ & $69.28$ & $85.58$ & $27.52$ & $41.03$ & $28.19$ \\
\ref{tab:scaling} & $4\times$ size, $1\times$ training steps & $85.91$ & $57.58$ & $94.38$ & $92.67$ & $89.95$ & $89.60$ & $89.60$ & $89.44$ & $92.14$ & $87.05$ & $87.12$ & $93.12$ & $83.39$ & $41.60$ & $19.73$ & $39.08$ & $83.86$ & $91.32$ & $78.04$ & $81.38$ & $89.09$ & $94.64$ & $73.00$ & $73.74$ & $40.40$ & $78.25$ & $77.40$ & $81.59$ & $70.22$ & $91.35$ & $27.47$ & $40.71$ & $28.10$ \\
\ref{tab:scaling} & $4\times$ ensembled & $84.77$ & $56.14$ & $93.46$ & $93.31$ & $90.67$ & $89.71$ & $89.60$ & $89.62$ & $92.24$ & $86.22$ & $86.53$ & $91.60$ & $77.98$ & $42.10$ & $20.10$ & $39.56$ & $83.09$ & $90.40$ & $71.74$ & $77.58$ & $89.85$ & $91.07$ & $66.00$ & $69.32$ & $29.49$ & $72.67$ & $71.94$ & $76.90$ & $69.12$ & $72.12$ & $28.05$ & $40.53$ & $28.09$ \\
\ref{tab:scaling} & $4\times$ ensembled, fine-tune only & $84.05$ & $54.78$ & $92.78$ & $93.15$ & $90.44$ & $88.34$ & $88.12$ & $89.27$ & $91.97$ & $85.33$ & $85.88$ & $90.98$ & $77.62$ & $41.66$ & $19.57$ & $39.12$ & $82.36$ & $89.86$ & $71.56$ & $77.43$ & $90.07$ & $92.86$ & $69.00$ & $67.31$ & $26.34$ & $70.47$ & $69.64$ & $75.45$ & $68.18$ & $74.04$ & $27.55$ & $40.22$ & $28.09$ \\
\bottomrule
\end{tabular}
\caption{
Score achieved on every task we consider for all of the experiments in this paper.
In the first column, we list the table where the condensed results were presented for a given experiment.
As in the main text, a row marked with $\bigstar$ denotes our baseline model (described in \cref{sec:baseline}).
}\label{tab:giant}
\end{minipage}
\end{table}

\clearpage

\eject \pdfpagewidth=8.5in \pdfpageheight=11in

\bibliography{biblio}

\end{document}